\documentclass[accepted]{uai2026}
          \usepackage{mathtools,siunitx,booktabs,tikz,multirow,soul,comment,hyperref,amsbsy,amssymb,amsthm,amsfonts,bbm,textcomp,graphicx,wrapfig,float,subcaption,algorithm,algpseudocode}

\usepackage[american]{babel}
\usepackage{natbib} 

\newtheorem{definition}{Definition}

\title{Beyond performance-wise Contribution Evaluation in Federated Learning}

\author[1]{\href{mailto:<pejo@crysys.hu>}{Balazs Pejo}{}}
\affil[1]{CrySyS Lab, Hungary}
  
\begin{document}

\maketitle

\begin{abstract}
    Federated learning offers a privacy-friendly collaborative learning framework, yet its success, like any joint venture, hinges on the contributions of its participants. Existing client evaluation methods predominantly focus on model performance, such as accuracy or loss, which represents only one dimension of a machine learning model's overall utility. In contrast, this work investigates the critical, yet overlooked, issue of client contributions towards a model's trustworthiness --- specifically, its reliability (tolerance to noisy data), resilience (resistance to adversarial examples), and fairness (measured via demographic parity). To quantify these multifaceted contributions, we employ the state-of-the-art approximation of the Shapley value, a principled method for value attribution. Our results reveal that no single client excels across all dimensions, which are largely independent from each other, highlighting a critical flaw in current evaluation scheme: no single metric is adequate for comprehensive evaluation and equitable rewarding allocation.
\end{abstract}

\section{Introduction}

The proliferation of digital technology has enabled the collection of vast amounts of user data, fueling the rapid advancement of Machine Learning (ML). However, this data is often fragmented into isolated silos, limiting the full potential of ML to entities with access to large, centralized datasets and potentially leading to biased or suboptimal models. Naive data pooling to overcome this fragmentation is often infeasible, as it can lead to severe privacy violations (i.e., GDPR) and erode competitive advantages for businesses where data is a key asset.

To address these challenges and democratize AI, Federated Learning (FL)~\cite{mcmahan2017communication} has emerged as a powerful paradigm. It enables multiple parties to collaboratively train a shared model without exposing their raw data, thus preserving privacy while improving model generalization. However, managing this collaborative process requires mechanisms to ensure its stability and success~\cite{kairouz2021advances}. An appropriate mechanism for this is data and client evaluation, which is foundational to building a trustworthy system.

Indeed, data (client) evaluation intersects with several dimensions of trustworthiness; for instance, quantifying clients is essential for ensuring accountability and transparency in the federation~\cite{siomos2023contribution}, while related valuation techniques that assess feature importance are a cornerstone of model explainability and interpretability~\cite{gunning2019xai}. Furthermore, client data directly impacts model's privacy~\cite{liu2021machine}, fairness~\cite{mehrabi2021survey}, reliability~\cite{goldblum2022dataset}, and resilience~\cite{ibitoye2019threat}. These dimensions are often intertwined, creating a complex web of trade-offs~\cite{liu2020privacy, shaham2023holistic, li2024triangular}. In this paper, we focus on last three of these critical objectives:

\begin{itemize}
    \item \textbf{Fairness}: Ensuring the model avoids discrimination against groups based on sensitive attributes.
    \item \textbf{Reliability}: Ensuring the model performs consistently, particularly when faced with noisy or corrupted data.
    \item \textbf{Resilience}: Ensuring the model maintains its performance when subjected to adversarial attacks.
\end{itemize}

From a regulatory and governance perspective, our focus is aligned with the European Trustworthy AI framework~\cite{kaur2022trustworthy} and the European Union's Artificial Intelligence Regulation (AI Act), which emphasize that AI systems must be simultaneously ethically acceptable and technically robust. 

Despite the importance of these multifaceted goals, the critical flaw of all existing evaluation methods is that they almost exclusively rely on a single performance metric, typically accuracy or model loss~\cite{naidu2023review}. This creates a narrow, and potentially misleading, view of a client's true value. For instance, most works apply data valuation with the singular goal of improving training efficiency (e.g., faster convergence or more accurate model) by weighting higher/lower or selecting more/less frequently high/low quality data/client, respectively. This paper challenges this performance-centric paradigm, arguing that a client's contribution to model trustworthiness is an equally, if not more, important consideration.

\paragraph{Contributions. }

We conduct a multifaceted analysis of client contributions in FL, extending the evaluation to key trustworthiness metrics. Our primary contributions are as follows:

\begin{itemize}
    \item We evaluate the contribution of FL clients based on four distinct aspects of the global model: performance (measured by accuracy), fairness (measured by demographic parity), reliability (measured by the tolerance to noise), and resilience (measured by the resistance against adversarial examples).
    \item We shed light on the independence of these trustworthy AI pillars: the various scores allocated to the same clients are statistically uncorrelated.
    \item We apply two distinct and well-established contribution evaluation schemes, the lightweight Leave-One-Out and the state-of-the-art Guided-Truncated-Gradient Shapley approximation, to demonstrate that our findings are present across different valuation methodology.
    \item We conduct a systematic empirical study across three different ML domains (tabular data, computer vision, and natural language processing) to validate the generalizability of our conclusions.
\end{itemize}

Our core finding is that client rankings differ dramatically across evaluation metrics. We provide compelling evidence that clients who are highly valuable for model performance can be neutral or even detrimental to its fairness, reliability, or resilience, and vice versa. This demonstrates that a performance-wise evaluation paints a skewed and incomplete picture, with profound implications for building trustworthy FL systems.

\paragraph{Organization. }

The remainder of this paper is organized as follows. Section~\ref{sec:RW} reviews related work, while in n Section~\ref{sec:mod}, we detail our multifaceted evaluation metrics. In Section~\ref{sec:exp}, we describe our experimental setup and present our empirical results along with our findings. Finally, we discuss the broader implications of our work and conclude the paper in Section~\ref{sec:con}.

\section{Related Works}
\label{sec:RW}

\subsection{Trustworthy AI}

As machine learning systems transition from research to integral components of high-stakes societal functions, the development paradigm has shifted from optimizing solely for predictive performance to ensuring overall trustworthiness~\cite{chkirbene2024large}. This is particularly critical in domains where model failures can have severe economic and social consequences. Trustworthiness typically encompasses several pillars, within this work we focus on three: 1) Fairness, which seeks to prevent unjust or prejudicial treatment, 2) Reliability, which ensure consistent operation under uncertain conditions, and 3) Resilience, which mitigates the impact of adverse manipulations. 

\paragraph{Fairness. }

Machine learning models, despite their mathematical objectivity, are trained on data generated by a society with historical and systemic biases~\cite{mehrabi2021survey}. Consequently, these models can inadvertently learn, reproduce, and even amplify these biases, leading to discriminatory outcomes in high-stakes applications~\cite{hall2022systematic}. It is not only an ethical imperative but also an emerging legal requirement, making it a critical component of trustworthy AI~\cite{wachter2020bias}. The concept of fairness is operationalized through various quantitative metrics, each capturing a different notion of equity. In this work, we focus on one of the most fundamental and widely understood metrics: Demographic Parity~\cite{calders2009building}, which requires that the model's prediction is statistically independent of the sensitive attribute. 

\paragraph{Robustness. }

While often discussed together, reliability and resilience are distinct parts of robustness within Trustworthy AI that address different types of data perturbations~\cite{fawzi2016robustness}. Reliability concerns a model's ability to perform consistently when faced with naturally occurring, random noise or data corruption~\cite{shannon1948}. This simulates real-world challenges like blurry images from camera shake, background noise in audio recordings, or simple data entry errors and the occurring white noise during communication. Resilience, in contrast, measures a model's tolerance against deliberately crafted, worst-case perturbations, commonly known as adversarial examples~\cite{goodfellow2014explaining}. These attacks introduce small, often imperceptible changes to an input with the specific goal of deceiving the model and causing a misclassifications. In this work, we use the well-known method called Projected Gradient Descent~\cite{madry2017towards} to generate such samples.

Despite this clear conceptual difference, the two terms are frequently conflated in the literature, often grouped under the umbrella term robustness. This confusion is reflected in the metrics used for their evaluation. For instance, Mean Corruption Error~\cite{mCE} and Effective Robustness~\cite{effectiverelative} focus on reliability, while Adversarial Accuracy~\cite{23metric} is tailored to measure resilience. 

Besides this confusion, a more significant limitation is that most of these metrics are heavily confounded by the model's (baseline) accuracy. A model with higher general performance will almost always score better on corrupted or adversarial data, making it difficult to discern whether the model is truly more reliable or simply more accurate to begin with. This effect makes it challenging to evaluate a client's specific contribution to the model's reliability or resilience, separate from their contribution to its overall performance. To address this gap, our work explicitly treats reliability and resilience as separate objectives. We seek to evaluate them using methods that isolate these properties from the model's general predictive accuracy. This approach allows for a more direct and nuanced assessment of a client's contribution to protecting the model against either naturally occurring random noise or specific adversarial threats.

\subsection{Contribution Evaluation}

Evaluating the contribution of individual clients to a collaborative model is a critical task. While a handful of strategies exist to tackle this~\cite{huang2020exploratory,jiang2023opendataval}, the cornerstone technique for most quality assessment problems within FL~\cite{jain2020overview} is the Shapley Value (SV)~\cite{winter2002shapley,rozemberczki2022shapley}. Originating from cooperative game theory~\cite{shapley1951notes}, The Shapley Value of a client is defined as their average marginal contribution to the model's utility across all possible subsets (coalitions) of clients. While it possesses desirable axiomatic properties that guarantee a fair and unique distribution, its primary drawback is its computational complexity, which makes exact computation infeasible for typical FL scenarios. Consequently, a variety of approximation methods have been developed, such as Monte-Carlo based approaches and gradient-based techniques~\cite{ghorbani2019data,liu2022gtg}, which aim to estimate the value in a computationally tractable manner.

\paragraph{Fairness. }

The term "fairness" carries a dual meaning in this context, referring to both the properties of the reward allocation scheme (game-theoretic fairness~\cite{rabin1993incorporating}) and the statistical properties of the machine learning model (algorithmic fairness~\cite{kleinberg2018algorithmic}). This creates a twofold intersection with contribution evaluation. First, one can assess the fairness of the allocation scheme itself. The Shapley Value is axiomatically the only "fair" reward allocation method; thus, any other scheme, including its approximations, inevitably compromises on some aspect of this game-theoretic fairness. Second, one can evaluate the algorithmic fairness of the trained model and allocate value to clients based on their contribution to that specific objective. 

Since the quality of data and clients has long been understood to extend beyond simple accuracy~\cite{wang1996beyond}, this second interpretation is our focus. Our work is inspired by recent efforts to view algorithmic fairness through a data-quality lens. The authors in~\cite{arnaiztowards} assign importance scores for individual training samples based on their contribution to a model's fairness, rather than its performance. We bridge this concept with FL, extending the unit of analysis from a single data point to an entire client. 



\paragraph{Robustness. }

The relationship between contribution evaluation and model robustness is also bidirectional. On one hand, research has explored how the evaluation process itself can be vulnerable. For example,the authors in~\cite{xu2024ace,pejo2025fragility} have demonstrated that contribution scores can be manipulated through data poisoning attacks, highlighting a lack of resilience in the evaluation schemes themselves. On the other hand, a perpendicular line of work uses contribution evaluation as a tool to enhance the final model's resilience, for instance, by identifying and up-weighting clients who contribute positively to this objective~\cite{guan2022few,otmani2024fedsv}.

Beyond adversarial manipulations, the inherent reliability of contribution scores is also a significant concern. The scores, including the Shapley Value estimates, are often subject to uncertainty stemming from the stochastic nature of both the approximation algorithms and the model training process. This instability has been studied from several angles: the authors in~\cite{pejo2025fragility} examined the effect of different aggregation techniques while others~\cite{goldwasser2024stabilizing} analyzed the randomness of the approximation mechanism. Yet, another work~\cite{wang2023data} investigated the consistency of different value functions against natural randomness. 

Lastly, Shapley-based rewards can motivate participants to strategically split their data to increase their payout, ultimately degrading the global model's performance in a classic "tragedy of the commons" scenario~\cite{qi2024mechanism}. While the existing literature has focused on the reliability and resilience of the evaluation schemes, our work addresses a different question. We shift the focus to using these schemes to evaluate clients based on their contribution to the model's reliability and resilience, treating these as explicit objectives alongside performance and fairness.

\section{Trustworthiness Scores}
\label{sec:mod}


\paragraph{Notation. }

The objective of ML is to iteratively (i.e., $1\leq t \leq T$) train a model $M(\theta)^t$ with parameters $\theta=\{\theta_1,\dots,\theta_m\}$ and with empirical loss function $\mathcal{L}(x,\theta)$ (where $x$ is a sample) such that the prediction is enhanced with respect to a specific task. During training the empirical loss is minimized: $\min_\theta \sum_{x\in D}\mathcal{L}(x,\theta)$. A summary of the utilized symbols are presented in Table~\ref{tab:notations}, which also contains the variables specific to the proposed evaluation schemes.

\begin{table}[!b]
    \centering
    \begin{tabular}{c|l}
        Symbol & Description \\
        \hline
        $k\in[K]$ & FL clients \\
        $M_k^t$ \& $M^t$ & The local and global models after round $t$ \\
        $D_k$ \& $D^\prime$ & Training \& Testing datasets \\
        $x$ \& $y$ & Data sample \& label \\
        $n$ \& $a$ & Gaussian \& malicious noise \\
        $v(\cdot)$ & Evaluation function for scoring \\ 
        & $v\in \{\mathtt{perf}, \mathtt{fair}, \mathtt{rel}, \mathtt{res}\}$\\
        $\mathtt{CS}^t_v$ & Contribution score in round $t$ based on $v$ \\
        & $\mathtt{CS}^t_v\in\{\mathtt{GTG}^t_v$ \& $\mathtt{LOO}^t_v$\} \\
    \end{tabular}
    \caption{Notations use in the paper.}
    \label{tab:notations}
\end{table}

\paragraph{Contribution Evaluation. }

Before defining our trustworthiness scores, we first outline the classical approach to client evaluation, which is based on model performance. The gold standard for this task is the Shapley Value: for a set of $K$ clients, the $\mathtt{SV}$ for client $k$ is defined in Eq.~\eqref{eq:GTG}, where $\Pi(K)$ is the set of all permutations of the $K$ clients, and $\pi_{<k}$ denotes the set of clients preceding $k$ in a given permutation $\pi$. Note, we present the formula to be suitable with evaluation functions $v(\cdot)$ where higher values mean better model. Indeed, for performance-wise evaluation, $v(\cdot)$ is usually defined as the accuracy. As calculating the exact value requires exponentially many model trainings (one for each possible client combination), we utilize Data Shapley~\cite{ghorbani2019data} instead, which only requires a single model to be trained: in each round all possible combination of gradients are used to update the model and the marginal differences are computed by comparing these intermediate submodels. The final score is the accumulated score from all rounds. 

\begin{equation}
    \label{eq:GTG}
    \mathtt{SV}_v(k) = \frac{1}{K!} \sum_{\pi \in \Pi(K)} \left( v(\pi_{<k} \cup \{k\}) - v(\pi_{<k}) \right)
\end{equation}


Guided-Truncated-Gradient Shapley ($\mathtt{GTG}$)~\cite{liu2022gtg}, the SotA approximation technique enhances this by 1) skipping entire rounds of FL if the global model's improvement is negligible ($<\epsilon_1$), 2) it samples a portion ($\epsilon_2$) of all subset from all possible client permutations (such that each client is equal times in the early places), and 3) for each sampled permutation, it only evaluates the marginal contributions of a client if they exceed a minimum threshold ($>\epsilon_3$). 

Another, even more lightweight and widely used alternative approach what we also utilize in this paper is the Leave-One-Out ($\mathtt{LOO}$) heuristic~\cite{evgeniou2004leave}. While less principled than GTG, it is computationally much cheaper. A client's contribution is simply the change in utility observed when they are removed from the grand coalition, as shown in Eq.~\eqref{eq:LOO}. 

\begin{equation}
    \label{eq:LOO}
    \mathtt{LOO}_v(k) = v([K]) - v([K]\setminus\{k\})
\end{equation}

\paragraph{Performance Scoring. }

\begin{figure}[!b]
    \centering
    \includegraphics[width=\linewidth]{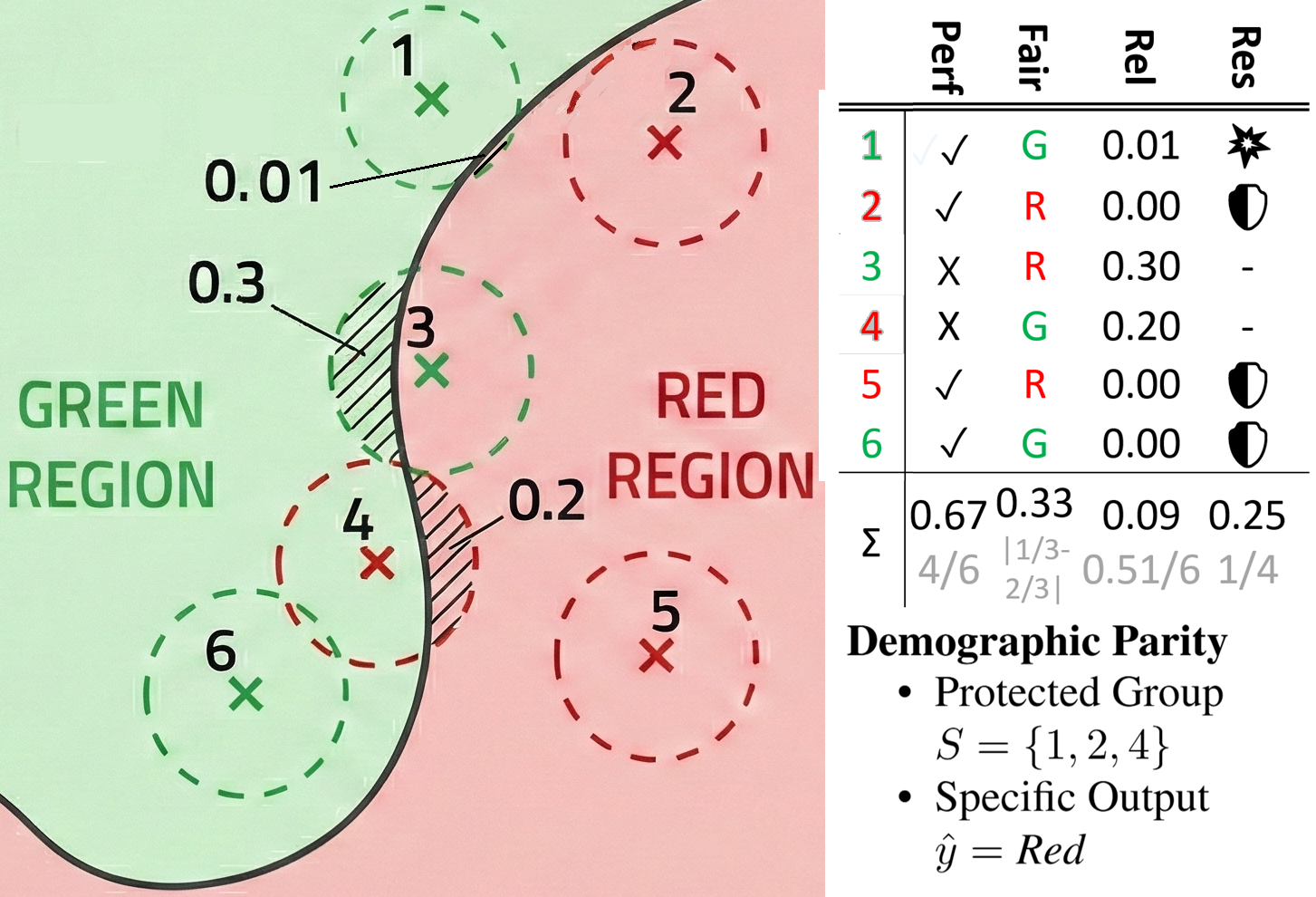}
    \caption{Example to illustrate what the different trustworthiness scores capture based on $D^\prime=\{(1,G),(2,R),(3,G),(4,R),(5,R),(6,G)\}$. While the model is $0.67$ accurate, its trustworthiness scores are widely different, e.g., $\mathtt{fair}(M)=0.33$ when Demographic Parity is used where the square numbers (1,2,4) are protected and 'Red' is the target class. Regarding reliability, the model might misclassify three (1,3,4) samples with various probabilities, resulting in $\mathtt{rel}(M)=0.09$ overall chance. Regarding resilience, the attacker is successful once (1) out of the four (1,2,5,6) correctly classified samples, i.e., $\mathtt{res}(M)=0.25$. Note, in the paper we transform these trustworthy metrics (i.e., $m\rightarrow(1-m)$) to be similar to accuracy: higher values mean better model. }
    \label{fig:illustrate}
\end{figure}

As far as we know, without a single exception, $v(\cdot)$ in the literature is always defined by some standard ML metric measuring the performance of the model, i.e., the accuracy/loss. We will use the former as the baseline (Eq.~\eqref{eq:perf}) and adapt this evaluation schemes to capture the model's fairness, reliability, and resilience, independently of its underlying performance. A visual illustration what each captures is presented in Fig.~\ref{fig:illustrate}, while the high level pseudo code encapsulating our scoring methodology is presented in Alg.~\ref{alg:alg}. 

\begin{equation}
    \label{eq:perf}
    \mathtt{perf}(M)=\Pr_{x\in D^\prime}[M(x) = y]
\end{equation} 

\begin{algorithm}[!t]
    \caption{Assigning scores to clients in round $t$.}
    \label{alg:alg}
    \textbf{Input}: $M^{t-1}$, $[D_k]_1^K$, $D^\prime$ \\
    \textbf{Output}: $[[\mathtt{CS}^t_{v}(D_k)]_{k=1}^K]_{v\in\{\mathtt{perf}, \mathtt{fair}, \mathtt{rel}, \mathtt{res}\}}$
    \begin{algorithmic}[1]
    \For{each client $k \in [K]$}
        \State $M^t_k \leftarrow Train(M^{t-1})$
        \State Share $M_k^t$ with the server
    \EndFor
    \For{$\mathtt{CS}\in\{\mathtt{GTG}, \mathtt{LOO}\}$}
        \State \# $\mathtt{GTG}$ via Eq.~\eqref{eq:GTG}, $\mathtt{LOO}$ via Eq.~\eqref{eq:LOO}
        \For{$v\in\{\mathtt{perf}, \mathtt{fair}, \mathtt{rel}, \mathtt{res}\}$}
            \State \# $\mathtt{perf}$/$\mathtt{fair}$/$\mathtt{rel}$/$\mathtt{res}$ via Eq.~\eqref{eq:perf}/~\eqref{eq:fair}/~\eqref{eq:rel}/~\eqref{eq:rob}
            \State $[\mathtt{CS}^t_v(D_k)]_1^K \leftarrow [M_k^t]_1^K$
        \EndFor
    \EndFor
    \end{algorithmic}
\end{algorithm}

\paragraph{Fairness Scoring. }

Our fairness score is based on Demographic Parity (Def.~\ref{def:DP}), which relies on the test dataset. It ensures that the proportion of samples assigned to a particular class by the model must be independent from its membership of the sensitive group. 

\begin{definition}[Demographic Parity~\cite{calders2009building}]
    \label{def:DP}
    For sensitive group $S\in D^\prime$ and for a specific outputs $\hat{y}$ Demographic Parity requires the formula below to hold:
    \begin{equation*}
         \Pr_{x\in D^\prime}[M(x)=\hat{y} | x\in S] =
         \Pr_{x\in D^\prime}[M(x)=\hat{y} | x\in D^\prime/S] 
    \end{equation*}
\end{definition}
    
We assign the fairness score to model $M$ such that we take the difference between these, as formalized in Eq.~\eqref{eq:fair}. Here, by subtracting the absolute difference from 1, higher values indicate a fairer model.

{\small\begin{equation}
    \begin{aligned}
    \label{eq:fair}
    \mathtt{fair}(M) = 1-\\
    \left|\Pr_{x\in D^\prime}[M(x)=\hat{y} | x\in S] -\Pr_{x\in D^\prime}[M(x)=\hat{y} | x\in D^\prime/S]\right|
    \end{aligned}
\end{equation}}

\paragraph{Reliability Scoring. }

To measure reliability, we assess the stability of a model's predictions when subjected to naturally occurring, random noise. To simulate this, we perturb each test sample $x\in D^\prime$ with noise drawn from a Gaussian distribution, $n\sim \mathcal{N}(0,\sigma)$, and compare the prediction between the original and the perturbed input, as formalized in Eq.~\eqref{eq:rel}. A key feature of this reliability metric is its independence from model accuracy. Such consistency focused metric captures the reliability of the trained models. In line with the other scores, it is also in the range $[0, 1]$: $1$ indicates perfect reliability, meaning the model's predictions are unaffected by the noise. 


\begin{equation}
    \label{eq:rel}
    \begin{aligned}
    \mathtt{rel}(M) = \\
    1 - \Pr_{x\in D^\prime}[M(x) \neq M(x+n)|n\sim\mathcal{N}(0,\sigma)]
    \end{aligned}
\end{equation}

\paragraph{Resilience Scoring. }

In contrast to reliability, which measures stability against random noise, our resilience metric quantifies a model's resilience against a deliberate, worst-case adversarial attack. To evaluate this, we adopt a standard methodology from the adversarial machine learning literature. We subject a trained model to a powerful and widely recognized attack, Projected Gradient Descent (PGD)~\cite{madry2017towards}, which is designed to find the minimal perturbation necessary to induce a misclassifications. The resilience score is then defined based on the success rate of this attack on correctly classified samples as defined in Eq.~\eqref{eq:rob}, i.e., 0 implies maximal attack success rate and minimal model resilience, 

\begin{equation}
    \label{eq:rob}
    \begin{aligned}
    \mathtt{res}(M) = \\
    1 -\Pr_{(x,y) \in D'}[M(x) \neq M(x+a)|M(x)=y]
    \end{aligned}
\end{equation}

\section{Experiments}
\label{sec:exp}

\paragraph{Setup. }

To ensure our findings are generalizable, we conduct experiments across three settings: using Cifar10 with a CNN model for Computer Vision, using IMBD with a USE encoder for Natural Language Processing, and using ADULT with an MLP model for Tabular data. Further details are presented in App.~\ref{app:setup}. The training commences for ten rounds but in our score computation we only take into account the last nine: the first corresponds to training an untrained model, thus, the corresponding change would dominate the rest and would skew the scoring results. We simulate a cross-silo setting with $K=4$ and $K=20$ clients. These small numbers are chosen so the individual client contributions remain meaningful, as scores tend to diminish with a large number of participants. To model realistic data heterogeneity, we evaluate clients under both IID and non-IID data distributions (using a Dirichlet distribution with $\alpha=0.5$). For aggregation we applied FedAVG and all experiments are repeated 5 fold. The implementation is based in Python using the Flower framework and TensorFlow, our code is available as open-source under the MIT License: \url{https://anonymous.4open.science/r/RobShap-F707/}.

For reliability, we inject Gaussian noise with a standard deviation of $\sigma=0.1$. For robustness, $l_\infty$-PGD was used with step size $\alpha=0.007$, $\varepsilon=0.3$ with 40 rounds and $\varepsilon=0.0015$ with 5 rounds for CIFAR/ADULT and IMDB, respectively. Regarding fairness, we utilize the \texttt{sex} attribute as the sensitive feature for the ADULT dataset. For CIFAR, where sensitive attributes are not naturally present, following~\cite{wang2020towards} we construct sensitive groups by converting 5\% of the samples per class into gray-scale. Finally, as IMDB also lack such sensitivity annotation, we disregarded it for this purpose. 

\paragraph{Trustworthy Model Scores. }

First we study how the model's various scores change during the training. It is clear that $\mathtt{perf}$ improves, at it is directly optimized. On the other hand, the metrics capturing trustworthiness are not explicitly considered, hence, they evolution is not necessarily improving. This is visualized in Fig.~\ref{fig:score_evol_NIID} for non-IID case (as the IID is similar, it is presented only in the Appendix in Fig.~\ref{fig:score_evol_IID}).

\begin{figure}[!b]
    \centering
    \begin{subfigure}{0.45\textwidth}
        \includegraphics[width=\linewidth]{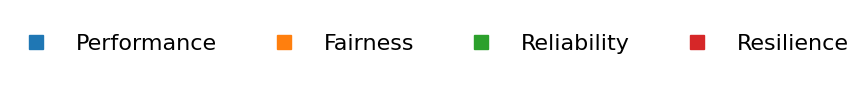}
    \end{subfigure}
    
    \begin{subfigure}{0.22\textwidth}
        \includegraphics[width=\linewidth]{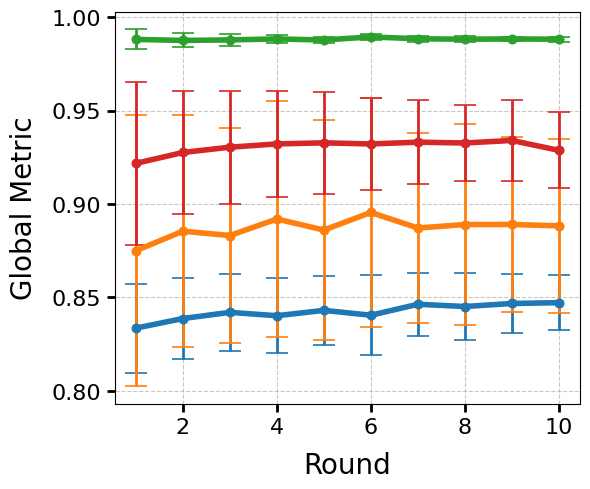}
        \caption{ADULT / 4 clients.}
    \end{subfigure}
    \hfill
    \begin{subfigure}{0.22\textwidth}
        \includegraphics[width=\linewidth]{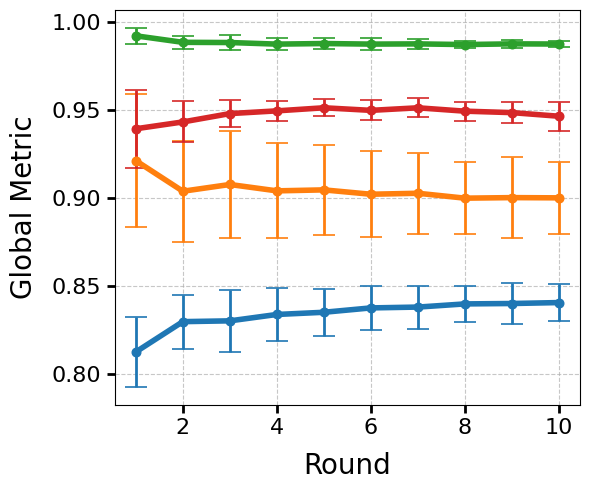}
        \caption{ADULT / 20 clients.}
    \end{subfigure}

    \begin{subfigure}{0.22\textwidth}
        \includegraphics[width=\linewidth]{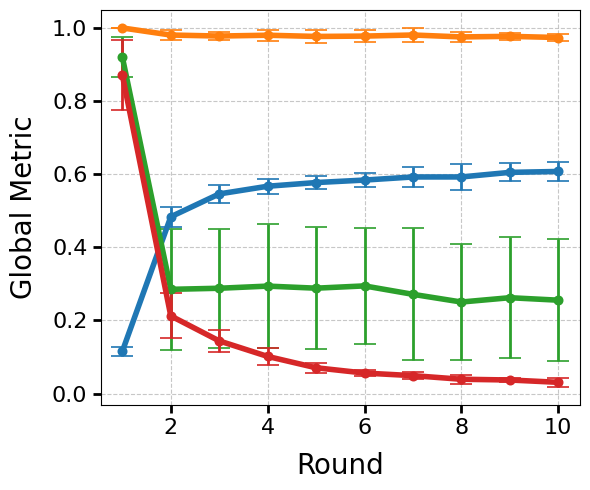}
        \caption{CIFAR / 4 clients.}
    \end{subfigure}
    \hfill
    \begin{subfigure}{0.22\textwidth}
        \includegraphics[width=\linewidth]{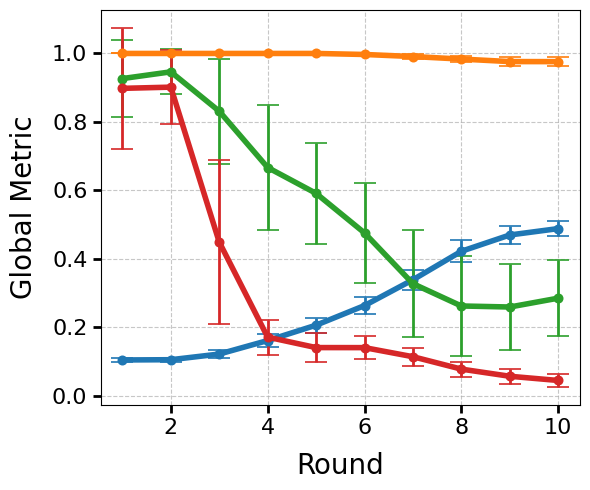}
        \caption{CIFAR / 20 clients.}
    \end{subfigure}

    \begin{subfigure}{0.22\textwidth}
        \includegraphics[width=\linewidth]{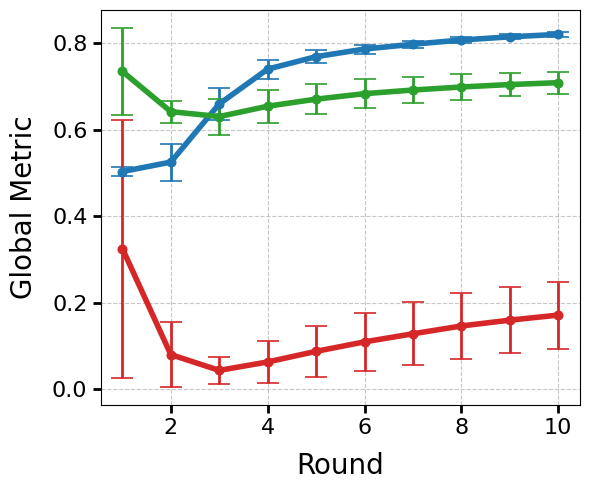}
        \caption{IMDB / 4 clients.}
    \end{subfigure}
    \hfill
    \begin{subfigure}{0.22\textwidth}
        \includegraphics[width=\linewidth]{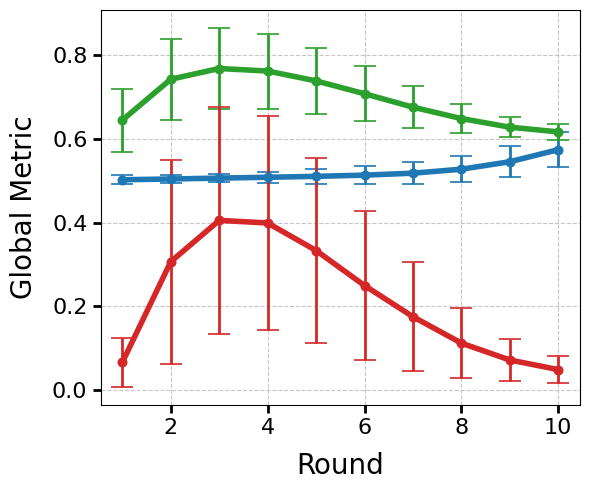}
        \caption{IMDB / 20 clients.}
    \end{subfigure}
    \caption{Score evolution of the model as the training progresses for the non-IID scenario.}
    \label{fig:score_evol_NIID}
\end{figure}

Here, one can see widely different trends across datasets. In case of ADULT, all scores are rather stable, which can be explained by the limited change in the model in each round (2-10): this simple task was already learned well in the first round, leaving no room for further significant changes. 

In contrast, considering CIFAR, the score trajectories are mostly different. Clearly, the $\mathtt{perf}$ score is increasing, as the model is explicitly optimized for that. On the other hand, the $\mathtt{fair}$ score is rather stationary and extremely high (due to the artificial uniform nature of creating the sensitivity class), implying the model is fair. Regarding the $\mathtt{rel}$ and $\mathtt{res}$ scores, they are both robustness metrics, so, not surprisingly, they are following a similar trend. The training process disregards these aspects, but instead of remaining static, these scores are deteriorating significantly: the model after the first round is quite robust ($\approx0.9$), but later these properties vanish ($\approx0.05-0.3$).

For the IMDB dataset, where the $\mathtt{fair}$ score is missing, changes in reliability ($0.6<\mathtt{rel}<0.8$) are considerably more moderate compared to resistance ($0.05<\mathtt{res}<0.4$), and neither of them show a clear monotone trend as before. These results shows that trustworthiness metric can behave differently in different settings, and even similar metric can paint a different picture about the same setting. While we advocate for a more and deeper research into the model's trustworthiness scores, our paper focuses on the client's scores, so we leave this as future work. 

\paragraph{Trustworthy Client Scores. }

\begin{figure}[!b]
\centering
    \begin{subfigure}{0.45\textwidth}
        \includegraphics[width=\linewidth]{figs/S_legend.png}
    \end{subfigure}
    \begin{subfigure}{0.22\textwidth}
        \includegraphics[width=\linewidth]{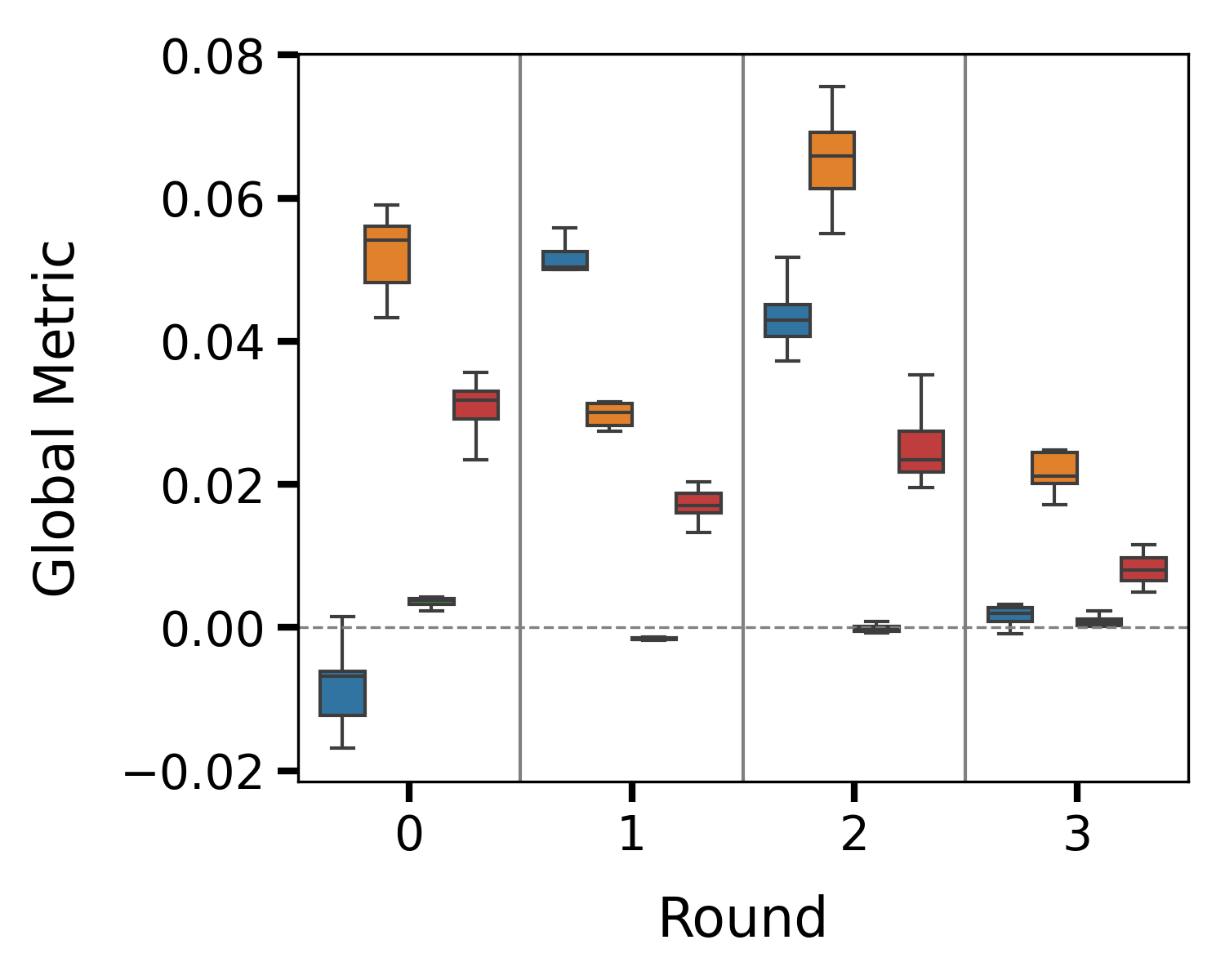}
        \caption{ADULT / LOO.}
    \end{subfigure}
    \hfill
    \begin{subfigure}{0.22\textwidth}
        \includegraphics[width=\linewidth]{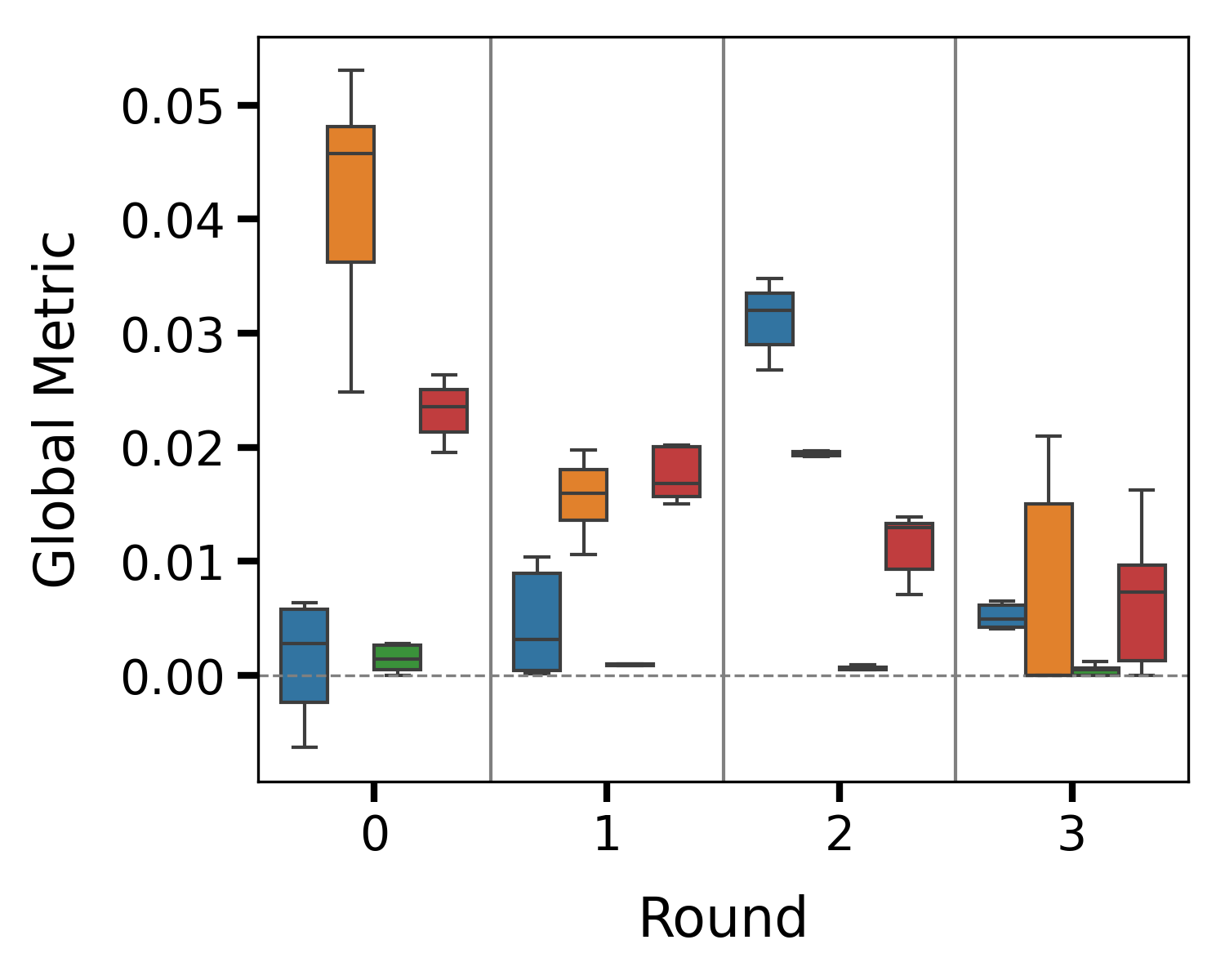}
        \caption{ADULT / GTG.}
    \end{subfigure}
    
    \begin{subfigure}{0.22\textwidth}
        \includegraphics[width=\linewidth]{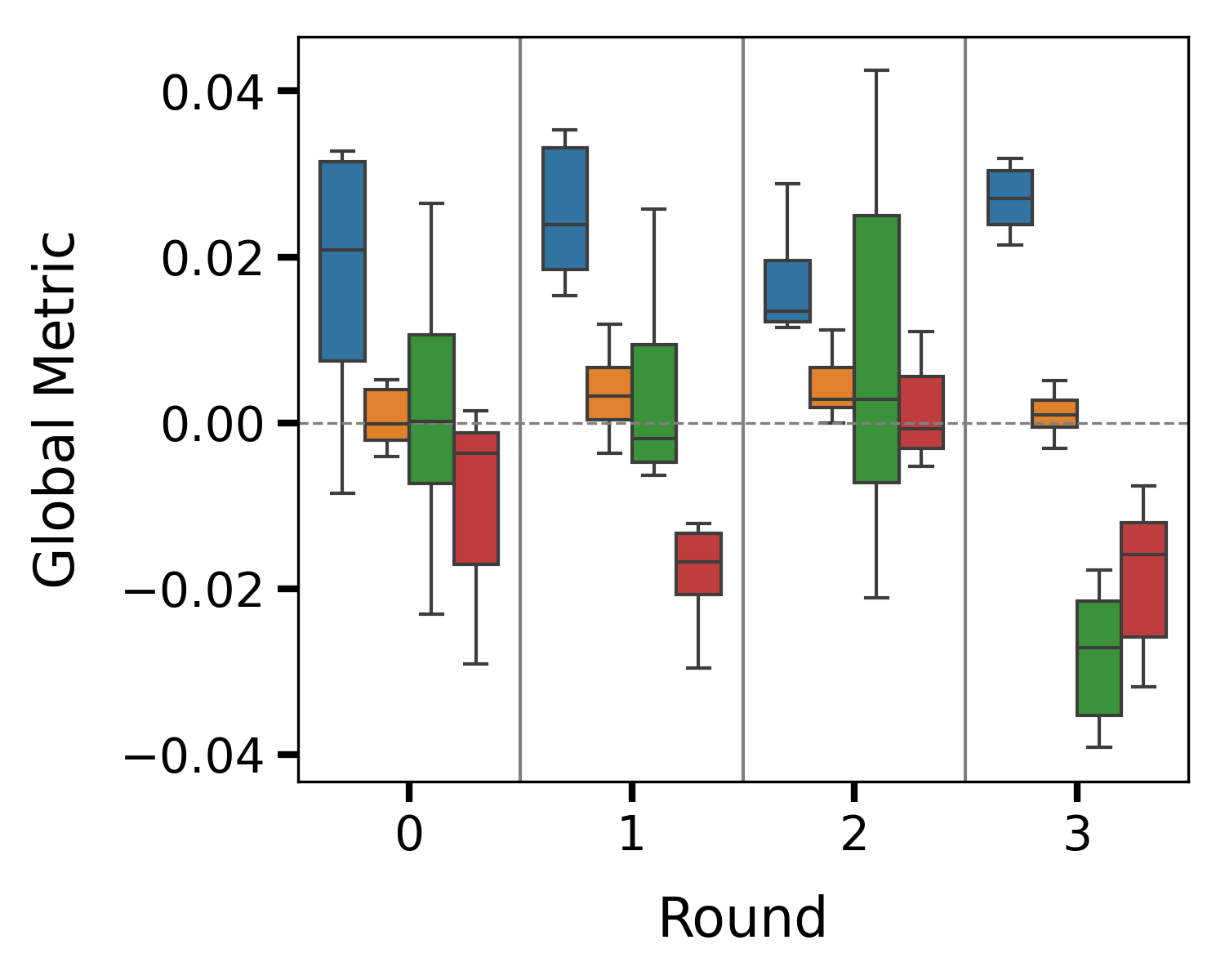}
        \caption{CIFAR / LOO.}
    \end{subfigure}
    \hfill
    \begin{subfigure}{0.22\textwidth}
        \includegraphics[width=\linewidth]{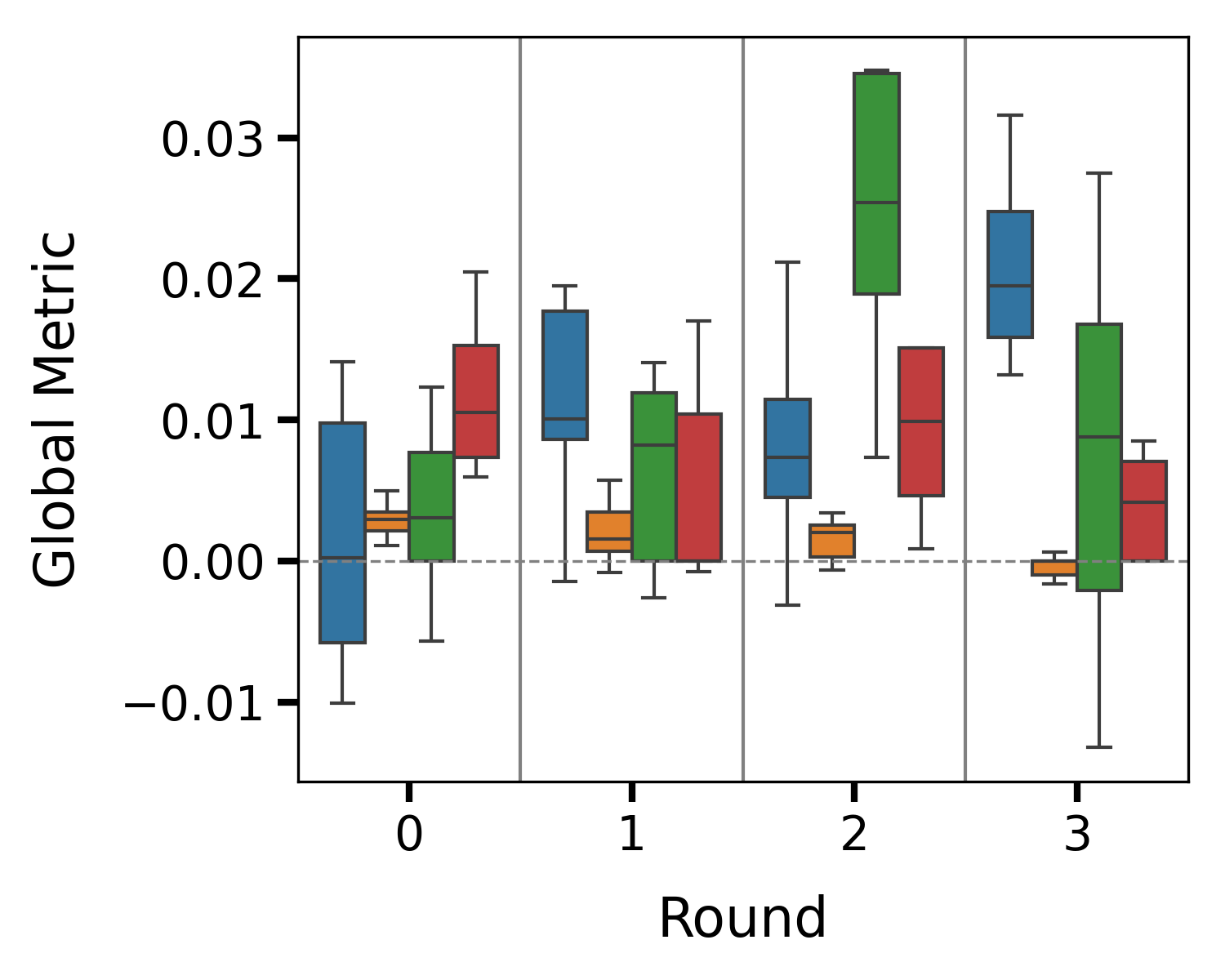}
        \caption{CIFAR / GTG.}
    \end{subfigure}

    \begin{subfigure}{0.22\textwidth}
        \includegraphics[width=\linewidth]{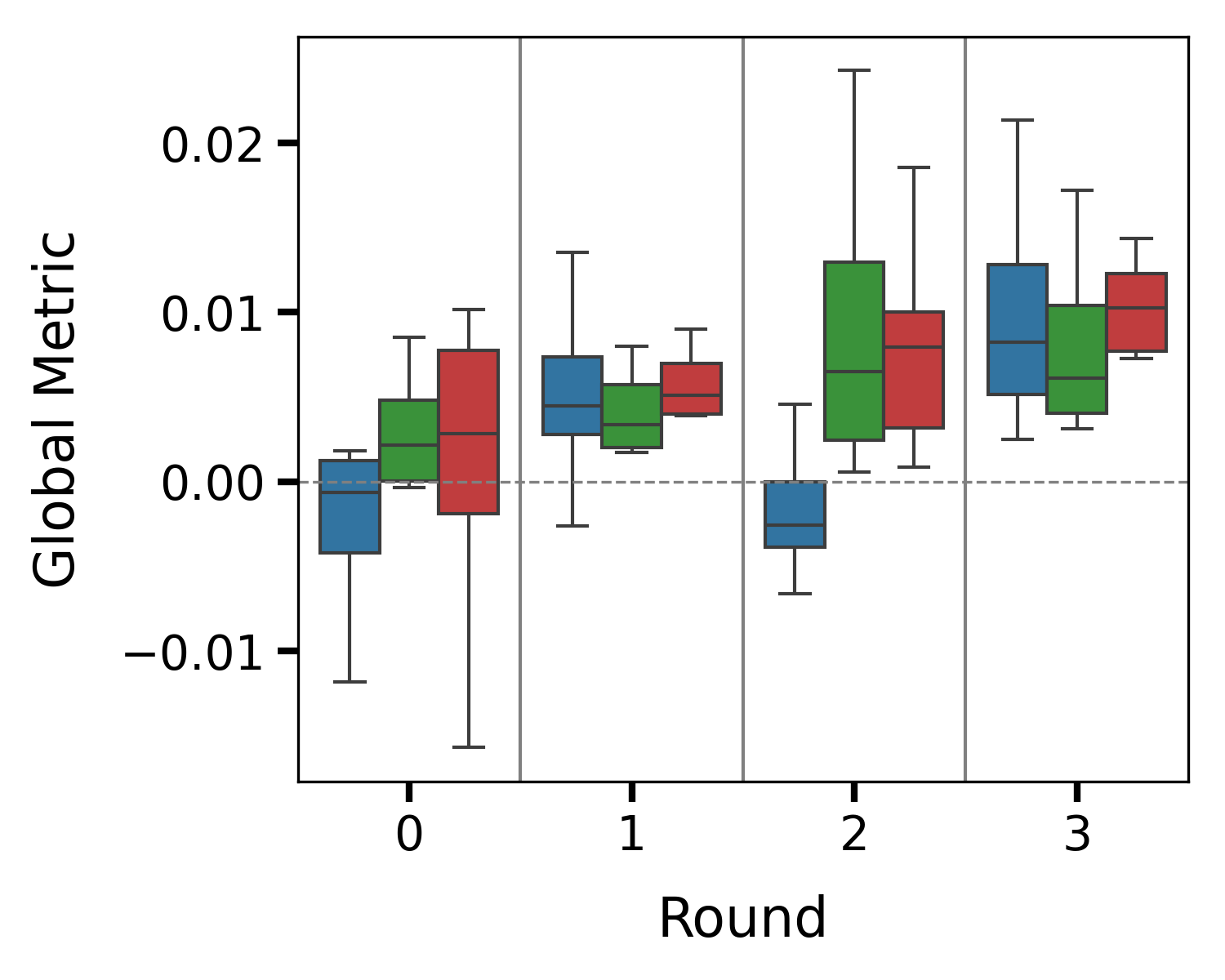}
        \caption{IMDB / LOO.}
    \end{subfigure}
    \hfill
    \begin{subfigure}{0.22\textwidth}
        \includegraphics[width=\linewidth]{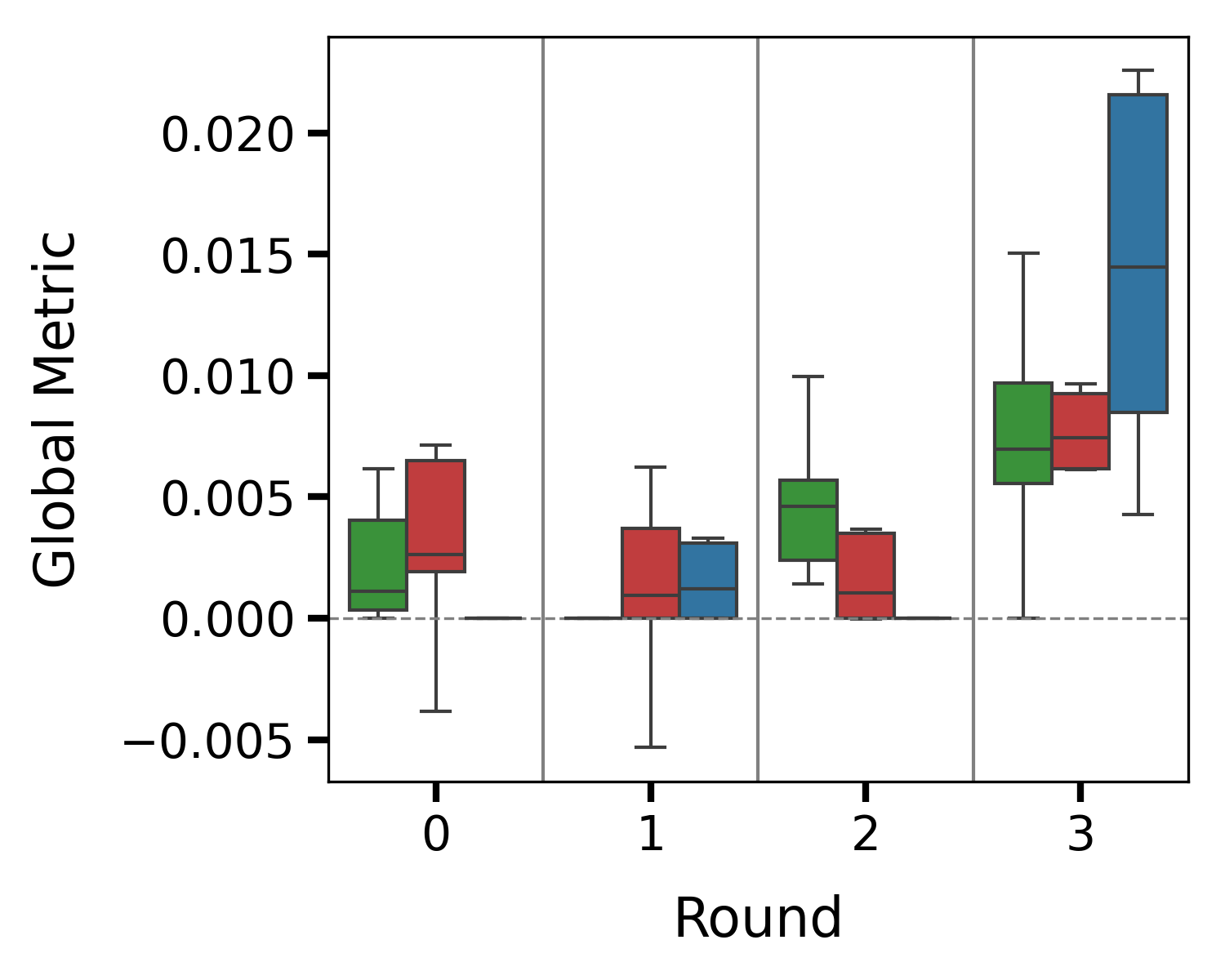}
        \caption{IMDB / GTG.}
    \end{subfigure}
    \caption{Computed scores for the 4 non-IID clients.}
    \label{fig:score_4_NIID}
\end{figure}

Next, based on these model scores, we allocate trustworthiness scores to the clients. While the model behaves in one way, that is the aggregated result of the client's contributions, which could drastically differ from each other. For space constrains, we only show in Fig.~\ref{fig:score_4_NIID} the obtained scores for the 4 client non-IID scenario, as with IID the scores are closer (as expected). The results corresponding to the other settings are placed in the Appendix: Fig.~\ref{fig:score_4_IID} and Fig.~\ref{fig:score_20}.

Some consistent pattern emerges across all three datasets and contribution evaluation scheme, namely that most scores have high variance and they are mostly positive. Several dominant clients appear that substantially improve the global model according to some metrics while not excelling (or even being harmful) in another. This is where the trade-offs between performance, fairness, reliability, and resilience are the most visible. For instance, some clients improve fairness or protection against attacks at the expense of accuracy, demonstrating that trustworthy behavior cannot be achieved by optimizing for global accuracy alone. 

Contribution analysis also reveals that trustworthiness in federated systems is inherently client-dependent: some clients consistently have positive, others consistently negative effects on a particular aspect, thus provides not only diagnostic insight, but also a basis for governance mechanisms that steer federated AI systems toward  operation. For further details on the various score evolutions across training rounds, we refer the reader to Appendix~\ref{app:evol}. 

\paragraph{Trustworthy Score Distributions. }

\begin{figure}[!b]
    \centering
    \begin{subfigure}{0.45\textwidth}
        \includegraphics[width=\linewidth]{figs/S_legend.png}
    \end{subfigure}
    
    \begin{subfigure}{0.22\textwidth}
        \includegraphics[width=\linewidth]{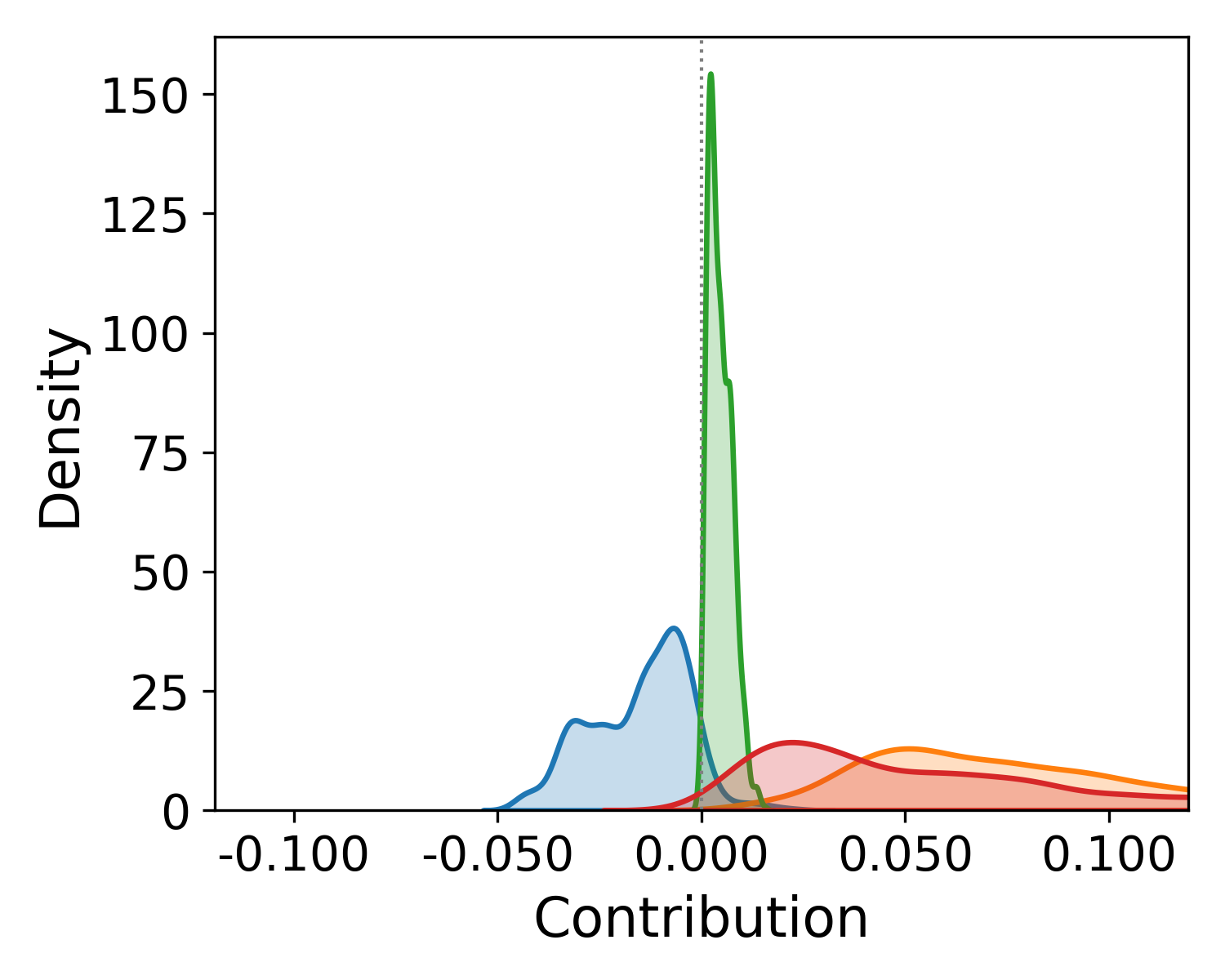}
        \caption{ADULT / LOO.}
    \end{subfigure}
    \hfill
    \begin{subfigure}{0.22\textwidth}
        \includegraphics[width=\linewidth]{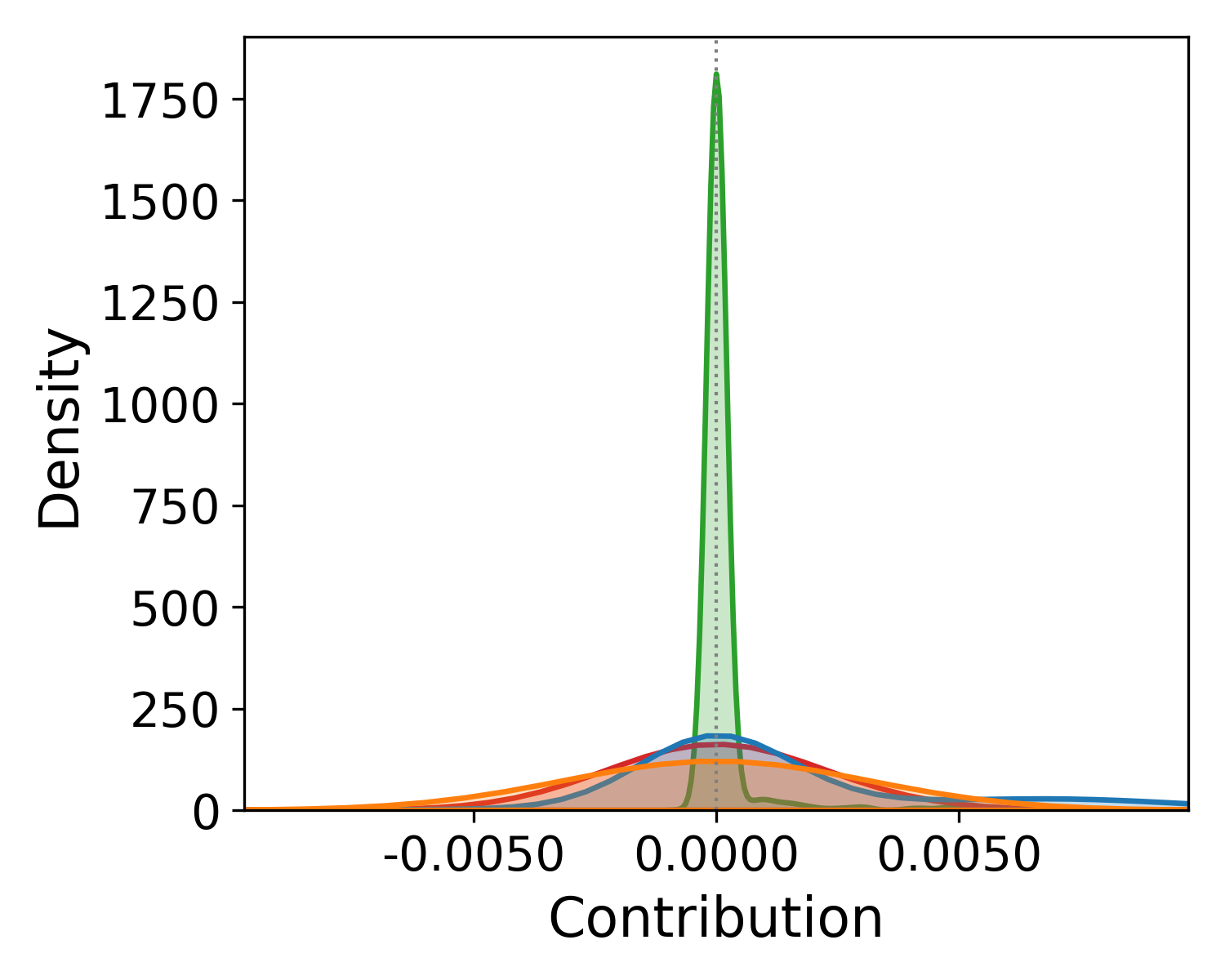}
        \caption{ADULT / GTG.}
    \end{subfigure}

    \begin{subfigure}{0.22\textwidth}
        \includegraphics[width=\linewidth]{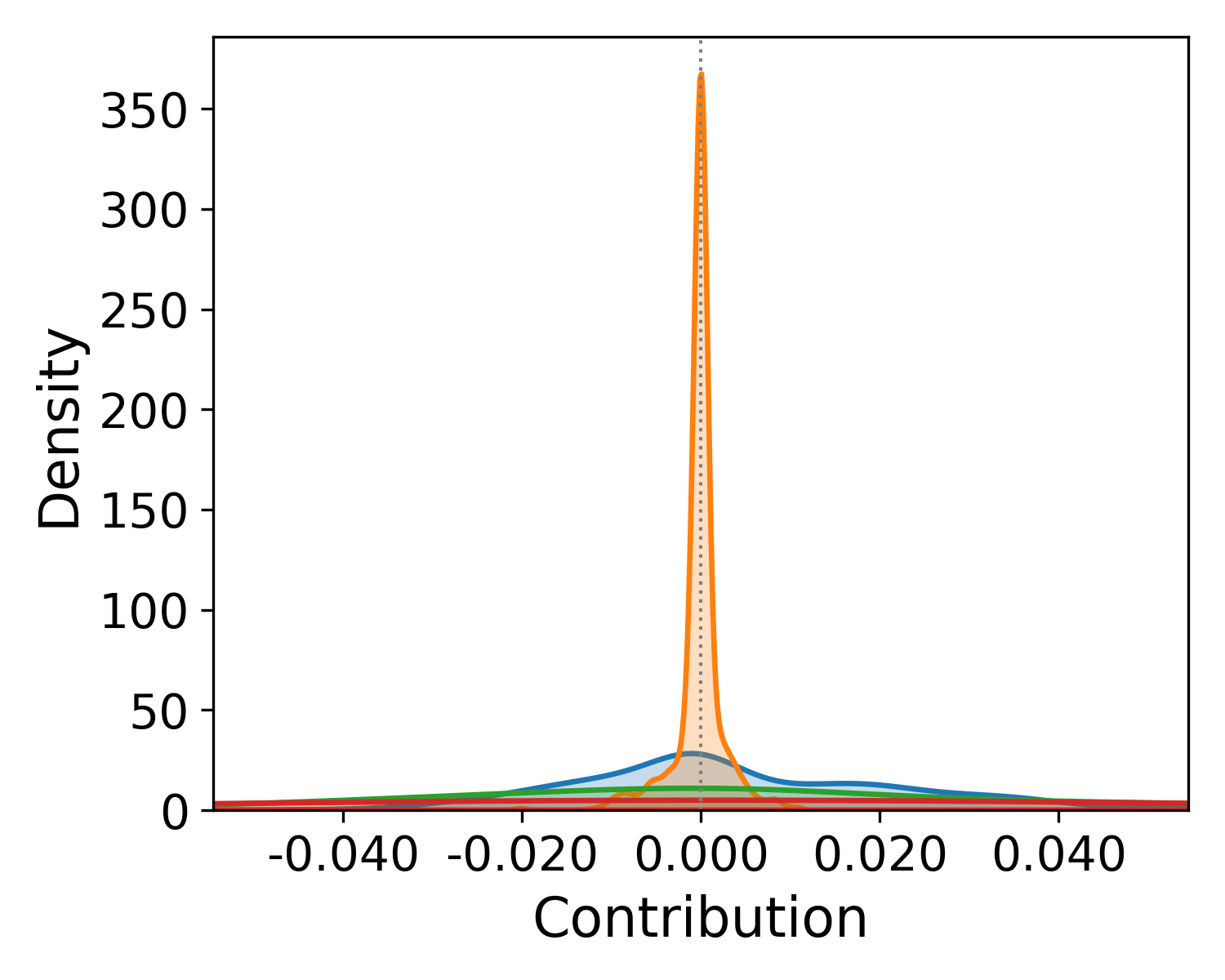}
        \caption{CIFAR / LOO.}
    \end{subfigure}
    \hfill
    \begin{subfigure}{0.22\textwidth}
        \includegraphics[width=\linewidth]{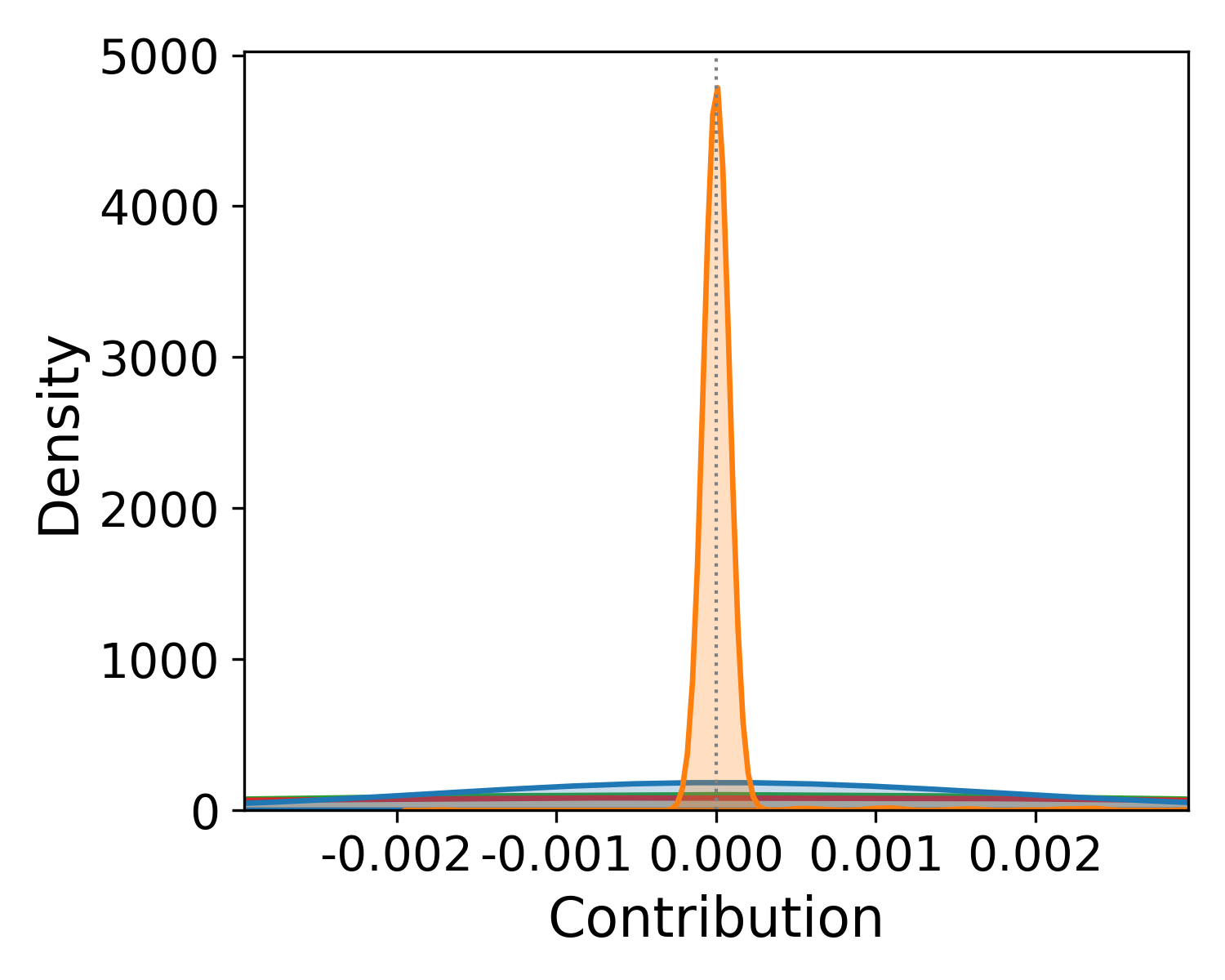}
        \caption{CIFAR / GTG.}
    \end{subfigure}

   \begin{subfigure}{0.22\textwidth}
        \includegraphics[width=\linewidth]{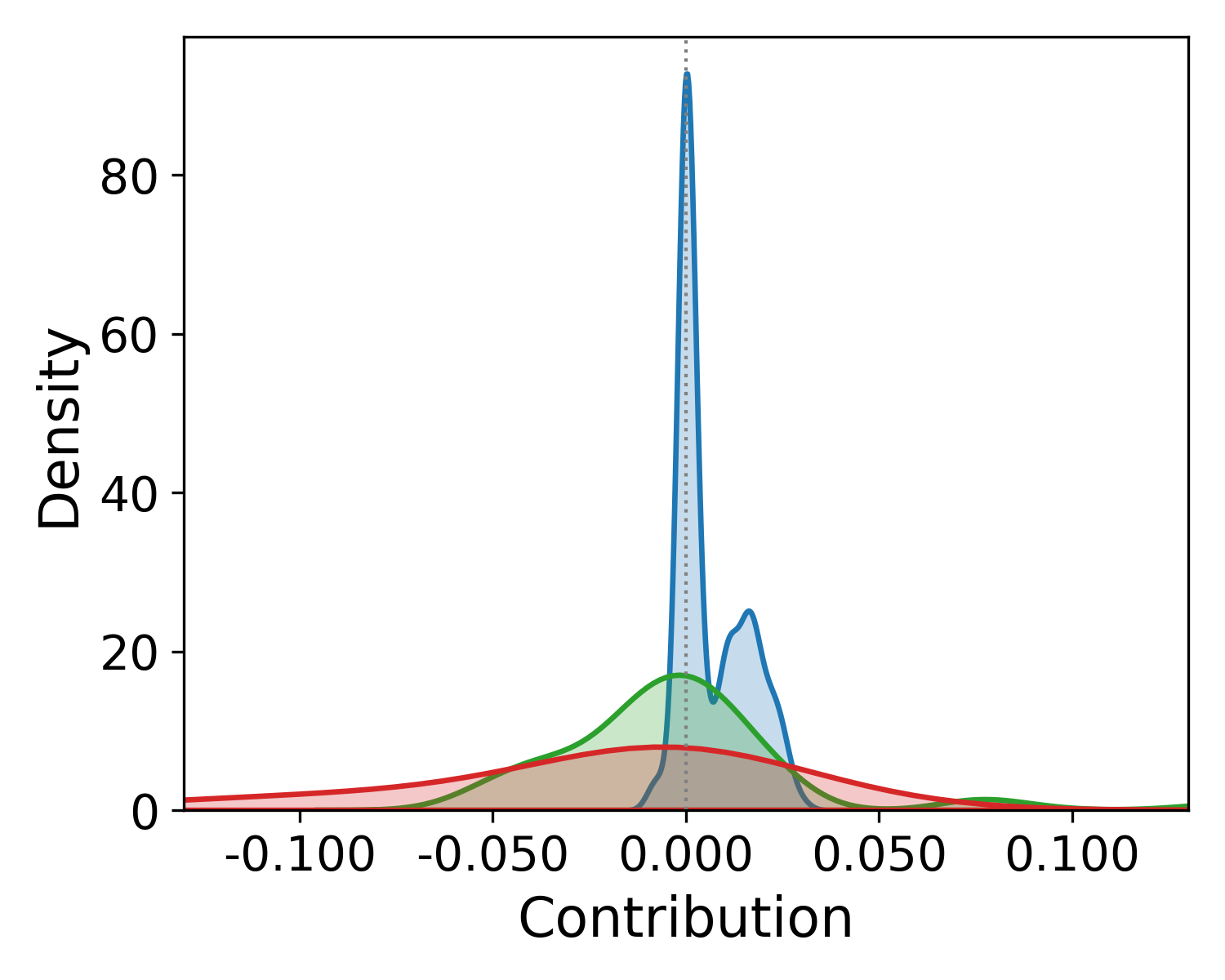}
        \caption{IMDB / LOO.}
    \end{subfigure}
    \hfill
    \begin{subfigure}{0.22\textwidth}
        \includegraphics[width=\linewidth]{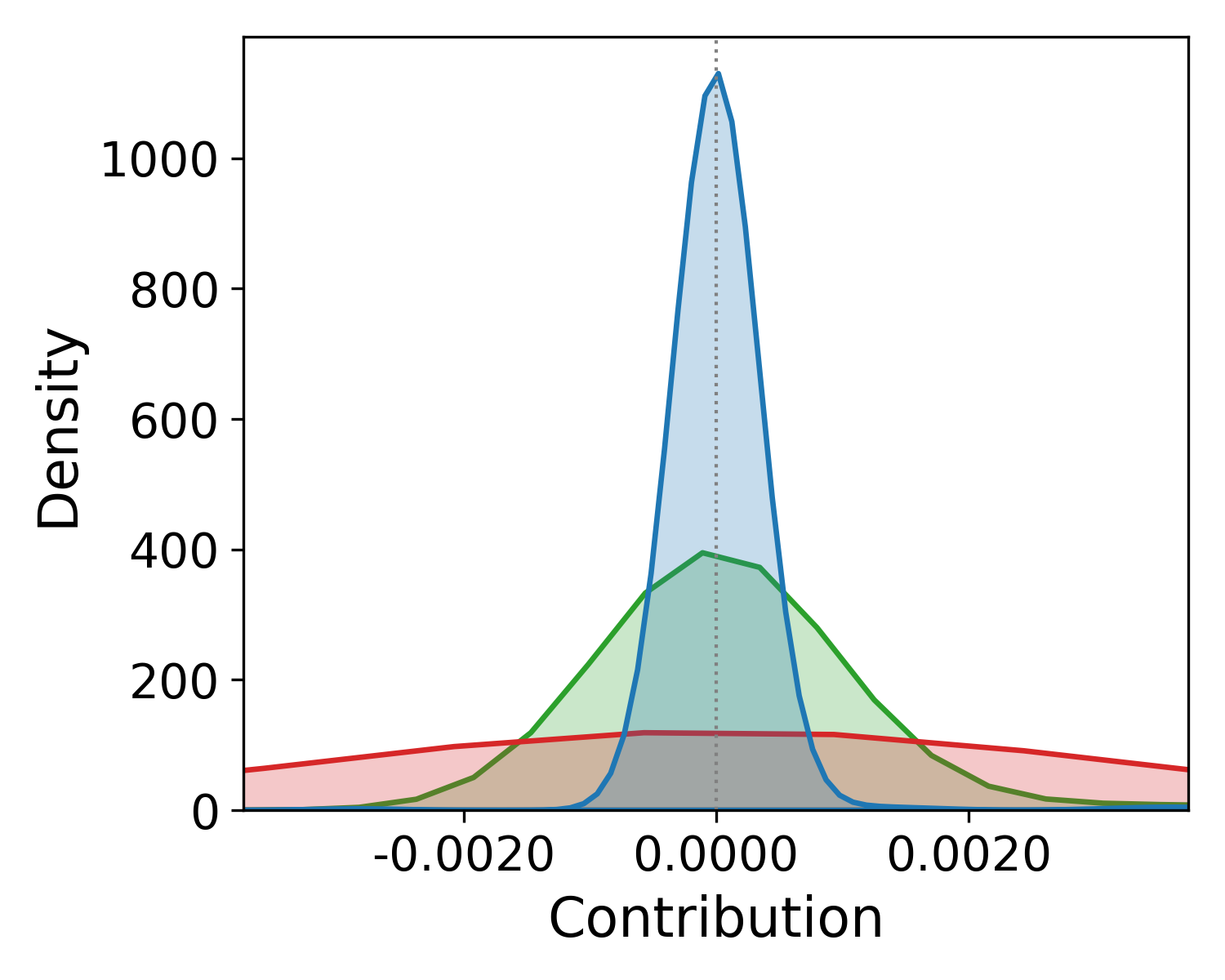}
        \caption{IMDB / GTG.}
    \end{subfigure}    
    \caption{Score distributions of the 20 non-IID clients.}
    \label{fig:score_dist_20_NIID}
\end{figure}

Instead of focusing on individuals, in Fig.~\ref{fig:score_dist_20_NIID} we show the trustworthiness score distributions in the 20 non-IID client setting, as for IID the trustworthy score distribution are 'bell' curves with different variances centered around 0 (as expected). This and the 4 client results can be find in the Appendix Fig.~\ref{fig:score_dist_20_IID} and Fig.~\ref{fig:score_dist_4}, respectively. 

The presented non-IID results reveal a deviations from the symmetric, zero-centered distributions observed in the IID case. This is most profound with $\mathtt{LOO}$ regarding ADULT and IMDB: distinct negative and positive shifts are present regarding $\mathtt{perf}$, $\mathtt{rel}$, and $\mathtt{res}$. In contrast, $\mathtt{GTG}$ based scores are more conservative: they fall within a smaller numerical scale and free of such severe shifts.

Furthermore, we find that the underlying scoring mechanism does matter, as $\mathtt{GTG}$ and $\mathtt{LOO}$ do not always align, showing inconsistencies even when utilizing the same evaluation function. Finally, the concentration of scores is highly dataset-dependent: $\mathtt{rel}$, $\mathtt{res}$, and $\mathtt{perf}$ are concentrated for ADULT, CIFAR, and  IMDB, respectively. This suggests that different data modalities are inherently more sensitive to specific trustworthiness dimensions, which in turn dictates how individual client contributions are perceived across the federation.

\paragraph{Trustworthy Score Differences. }

We also study the difference of the trustworthiness scores from the baseline performance score. For this we use an absolute and a relative metric. We apply the Spearman correlation coefficient ($\Phi$)~\cite{wissler1905spearman} to capture the discrepancies in the client's orderings that the scores imply. This value is in the range of $[-1,1]$, where positive/negative values imply a positive/negative correlation which is stronger if the value is further from 0. While such correlations are important for contribution evaluation, they are unable to capture the full picture: $[1,2,3]$ is the same order-wise as $[1,2,100]$. Thus, we employ the $L_2$ difference (RMSE) as well to capture how far the client's various scores are. This is especially important when any monetary gain or reward is allocated between the clients. Both values are reported in Tab.~\ref{tab:score_diff_GTG} when $\mathtt{GTG}$ is used, in the Appendix in Tab.~\ref{tab:score_diff_LOO} we show the result corresponding to $\mathtt{LOO}$. 

\begin{table}[!t]
    \centering
    \setlength{\tabcolsep}{2pt}
    \begin{tabular}{ccc||cc|cc|cc}
        \multicolumn{3}{c||}{\multirow{2}{*}{\textbf{Setting}}}
        & \multicolumn{2}{c|}{\textbf{$\mathtt{fair}$}}
        & \multicolumn{2}{c|}{\textbf{$\mathtt{rel}$}}
        & \multicolumn{2}{c}{\textbf{$\mathtt{res}$}} \\
        &&& $\Phi$ & $L_2$ & $\Phi$ & $L_2$ & $\Phi$ & $L_2$ \\
        \hline\hline
        \multirow{4}{*}{\rotatebox{90}{ADULT}} & \multirow{2}{*}{4} & I &
        -0.145 & 0.029 & 0.056 & 0.026 & -0.021 & 0.025 \\
        && N &
        -0.162 & 0.053 & -0.152 & 0.036 & -0.127 & 0.050 \\
        \cline{2-9}
        & \multirow{2}{*}{20} & I &
        -0.000 & 0.004 & 0.080 & 0.003 & -0.021 & 0.004 \\
        && N &
        -0.040 & 0.014 & -0.042 & 0.006 & -0.036 & 0.011 \\
        \hline
        \multirow{4}{*}{\rotatebox{90}{CIFAR}} & \multirow{2}{*}{4} & I &
        -0.007 & 0.037 & -0.166 & 0.048 & -0.250 & 0.058 \\
        && N &
        -0.096 & 0.040 & -0.320 & 0.075 & -0.386 & 0.095 \\
        \cline{2-9}
        & \multirow{2}{*}{20} & I &
        0.004 & 0.004 & -0.043 & 0.011 & -0.074 & 0.018 \\
        && N &
        0.009 & 0.007 & -0.043 & 0.016 & -0.031 & 0.020 \\
        \hline
        \multirow{4}{*}{\rotatebox{90}{IMDB}} & \multirow{2}{*}{4} & I &
        \textemdash & \textemdash & 0.110 & 0.014 & 0.087 & 0.026 \\
        && N &
        \textemdash & \textemdash & 0.306 & 0.023 & 0.237 & 0.066 \\
        \cline{2-9}
        & \multirow{2}{*}{20} & I &
        \textemdash & \textemdash & 0.064 & 0.001 & -0.016 & 0.003 \\
        && N &
        \textemdash & \textemdash & -0.016 & 0.004 & -0.010 & 0.013 \\
    \end{tabular}
    \caption{Comparing the $\mathtt{GTG}$-based performance score to the trustworthy scores where I and N stands for IID and non-IID, respectively.}
    \label{tab:score_diff_GTG}
\end{table}

The key finding of this paper is presented in this table: with a few exceptions, the absolute value of the $\Phi$ correlation coefficients are below $0.1$, which imply these trustworthiness scores are largely independent from the performance-based score. Thus, for collaborative systems with trustworthy emphasis, it is inadequate to evaluate the participants solely using a single metric, and it is imperative to consider all the dimensions the consortium cares about. 

\paragraph{Trustworthy Score Correlations. }

Lastly, instead of comparing the trustworthy scores to the performance score, we are comparing them against each other. This is visualized as a heatmap using the Spearman correlations in Fig.~\ref{fig:score_heat_4_N} for 4 non-IID clients. Similar results are reported in the Appendix for the other studied cases (Fig.~\ref{fig:score_heat_4_I} and Fig.~\ref{fig:score_heat_20}).

\begin{figure}[!t]
    \centering    
    \begin{subfigure}{0.22\textwidth}
        \includegraphics[width=\linewidth]{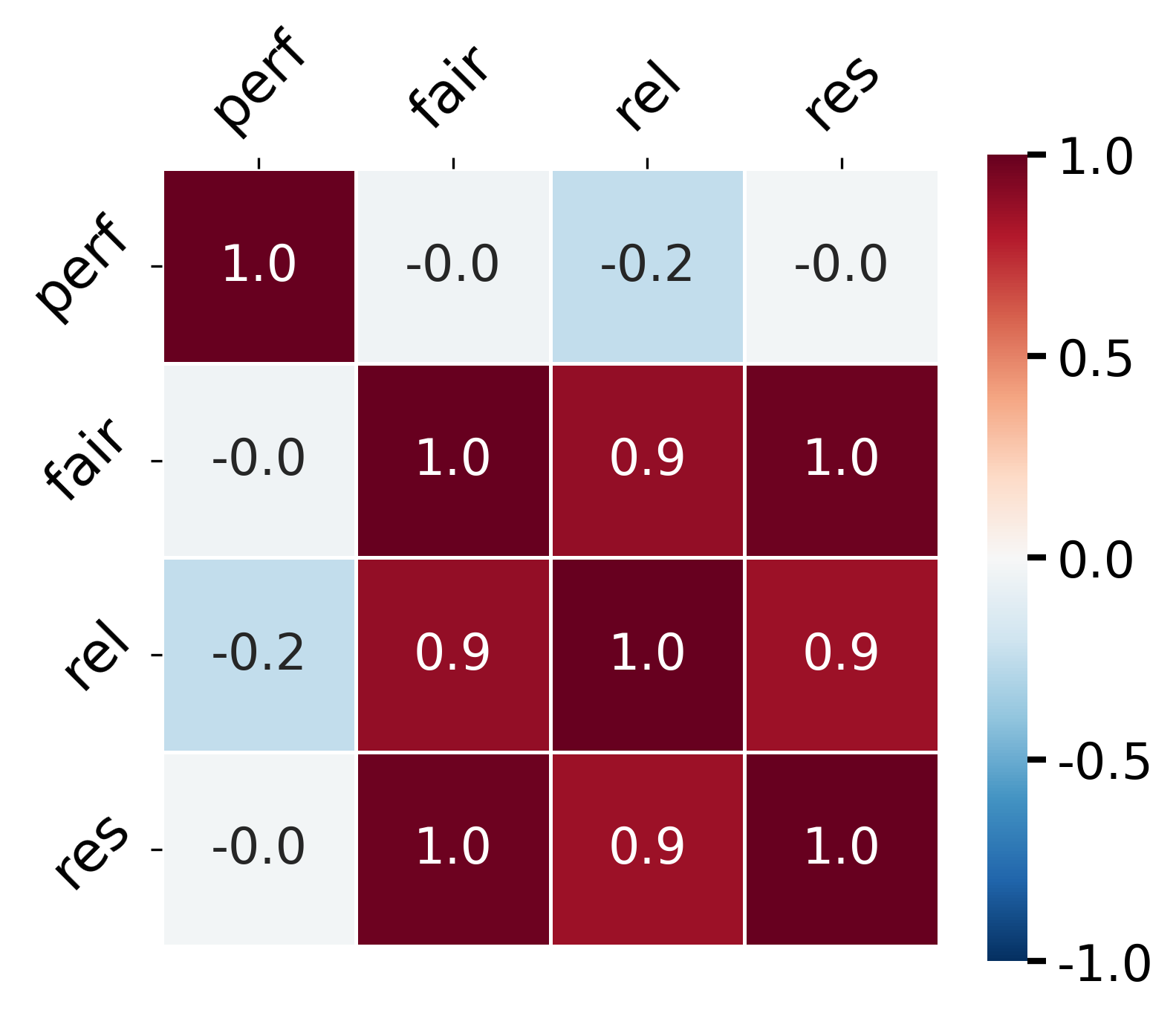}
        \caption{ADULT / LOO.}
    \end{subfigure}
    \hfill
    \begin{subfigure}{0.22\textwidth}
        \includegraphics[width=\linewidth]{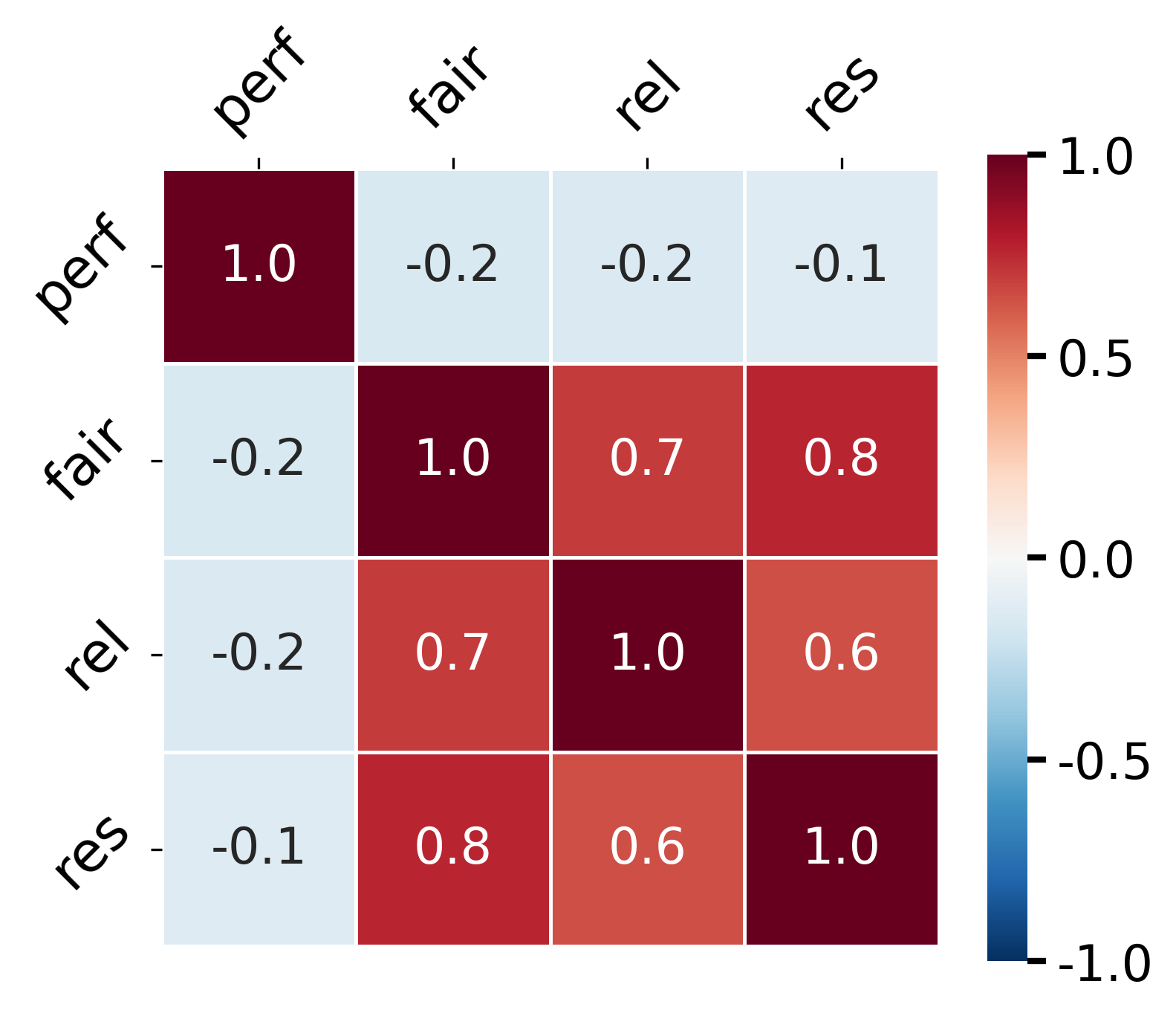}
        \caption{ADULT / GTG.}
    \end{subfigure}

    \begin{subfigure}{0.22\textwidth}
        \includegraphics[width=\linewidth]{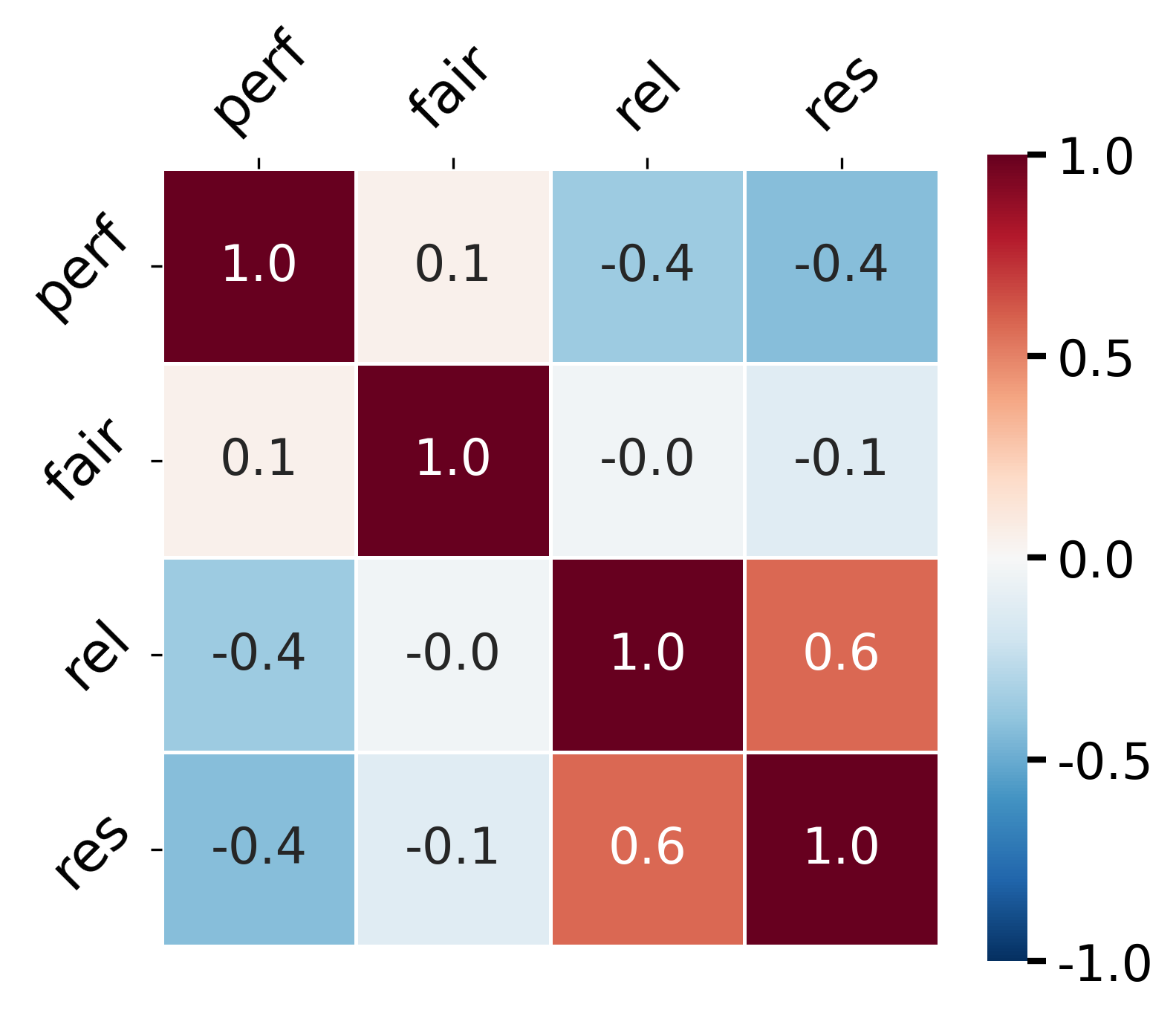}
        \caption{CIFAR / LOO.}
    \end{subfigure}
    \hfill
    \begin{subfigure}{0.22\textwidth}
        \includegraphics[width=\linewidth]{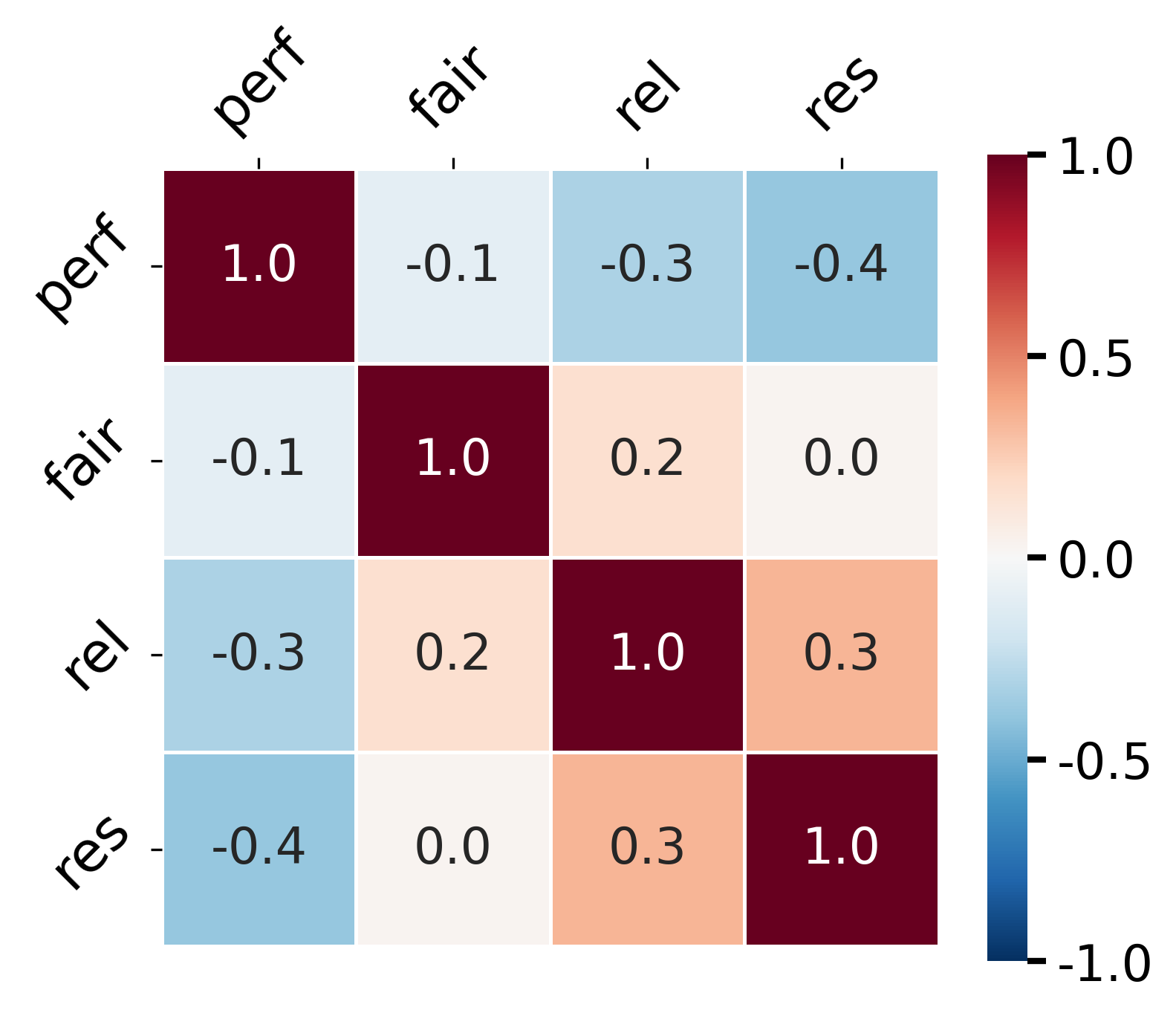}
        \caption{CIFAR / GTG.}
    \end{subfigure}

    \begin{subfigure}{0.22\textwidth}
        \includegraphics[width=\linewidth]{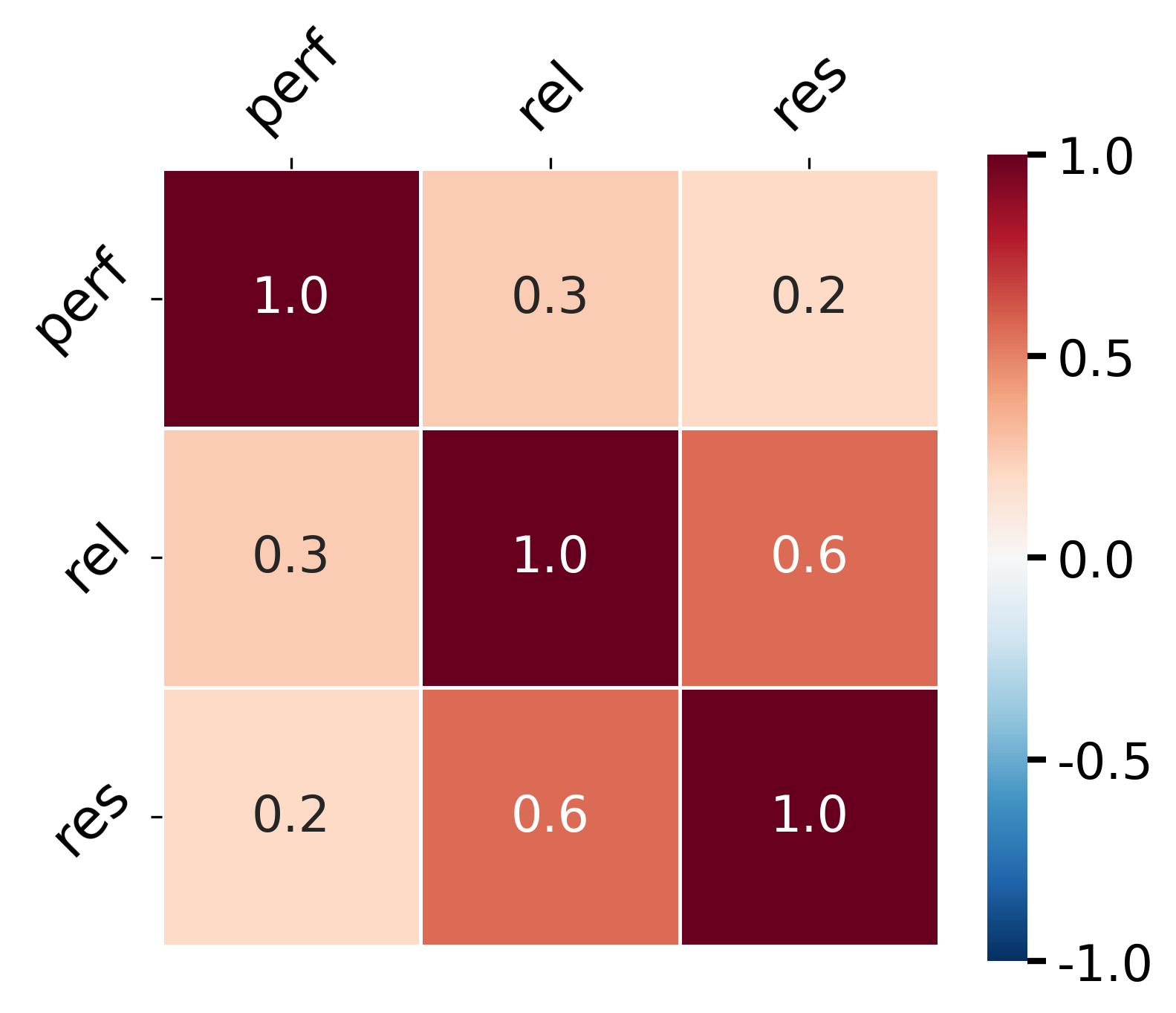}
        \caption{IMDB / LOO.}
    \end{subfigure}
    \hfill
    \begin{subfigure}{0.22\textwidth}
        \includegraphics[width=\linewidth]{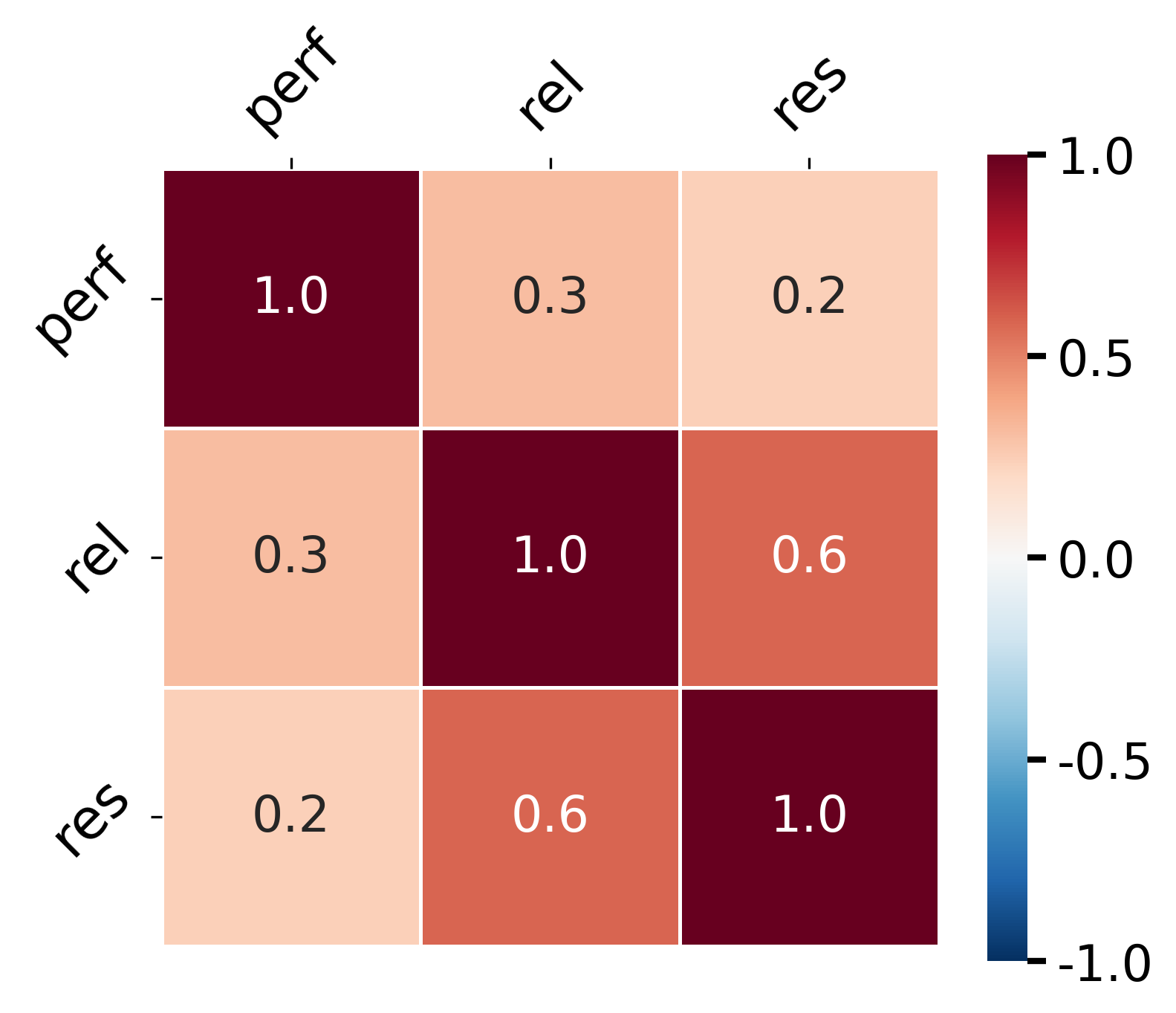}
        \caption{IMDB / GTG.}
    \end{subfigure}
    \caption{Heatmap of the pair-wise score correlations for the 4 non-IID client setting.  .}
    \label{fig:score_heat_4_N}
\end{figure}

The pairwise correlation analysis confirms that performance-based evaluations are poor proxies for the client's trustworthiness. Across all datasets, correlations between $\mathtt{perf}$ and the three trustworthiness metrics are consistently low, even trending in negative directions (CIFAR). In contrast, trustworthiness dimensions exhibit significantly stronger inter-correlations; notably, $\mathtt{rel}$ and $\mathtt{res}$ consistently correlate above $0.5$, reflecting their shared focus on model robustness. Another finding regarding the ADULT dataset is the high correlation of $\mathtt{fair}$ with both robustness metrics, suggesting a link between equitable data representation and resistance to perturbations.

\section{Conclusion}
\label{sec:con}

Our investigation demonstrates that evaluating clients in Federated Learning solely through performance metrics provides an incomplete, and often misleading, picture of their true contribution. By extending client evaluation to encompass the pillars of trustworthiness (i.e., fairness and robustness), we reveal that a client's value is highly context-dependent. Our empirical results show that performance-wise 'good' clients are not necessarily trustworthy; in fact, the rankings of participants shift dramatically depending on the chosen metric. These findings have profound implications for incentive design: a reward system based only on accuracy may inadvertently penalize clients who are essential for ensuring the model's trustworthiness. Future work should explore how these multi-dimensional scores can be integrated into a unified, holistic reward allocation scheme that prioritizes the deployment of truly dependable and trustworthy AI systems.

\paragraph{Limitations \& Future Works. }

It is important to note that our choice of any single metric is one of many possibilities. We hypothesize that replacing these underlying definitions --- using another fairness definition, a different noise distribution, another adversarial attack --- would likely result in subtle but meaningful differences in the resulting specific client scores. A comprehensive investigation into the sensitivity of client rankings to different metric choices within a single aspect (e.g., accuracy vs loss for performance) is a promising avenue for future work. Further trustworthy AI dimensions could also be considered, e.g., measuring the privacy of the models (and consequently assigning a privacy score to the clients) via the success rate of a membership inference attack. Another avenue to explore is the measurements of such aspects when they are explicitly considered during training (just as the model performance), i.e., applying robust training, using a fairness regularization, or enforcing differential privacy, as these techniques would directly impact the corresponding contribution scores.

\begin{acknowledgements}
    Project no. 145832, implemented with the support provided by the Ministry of Innovation and Technology from the NRDI Fund, financed under the PD\_23 funding scheme and by the European Union project RRF-2.3.1-21-2022-00004 within the framework of the Artificial Intelligence National Laboratory.
\end{acknowledgements}

\newpage
\bibliography{ref}

\begin{thebibliography}{47}
\providecommand{\natexlab}[1]{#1}
\providecommand{\url}[1]{\texttt{#1}}
\expandafter\ifx\csname urlstyle\endcsname\relax
  \providecommand{\doi}[1]{doi: #1}\else
  \providecommand{\doi}{doi: \begingroup \urlstyle{rm}\Url}\fi

\bibitem[Arnaiz-Rodriguez and Oliver(2023)]{arnaiztowards}
Adrian Arnaiz-Rodriguez and Nuria Oliver.
\newblock Towards algorithmic fairness by means of instance-level data
  re-weighting based on shapley values.
\newblock \emph{arXiv preprint arXiv:2303.01928}, 2023.

\bibitem[Calders et~al.(2009)Calders, Kamiran, and
  Pechenizkiy]{calders2009building}
Toon Calders, Faisal Kamiran, and Mykola Pechenizkiy.
\newblock Building classifiers with independency constraints.
\newblock In \emph{2009 IEEE international conference on data mining
  workshops}, 2009.

\bibitem[Cera et~al.(2018)Cera, Yanga, Konga, Huaa, Limtiacob, Johna,
  Constanta, Guajardo-C{\'e}spedesa, Yuanc, Tara, et~al.]{use}
Daniel Cera, Yinfei Yanga, Sheng-yi Konga, Nan Huaa, Nicole Limtiacob,
  Rhomni~St Johna, Noah Constanta, Mario Guajardo-C{\'e}spedesa, Steve Yuanc,
  Chris Tara, et~al.
\newblock Universal sentence encoder.
\newblock 2018.

\bibitem[Chkirbene et~al.(2024)Chkirbene, Hamila, Gouissem, and
  Devrim]{chkirbene2024large}
Zina Chkirbene, Ridha Hamila, Ala Gouissem, and Unal Devrim.
\newblock Large language models (llm) in industry: A survey of applications,
  challenges, and trends.
\newblock In \emph{2024 IEEE 21st International Conference on Smart
  Communities: Improving Quality of Life using AI, Robotics and IoT (HONET)},
  2024.

\bibitem[Evgeniou et~al.(2004)Evgeniou, Pontil, and
  Elisseeff]{evgeniou2004leave}
Theodoros Evgeniou, Massimiliano Pontil, and Andr{\'e} Elisseeff.
\newblock Leave one out error, stability, and generalization of voting
  combinations of classifiers.
\newblock \emph{Machine learning}, 2004.

\bibitem[Fawzi et~al.(2016)Fawzi, Moosavi-Dezfooli, and
  Frossard]{fawzi2016robustness}
Alhussein Fawzi, Seyed-Mohsen Moosavi-Dezfooli, and Pascal Frossard.
\newblock Robustness of classifiers: from adversarial to random noise.
\newblock \emph{Advances in neural information processing systems}, 2016.

\bibitem[Ghorbani and Zou(2019)]{ghorbani2019data}
Amirata Ghorbani and James Zou.
\newblock Data shapley: Equitable valuation of data for machine learning.
\newblock In \emph{International conference on machine learning}, 2019.

\bibitem[Goldblum et~al.(2022)Goldblum, Tsipras, Xie, Chen, Schwarzschild,
  Song, Madry, Li, and Goldstein]{goldblum2022dataset}
Micah Goldblum, Dimitris Tsipras, Chulin Xie, Xinyun Chen, Avi Schwarzschild,
  Dawn Song, Aleksander Madry, Bo~Li, and Tom Goldstein.
\newblock Dataset security for machine learning: Data poisoning, backdoor
  attacks, and defenses.
\newblock \emph{IEEE Transactions on Pattern Analysis and Machine
  Intelligence}, 2022.

\bibitem[Goldwasser and Hooker(2024)]{goldwasser2024stabilizing}
Jeremy Goldwasser and Giles Hooker.
\newblock Stabilizing estimates of shapley values with control variates.
\newblock In \emph{World Conference on Explainable Artificial Intelligence},
  2024.

\bibitem[Goodfellow et~al.(2015)Goodfellow, Shlens, and
  Szegedy]{goodfellow2014explaining}
Ian~J Goodfellow, Jonathon Shlens, and Christian Szegedy.
\newblock Explaining and harnessing adversarial examples.
\newblock \emph{stats}, 2015.

\bibitem[Guan et~al.(2022)Guan, Tu, He, and Tao]{guan2022few}
Jiyang Guan, Zhuozhuo Tu, Ran He, and Dacheng Tao.
\newblock Few-shot backdoor defense using shapley estimation.
\newblock In \emph{Proceedings of the IEEE/CVF Conference on Computer Vision
  and Pattern Recognition}, 2022.

\bibitem[Gunning et~al.(2019)Gunning, Stefik, Choi, Miller, Stumpf, and
  Yang]{gunning2019xai}
David Gunning, Mark Stefik, Jaesik Choi, Timothy Miller, Simone Stumpf, and
  Guang-Zhong Yang.
\newblock Xai—explainable artificial intelligence.
\newblock \emph{Science robotics}, 2019.

\bibitem[Guo et~al.(2023)Guo, Bao, Wang, Ma, Gao, Xiao, Liu, Dong, Liu, and
  Wu]{23metric}
Jun Guo, Wei Bao, Jiakai Wang, Yuqing Ma, Xinghai Gao, Gang Xiao, Aishan Liu,
  Jian Dong, Xianglong Liu, and Wenjun Wu.
\newblock A comprehensive evaluation framework for deep model robustness.
\newblock \emph{Pattern Recognition}, 2023.

\bibitem[Hall et~al.(2022)Hall, van~der Maaten, Gustafson, Jones, and
  Adcock]{hall2022systematic}
Melissa Hall, Laurens van~der Maaten, Laura Gustafson, Maxwell Jones, and Aaron
  Adcock.
\newblock A systematic study of bias amplification.
\newblock \emph{arXiv preprint arXiv:2201.11706}, 2022.

\bibitem[Hendrycks and Dietterich(2019)]{mCE}
Dan Hendrycks and Thomas Dietterich.
\newblock Benchmarking neural network robustness to common corruptions and
  perturbations.
\newblock In \emph{International Conference on Learning Representations}, 2019.

\bibitem[Huang et~al.(2020)Huang, Talbi, Zhao, Boucchenak, Chen, and
  Roos]{huang2020exploratory}
Jiyue Huang, Rania Talbi, Zilong Zhao, Sara Boucchenak, Lydia~Y Chen, and
  Stefanie Roos.
\newblock An exploratory analysis on users’ contributions in federated
  learning.
\newblock In \emph{IEEE international conference on trust, privacy and security
  in intelligent systems and applications (TPS-ISA)}, 2020.

\bibitem[Ibitoye et~al.(2019)Ibitoye, Abou-Khamis, Shehaby, Matrawy, and
  Shafiq]{ibitoye2019threat}
Olakunle Ibitoye, Rana Abou-Khamis, Mohamed~el Shehaby, Ashraf Matrawy, and
  M~Omair Shafiq.
\newblock The threat of adversarial attacks on machine learning in network
  security--a survey.
\newblock \emph{arXiv preprint arXiv:1911.02621}, 2019.

\bibitem[Jain et~al.(2020)Jain, Patel, Nagalapatti, Gupta, Mehta, Guttula,
  Mujumdar, Afzal, Sharma~Mittal, and Munigala]{jain2020overview}
Abhinav Jain, Hima Patel, Lokesh Nagalapatti, Nitin Gupta, Sameep Mehta,
  Shanmukha Guttula, Shashank Mujumdar, Shazia Afzal, Ruhi Sharma~Mittal, and
  Vitobha Munigala.
\newblock Overview and importance of data quality for machine learning tasks.
\newblock In \emph{Proceedings of the 26th ACM SIGKDD international conference
  on knowledge discovery \& data mining}, 2020.

\bibitem[Jiang et~al.(2023)Jiang, Liang, Zou, and Kwon]{jiang2023opendataval}
Kevin Jiang, Weixin Liang, James~Y Zou, and Yongchan Kwon.
\newblock Opendataval: a unified benchmark for data valuation.
\newblock \emph{Advances in Neural Information Processing Systems}, 2023.

\bibitem[Kairouz et~al.(2021)Kairouz, McMahan, Avent, Bellet, Bennis, Bhagoji,
  Bonawitz, Charles, Cormode, Cummings, et~al.]{kairouz2021advances}
Peter Kairouz, H~Brendan McMahan, Brendan Avent, Aur{\'e}lien Bellet, Mehdi
  Bennis, Arjun~Nitin Bhagoji, Kallista Bonawitz, Zachary Charles, Graham
  Cormode, Rachel Cummings, et~al.
\newblock Advances and open problems in federated learning.
\newblock \emph{Foundations and trends{\textregistered} in machine learning},
  2021.

\bibitem[Kaur et~al.(2022)Kaur, Uslu, Rittichier, and
  Durresi]{kaur2022trustworthy}
Davinder Kaur, Suleyman Uslu, Kaley~J Rittichier, and Arjan Durresi.
\newblock Trustworthy artificial intelligence: a review.
\newblock \emph{ACM computing surveys (CSUR)}, 2022.

\bibitem[Kleinberg et~al.(2018)Kleinberg, Ludwig, Mullainathan, and
  Rambachan]{kleinberg2018algorithmic}
Jon Kleinberg, Jens Ludwig, Sendhil Mullainathan, and Ashesh Rambachan.
\newblock Algorithmic fairness.
\newblock In \emph{Aea papers and proceedings}, 2018.

\bibitem[Li and Li(2024)]{li2024triangular}
Jingyang Li and Guoqiang Li.
\newblock The triangular trade-off between robustness, accuracy and fairness in
  deep neural networks: A survey.
\newblock \emph{ACM Computing Surveys}, 2024.

\bibitem[Liu et~al.(2021)Liu, Ding, Shaham, Rahayu, Farokhi, and
  Lin]{liu2021machine}
Bo~Liu, Ming Ding, Sina Shaham, Wenny Rahayu, Farhad Farokhi, and Zihuai Lin.
\newblock When machine learning meets privacy: A survey and outlook.
\newblock \emph{ACM Computing Surveys (CSUR)}, 2021.

\bibitem[Liu et~al.(2020)Liu, Xie, Wang, Zou, Xiong, Ying, and
  Vasilakos]{liu2020privacy}
Ximeng Liu, Lehui Xie, Yaopeng Wang, Jian Zou, Jinbo Xiong, Zuobin Ying, and
  Athanasios~V Vasilakos.
\newblock Privacy and security issues in deep learning: A survey.
\newblock \emph{IEEE Access}, 2020.

\bibitem[Liu et~al.(2022)Liu, Chen, Yu, Liu, and Cui]{liu2022gtg}
Zelei Liu, Yuanyuan Chen, Han Yu, Yang Liu, and Lizhen Cui.
\newblock Gtg-shapley: Efficient and accurate participant contribution
  evaluation in federated learning.
\newblock \emph{ACM Transactions on intelligent Systems and Technology (TIST)},
  2022.

\bibitem[Madry et~al.(2017)Madry, Makelov, Schmidt, Tsipras, and
  Vladu]{madry2017towards}
Aleksander Madry, Aleksandar Makelov, Ludwig Schmidt, Dimitris Tsipras, and
  Adrian Vladu.
\newblock Towards deep learning models resistant to adversarial attacks.
\newblock \emph{arXiv preprint arXiv:1706.06083}, 2017.

\bibitem[McMahan et~al.(2017)McMahan, Moore, Ramage, Hampson, and
  y~Arcas]{mcmahan2017communication}
Brendan McMahan, Eider Moore, Daniel Ramage, Seth Hampson, and Blaise~Aguera
  y~Arcas.
\newblock Communication-efficient learning of deep networks from decentralized
  data.
\newblock In \emph{Artificial intelligence and statistics}, 2017.

\bibitem[Mehrabi et~al.(2021)Mehrabi, Morstatter, Saxena, Lerman, and
  Galstyan]{mehrabi2021survey}
Ninareh Mehrabi, Fred Morstatter, Nripsuta Saxena, Kristina Lerman, and Aram
  Galstyan.
\newblock A survey on bias and fairness in machine learning.
\newblock \emph{ACM computing surveys (CSUR)}, 2021.

\bibitem[Naidu et~al.(2023)Naidu, Zuva, and Sibanda]{naidu2023review}
Gireen Naidu, Tranos Zuva, and Elias~Mmbongeni Sibanda.
\newblock A review of evaluation metrics in machine learning algorithms.
\newblock In \emph{Computer science on-line conference}, 2023.

\bibitem[Otmani et~al.(2024)Otmani, El-Azouzi, and Labatut]{otmani2024fedsv}
Khaoula Otmani, Rachid El-Azouzi, and Vincent Labatut.
\newblock Fedsv: Byzantine-robust federated learning via shapley value.
\newblock In \emph{ICC 2024-IEEE International Conference on Communications},
  2024.

\bibitem[Pejo et~al.(2026)Pejo, Frank, Varga, and Veliczky]{pejo2025fragility}
Balazs Pejo, Marcell Frank, Krisztian Varga, and Peter Veliczky.
\newblock On the fragility of contribution score computation in federated
  learning.
\newblock \emph{IEEE International Conference on Secure and Trusthworthy
  Machine Learning}, 2026.

\bibitem[Qi and Zhu(2024)]{qi2024mechanism}
Meng Qi and Mingxi Zhu.
\newblock Mechanism for decision-aware collaborative federated learning: A
  pitfall of shapley values.
\newblock \emph{arXiv preprint arXiv:2403.04753}, 2024.

\bibitem[Rabin(1993)]{rabin1993incorporating}
Matthew Rabin.
\newblock Incorporating fairness into game theory and economics.
\newblock \emph{The American economic review}, 1993.

\bibitem[Rozemberczki et~al.(2022)Rozemberczki, Watson, Bayer, Yang, Kiss,
  Nilsson, and Sarkar]{rozemberczki2022shapley}
Benedek Rozemberczki, Lauren Watson, P{\'e}ter Bayer, Hao-Tsung Yang,
  Oliv{\'e}r Kiss, Sebastian Nilsson, and Rik Sarkar.
\newblock The shapley value in machine learning.
\newblock In \emph{International Joint/European Conference on Artificial
  Intelligence}, 2022.

\bibitem[Shaham et~al.(2023)Shaham, Hajisafi, Quan, Nguyen, Krishnamachari,
  Peris, Ghinita, Shahabi, and Pathirana]{shaham2023holistic}
Sina Shaham, Arash Hajisafi, Minh~K Quan, Dinh~C Nguyen, Bhaskar
  Krishnamachari, Charith Peris, Gabriel Ghinita, Cyrus Shahabi, and Pubudu~N
  Pathirana.
\newblock Holistic survey of privacy and fairness in machine learning.
\newblock \emph{arXiv preprint arXiv:2307.15838}, 2023.

\bibitem[Shannon(1948)]{shannon1948}
Claude~E. Shannon.
\newblock A mathematical theory of communication.
\newblock \emph{Bell System Technical Journal}, 1948.

\bibitem[Shapley and Corporation(1951)]{shapley1951notes}
L.S. Shapley and Rand Corporation.
\newblock Notes on the n-person game.
\newblock \emph{Notes on the N-person Game}, 1951.

\bibitem[Siomos and Passerat-Palmbach(2023)]{siomos2023contribution}
Vasilis Siomos and Jonathan Passerat-Palmbach.
\newblock Contribution evaluation in federated learning: Examining current
  approaches.
\newblock \emph{arXiv preprint arXiv:2311.09856}, 2023.

\bibitem[Taori et~al.(2020)Taori, Dave, Shankar, Carlini, Recht, and
  Schmidt]{effectiverelative}
Rohan Taori, Achal Dave, Vaishaal Shankar, Nicholas Carlini, Benjamin Recht,
  and Ludwig Schmidt.
\newblock Measuring robustness to natural distribution shifts in image
  classification.
\newblock \emph{Advances in Neural Information Processing Systems}, 2020.

\bibitem[Wachter et~al.(2020)Wachter, Mittelstadt, and
  Russell]{wachter2020bias}
Sandra Wachter, Brent Mittelstadt, and Chris Russell.
\newblock Bias preservation in machine learning: the legality of fairness
  metrics under eu non-discrimination law.
\newblock \emph{W. Va. L. Rev.}, 2020.

\bibitem[Wang and Jia(2023)]{wang2023data}
Jiachen~T Wang and Ruoxi Jia.
\newblock Data banzhaf: A robust data valuation framework for machine learning.
\newblock In \emph{International Conference on Artificial Intelligence and
  Statistics}, 2023.

\bibitem[Wang and Strong(1996)]{wang1996beyond}
Richard~Y Wang and Diane~M Strong.
\newblock Beyond accuracy: What data quality means to data consumers.
\newblock \emph{Journal of management information systems}, 1996.

\bibitem[Wang et~al.(2020)Wang, Qinami, Karakozis, Genova, Nair, Hata, and
  Russakovsky]{wang2020towards}
Zeyu Wang, Klint Qinami, Ioannis~Christos Karakozis, Kyle Genova, Prem Nair,
  Kenji Hata, and Olga Russakovsky.
\newblock Towards fairness in visual recognition: Effective strategies for bias
  mitigation.
\newblock In \emph{Proceedings of the IEEE/CVF conference on computer vision
  and pattern recognition}, 2020.

\bibitem[Winter(2002)]{winter2002shapley}
Eyal Winter.
\newblock The shapley value.
\newblock \emph{Handbook of game theory with economic applications}, 2002.

\bibitem[Wissler(1905)]{wissler1905spearman}
Clark Wissler.
\newblock The spearman correlation formula.
\newblock \emph{Science}, 1905.

\bibitem[Xu et~al.(2024)Xu, Jiang, Niu, Jia, Li, and Poovendran]{xu2024ace}
Zhangchen Xu, Fengqing Jiang, Luyao Niu, Jinyuan Jia, Bo~Li, and Radha
  Poovendran.
\newblock $\{$ACE$\}$: A model poisoning attack on contribution evaluation
  methods in federated learning.
\newblock In \emph{33rd USENIX Security Symposium (USENIX Security 24)}, 2024.

\end{thebibliography}
\newpage

\onecolumn

\title{Beyond performance-wise Contribution Evaluation in Federated Learning\\(Supplementary Material)}
\maketitle

\appendix

\section{Experiment Details}
\label{app:setup}

\begin{itemize}
    \item For Computer Vision we used CIFAR-10 along with a CNN consisting of six convolutional layers arranged in three blocks (2×64, 2×128, 2×256 filters) with batch normalization and dropout, followed by a global average pooling layer, a dense ReLU layer with 256 units and a 10-way softmax output, trained with the Adam optimizer (learning rate $1\times10^{-3}$), sparse categorical crossentropy loss, and $L_2$ regularization ($5\times10^{-4}$). 
    \item For Natural Language Processing we utilize the IMDB dataset and employ the Universal Sentence Encoder~\cite{use} to generate 512-dimensional embeddings for each review, which are then fed into a two-layer MLP with 256 and 128 hidden units, ReLU activations, batch normalization and dropout, and a final 2-way softmax output layer that is trained with Adam, sparse categorical crossentropy loss. 
    \item For Tabular Data we selected ADULT and applied a multilayer perceptron with three ReLU-activated dense layers with 256, 128 and 64 units, Layer Normalization, dropout, and a final sigmoid activation, trained with Adam. 
\end{itemize}

While we aim for performance-free metrics to measure the contributions, the underlying model should be accurate, as that is the point of the collaboration after all. Hence, we performed a hyperparameter optimization; the parameters are shown in Tab.~\ref{tab:hyper}.

\begin{table}[h]
    \centering
    \setlength{\tabcolsep}{2pt}
    \begin{tabular}{|l||c|c|c|}
        \hline
        \textbf{Property} & ADULT & \textbf{CIFAR-10} & IMDB \\
        \hline\hline
        Train / Test & 40K / 10K & 30K / 8K & 20K / 5K \\
        \hline
        Input shape & 108 features & (32, 32, 3) & 512 (USE) \\
        Model & MLP  & CNN  & NN  \\
        Act. func. & ReLU + sigmoid & ReLU + softmax & ReLU + softmax \\
        Optimizer & Adam & Adam & Adam \\
        LR / DO & $1\times10^{-3}$/0.1/0.05 & $1\times10^{-3}$/0.25/0.35/0.5 & $3\times10^{-5}$/0.5/0.5 \\
        $L_2$ Reg & $10^{-4}$ & $5\times10^{-4}$ & $10^{-4}$ \\
        Norm & Layer & Batch & Batch \\
        Loss & Binary CCE & Sparse CCE & Sparse CCE \\
        \hline
        Noise $\sigma$ & 0.1 & 0.1 & 0.1 \\
        \hline
        GTG metrics & \multicolumn{3}{c}{ $\epsilon_1=0.001$, $\epsilon_2=0.05$, $\epsilon_3=0.002$ } \\
        \hline
    \end{tabular}
    \caption{Datasets and model configurations used in the experiments.}
    \label{tab:hyper}
\end{table}

\section{Trustworthy Score Evolution}
\label{app:evol}

As the reported client scores are the accumulated values across the training rounds, we measure the round-wise deviations in each round separately in Tab.~\ref{tab:score_fluct}. Here the low variance suggests the scores are rather consistent across round; foretelling that it could be enough to measure the scores in a single round.   

The variance values show a clear pattern across datasets and settings. On ADULT, client contributions are almost time-invariant in every configuration: variances are effectively zero after thresholding, indicating that a single round is already representative of each client’s long-term effect on the model. In contrast, CIFAR exhibits noticeably higher variance, especially in the non-IID cases and for small client counts, which suggests that heterogeneous and high-dimensional image data leads to unstable client contributions over rounds. IMDB lies between these extremes: IID partitions again yield very small fluctuations, while Non-IID settings produce visible variance in robustness- and adversarial-related scores, reflecting the sensitivity of text models to distributional shifts in local corpora. Overall, the results indicate that temporal stability of client contributions is primarily driven by data modality and heterogeneity: IID and higher client counts lead to stable, interpretable contributions, whereas non-IID, low-client regimes introduce substantial instability that must be considered when analyzing robustness and fairness at client level.

\begin{table}[h]
    \centering
    \begin{tabular}{ccc||c|c|c|c}
        \multicolumn{3}{c||}{\textbf{Setting}}
        & \textbf{$\mathtt{perf}$}
        & \textbf{$\mathtt{fair}$}
        & \textbf{$\mathtt{rel}$}
        & \textbf{$\mathtt{res}$} \\
        \hline\hline

        \multirow{4}{*}{\rotatebox{90}{ADULT}} & \multirow{2}{*}{4}
            & IID
            & 9.9$e${-7}±0.0
            & 2.4$e${-5}±0.0
            & 3.0$e${-7}±0.0
            & 2.3$e${-5}±0.0 \\
        && non-IID
            & 1.1$e${-4}±0.0
            & 1.5$e${-4}±0.0
            & 2.7$e${-6}±0.0
            & 1.0$e${-4}±0.0 \\
        \cline{2-7}

        & \multirow{2}{*}{20}
            & IID
            & 2.8$e${-7}±0.0
            & 1.6$e${-6}±0.0
            & 0.0±0.0
            & 2.2$e${-6}±0.0 \\
        && non-IID
            & 2.0$e${-5}±0.0
            & 2.0$e${-4}±0.0
            & 1.4$e${-6}±0.0
            & 2.1$e${-4}±0.0 \\
        \hline

        \multirow{4}{*}{\rotatebox{90}{CIFAR}} & \multirow{2}{*}{4}
            & IID
            & 2.9$e${-4}±0.0
            & 5.4$e${-5}±0.0
            & 8.4$e${-4}±0.0
            & 8.8$e${-4}±0.0 \\
        && non-IID
            & 1.4$e${-3}±0.0
            & 8.9$e${-5}±0.0
            & 3.6$e${-3}±5.0$e${-5}
            & 6.5$e${-3}±1.1$e${-4} \\
        \cline{2-7}

        & \multirow{2}{*}{20}
            & IID
            & 6.1$e${-6}±0.0
            & 1.6$e${-6}±0.0
            & 6.6$e${-5}±0.0
            & 2.6$e${-4}±0.0 \\
        && non-IID
            & 2.4$e${-4}±0.0
            & 7.2$e${-6}±0.0
            & 2.2$e${-3}±0.0
            & 3.7$e${-2}±1.7$e${-3} \\
        \hline

        \multirow{4}{*}{\rotatebox{90}{IMDB}} & \multirow{2}{*}{4}
            & IID
            & 1.1$e${-6}±0.0
            & --
            & 1.5$e${-6}±0.0
            & 8.4$e${-6}±0.0 \\
        && non-IID
            & 9.5$e${-5}±0.0
            & --
            & 5.8$e${-4}±0.0
            & 4.0$e${-3}±3.6$e${-5} \\
        \cline{2-7}

        & \multirow{2}{*}{20}
            & IID
            & 0.0±0.0
            & --
            & 1.0$e${-6}±0.0
            & 1.4$e${-6}±0.0 \\
        && non-IID
            & 8.4$e${-5}±0.0
            & --
            & 1.8$e${-3}±0.0
            & 2.1$e${-2}±4.2$e${-4} \\
    \end{tabular}
    \caption{Scores fluctuations across training rounds.}
    \label{tab:score_fluct}
\end{table}

\section{Supplementary Results}

\begin{figure}[h]
\centering
    \begin{subfigure}{0.6\textwidth}
        \includegraphics[width=\linewidth]{figs/S_legend.png}
    \end{subfigure}
    
    \begin{subfigure}{0.3\textwidth}
        \includegraphics[width=\linewidth]{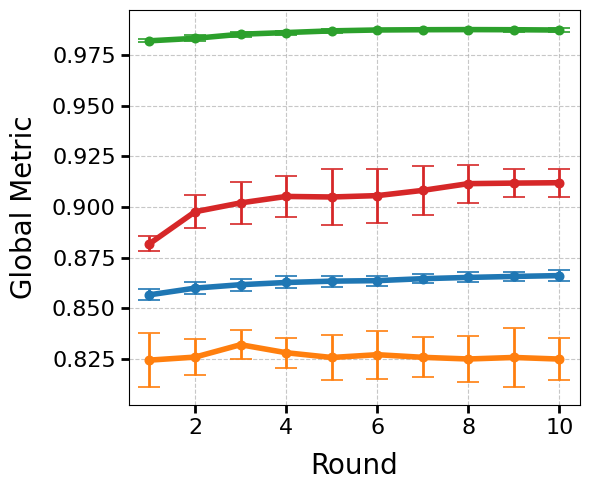}
        \caption{ADULT / 4 clients.}
    \end{subfigure}
    \hfill
    \begin{subfigure}{0.3\textwidth}
        \includegraphics[width=\linewidth]{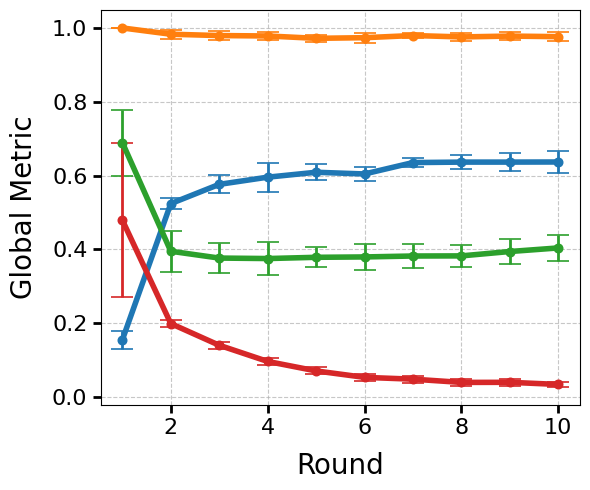}
        \caption{CIFAR / 4 clients.}
    \end{subfigure}
    \hfill
    \begin{subfigure}{0.3\textwidth}
        \includegraphics[width=\linewidth]{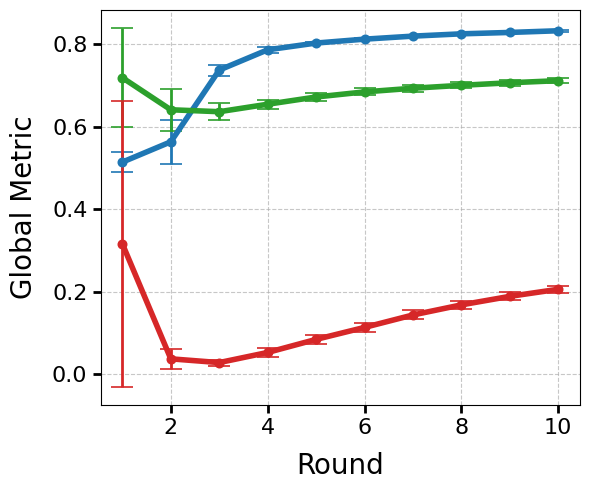}
        \caption{IMDB / 4 clients.}
    \end{subfigure}
    
    \begin{subfigure}{0.3\textwidth}
        \includegraphics[width=\linewidth]{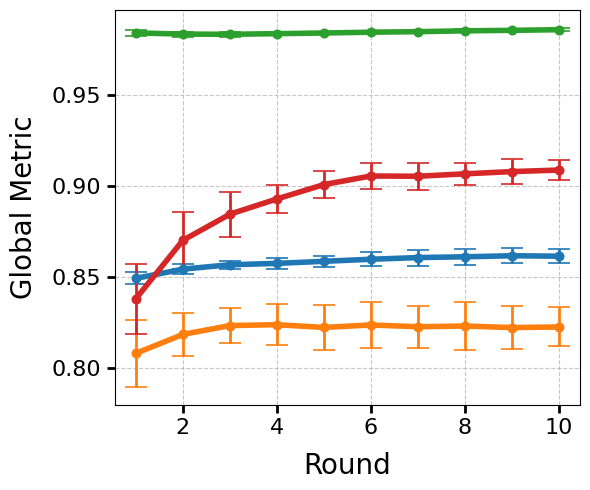}
        \caption{ADULT / 20 clients.}
    \end{subfigure}
    \hfill
    \begin{subfigure}{0.3\textwidth}
        \includegraphics[width=\linewidth]{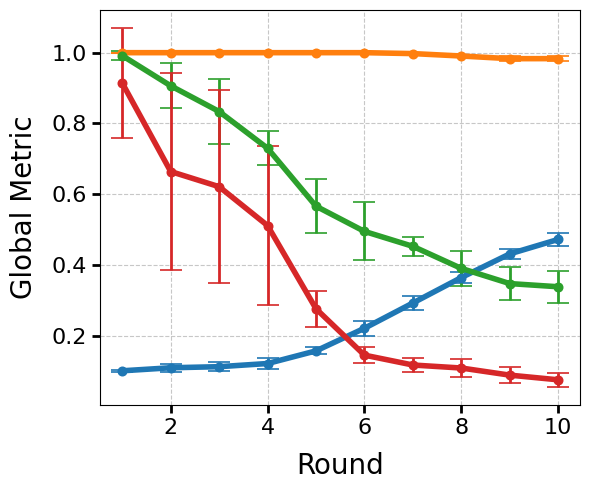}
        \caption{CIFAR / 20 clients.}
    \end{subfigure}
    \hfill 
    \begin{subfigure}{0.3\textwidth}
        \includegraphics[width=\linewidth]{figs/SE_I_20_I.png}
        \caption{IMDB / 20 clients.}
    \end{subfigure}

    \caption{Score evolution of the model as the training progresses with IID distribution.}
    \label{fig:score_evol_IID}
\end{figure}

\begin{figure}[h]
    \centering
    \begin{subfigure}{0.6\textwidth}
        \includegraphics[width=\linewidth]{figs/S_legend.png}
    \end{subfigure}
    
    \begin{subfigure}{0.3\textwidth}
        \includegraphics[width=\linewidth]{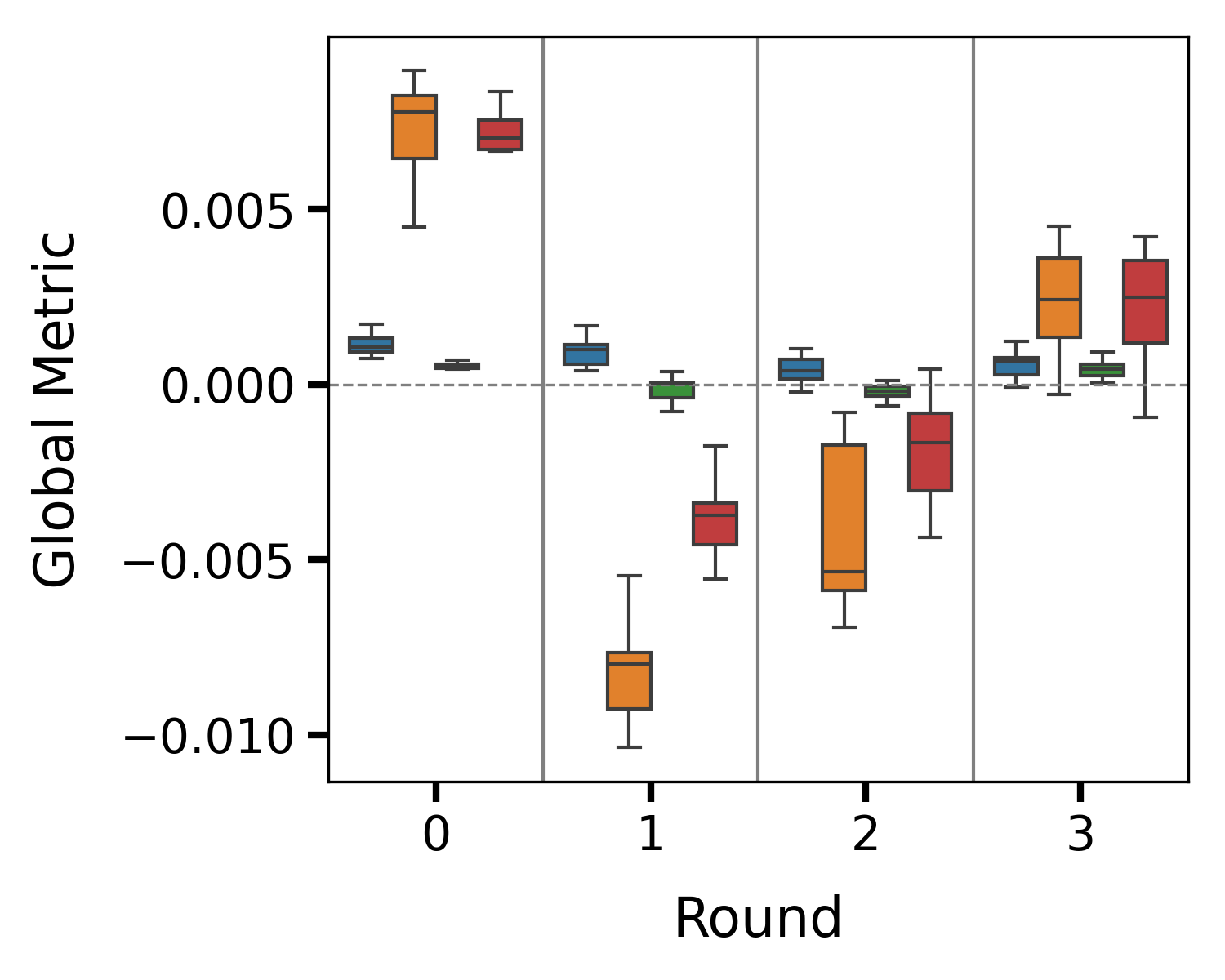}
        \caption{ADULT / LOO.}
    \end{subfigure}
    \hfill
    \begin{subfigure}{0.3\textwidth}
        \includegraphics[width=\linewidth]{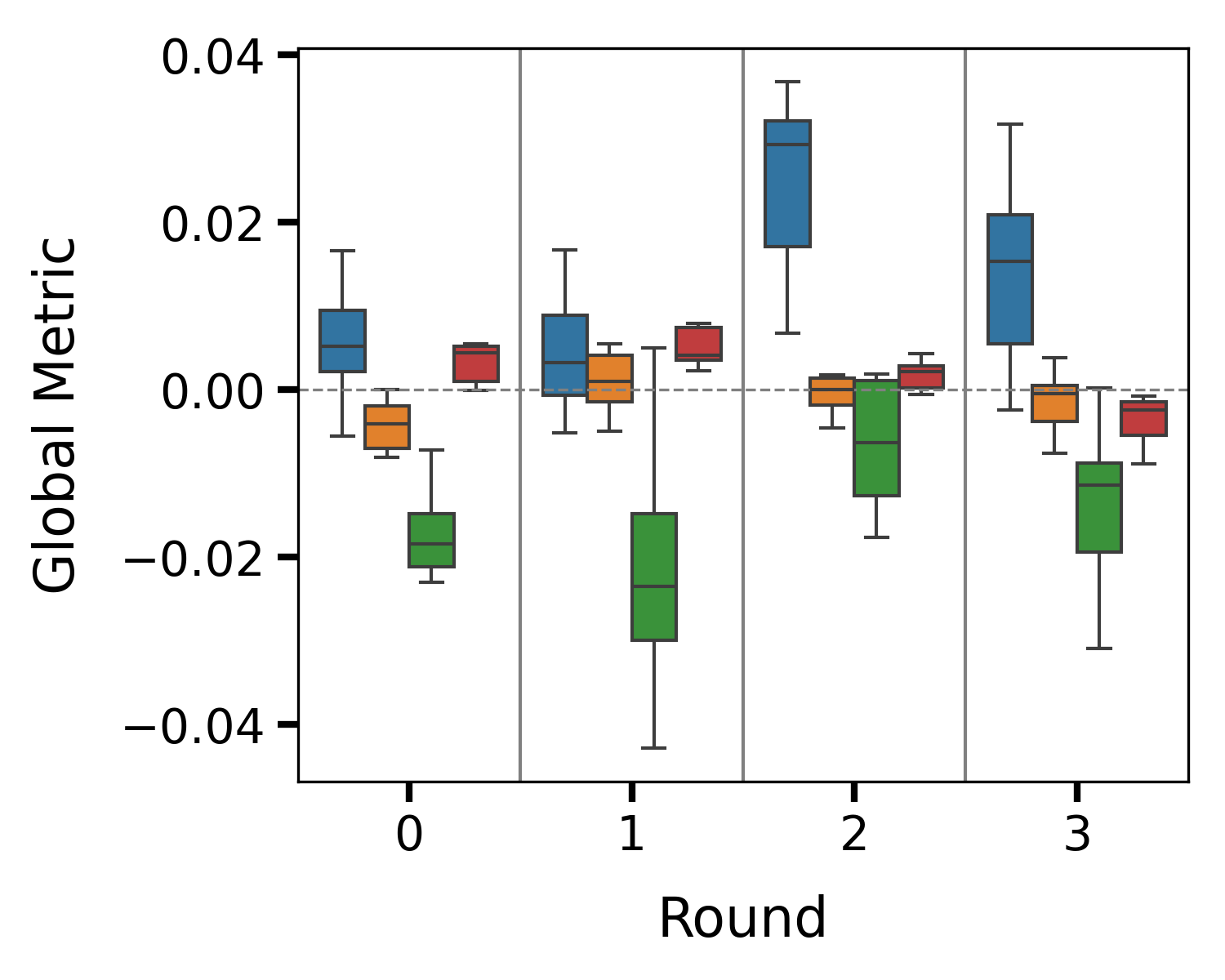}
        \caption{CIFAR / LOO.}
    \end{subfigure}
    \hfill
    \begin{subfigure}{0.3\textwidth}
        \includegraphics[width=\linewidth]{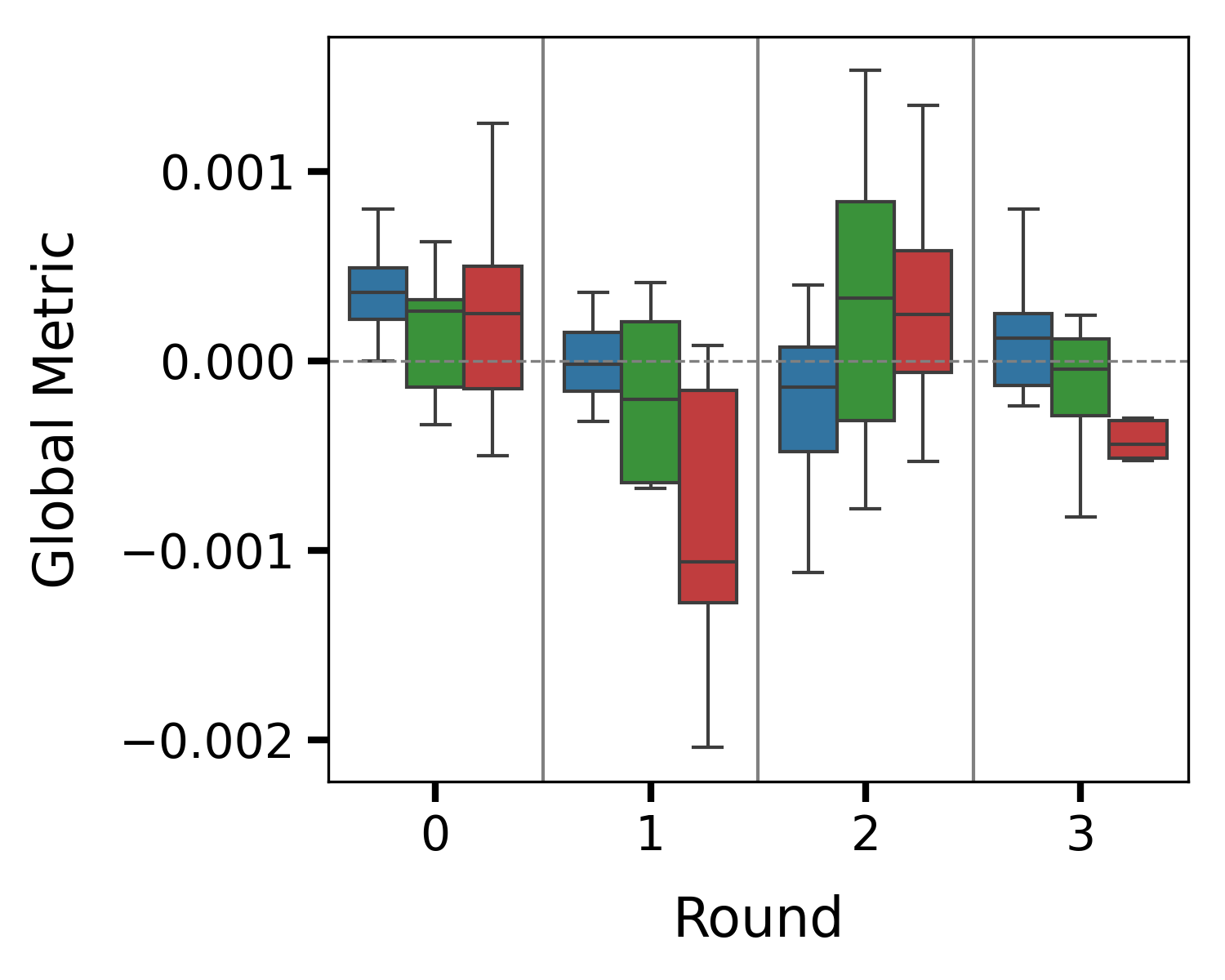}
        \caption{IMDB / LOO.}
    \end{subfigure}
    
    \begin{subfigure}{0.3\textwidth}
        \includegraphics[width=\linewidth]{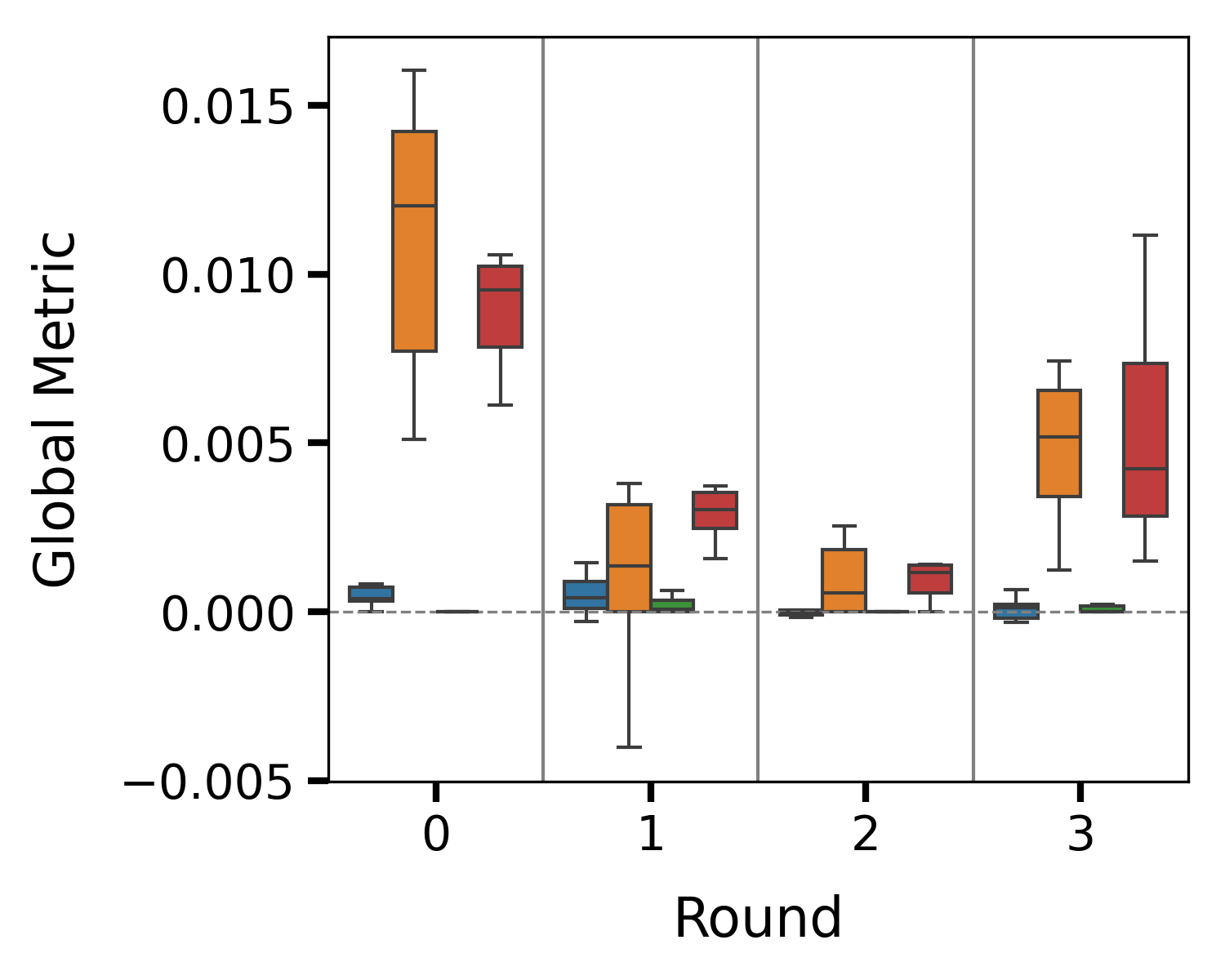}
        \caption{ADULT / GTG.}
    \end{subfigure}
    \hfill
    \begin{subfigure}{0.3\textwidth}
        \includegraphics[width=\linewidth]{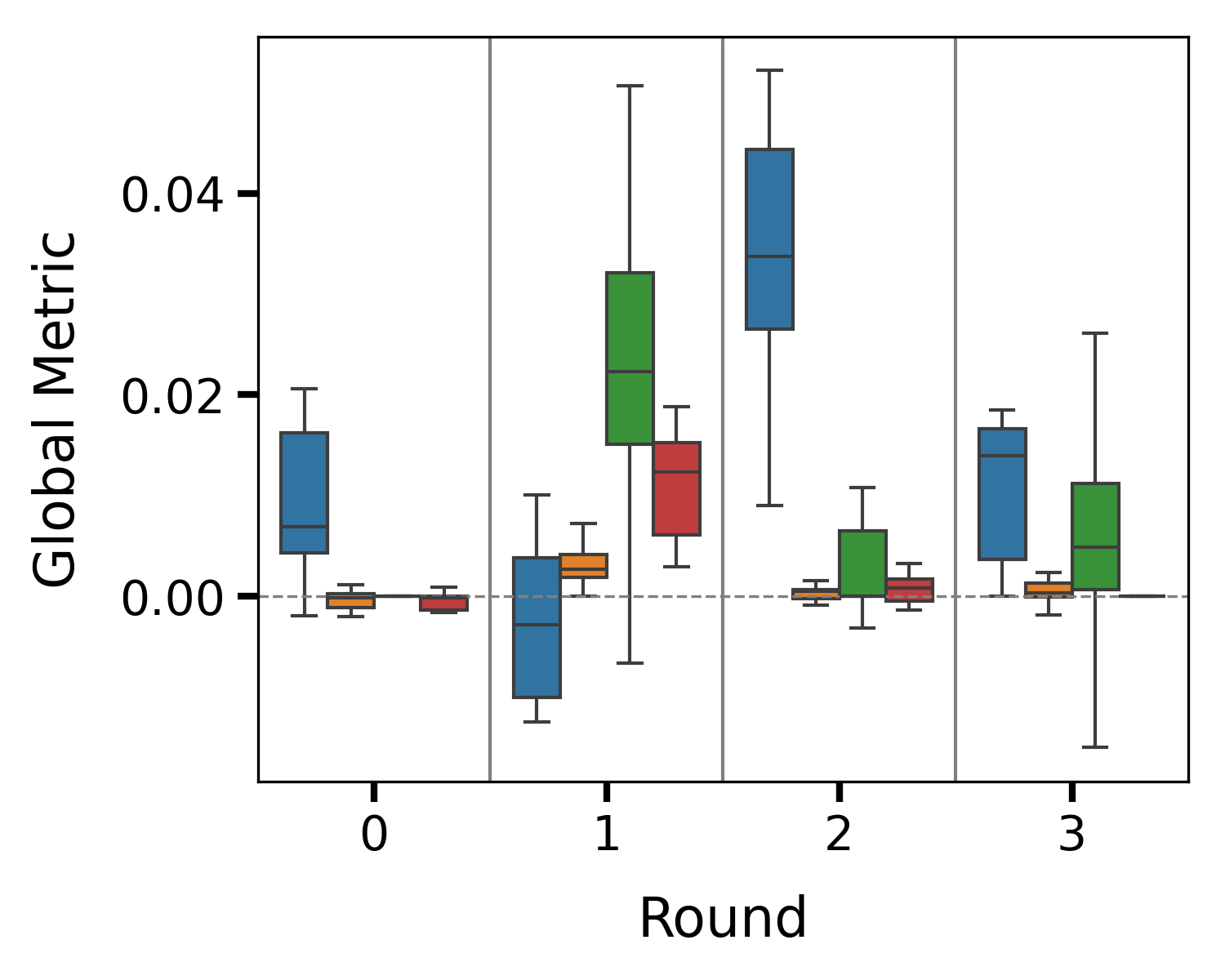}
        \caption{CIFAR / GTG.}
    \end{subfigure}
    \hfill
    \begin{subfigure}{0.3\textwidth}
        \includegraphics[width=\linewidth]{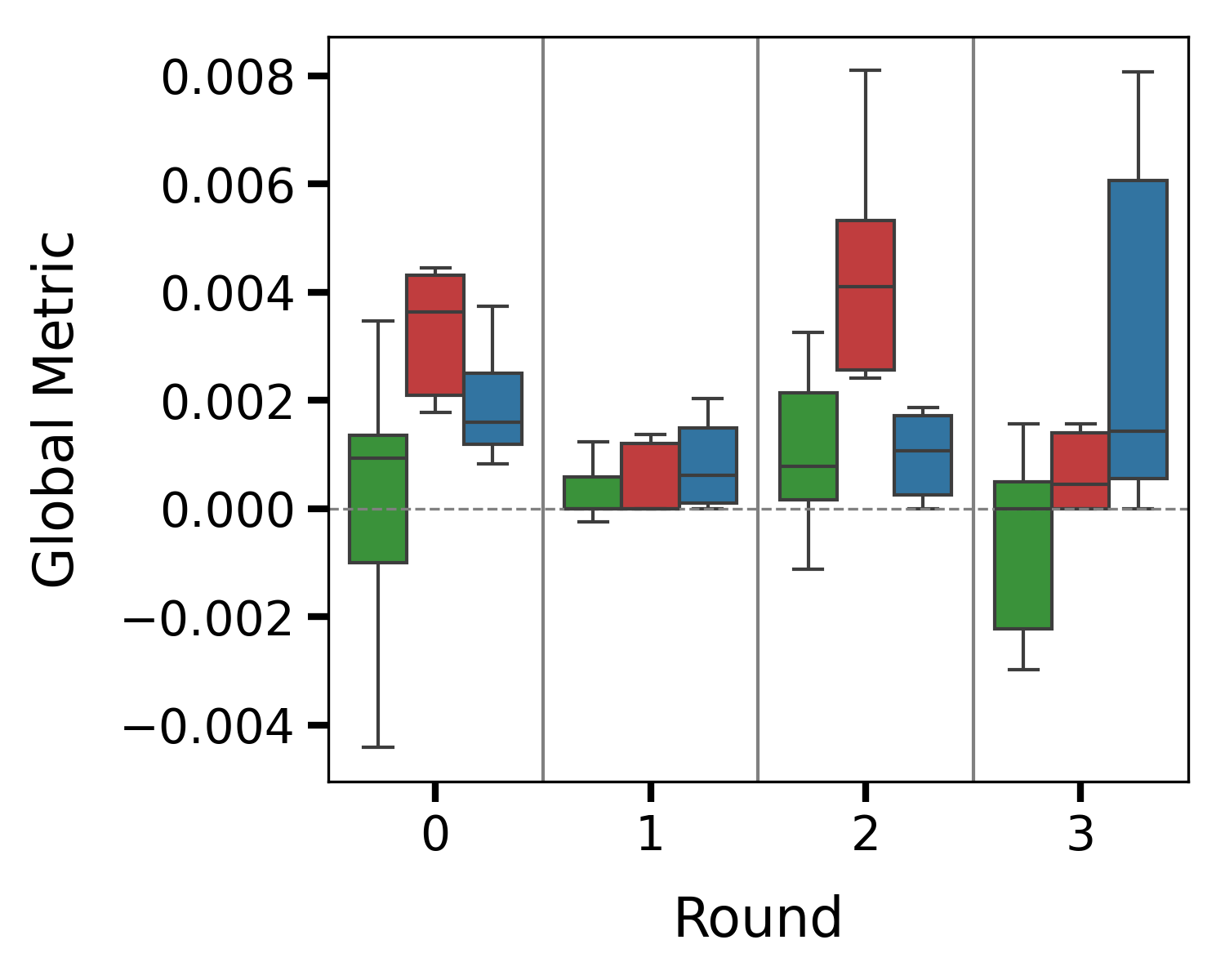}
        \caption{IMDB / GTG.}
    \end{subfigure}
    \caption{Scores of the 4 IID clients.}
    \label{fig:score_4_IID}
\end{figure}

\begin{figure}[h]
    \centering
    \begin{subfigure}{0.6\textwidth}
        \includegraphics[width=\linewidth]{figs/S_legend.png}
    \end{subfigure}
    
    \begin{subfigure}{0.22\textwidth}
        \includegraphics[width=\linewidth]{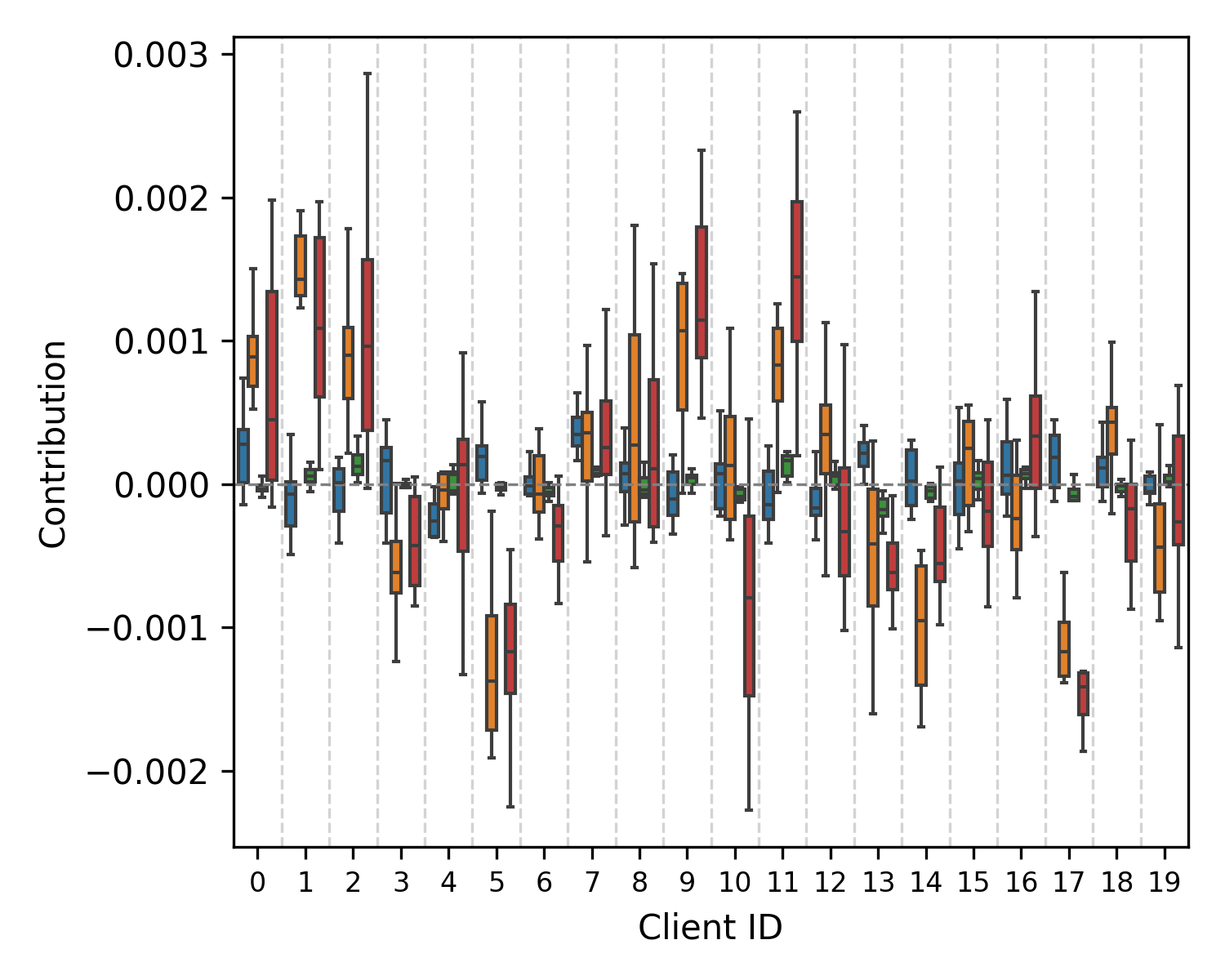}
        \caption{ADULT / IID / LOO.}
    \end{subfigure}
    \hfill
    \begin{subfigure}{0.22\textwidth}
        \includegraphics[width=\linewidth]{figs/S_A_20_I_loo.png}
        \caption{ADULT / IID / GTG.}
    \end{subfigure}
    \hfill
    \begin{subfigure}{0.22\textwidth}
        \includegraphics[width=\linewidth]{figs/S_A_20_I_loo.png}
        \caption{ADULT / NIID / LOO.}
    \end{subfigure}
    \hfill
    \begin{subfigure}{0.22\textwidth}
        \includegraphics[width=\linewidth]{figs/S_A_20_I_loo.png}
        \caption{ADULT / NIID / LOO.}
    \end{subfigure}

    \begin{subfigure}{0.22\textwidth}
        \includegraphics[width=\linewidth]{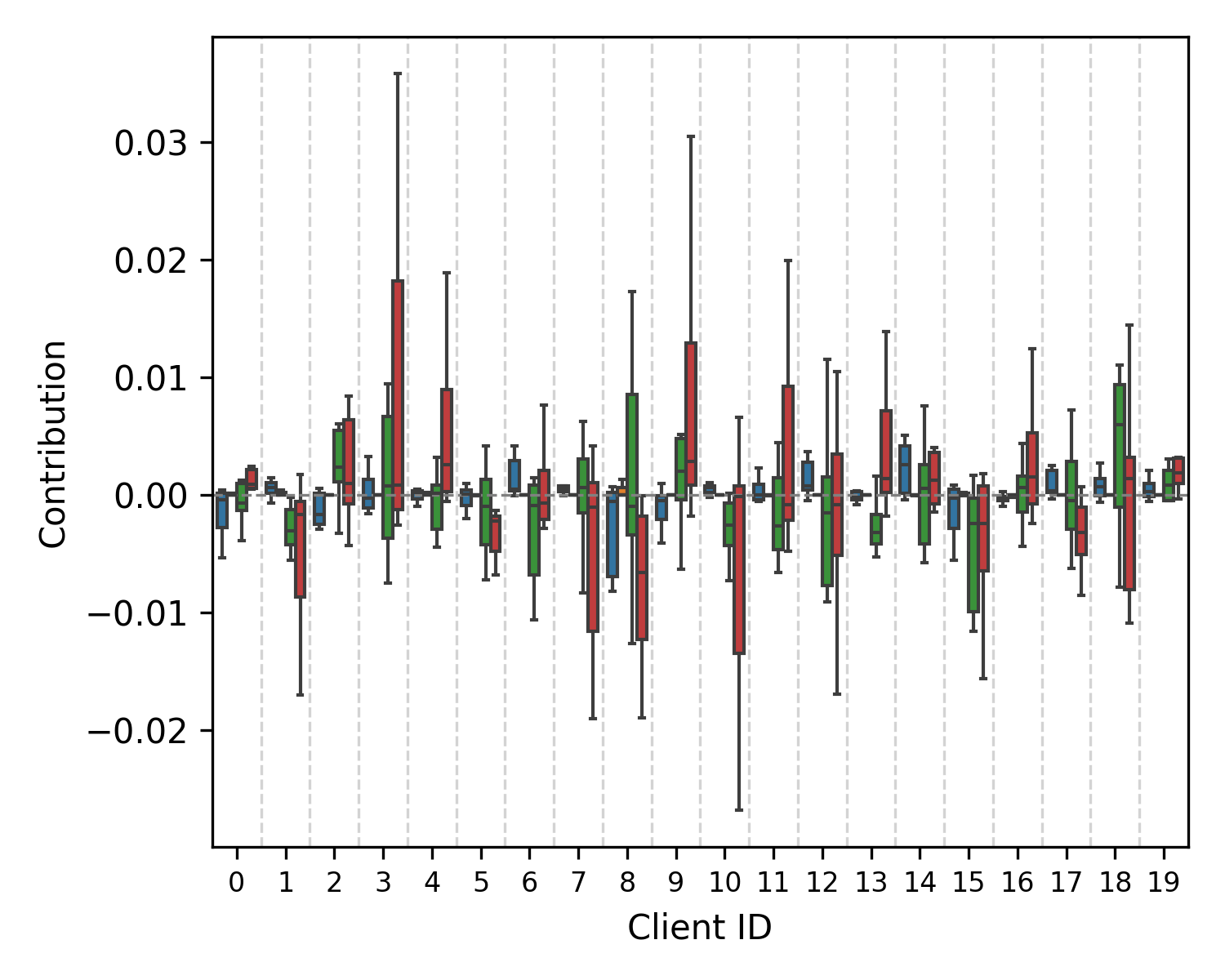}
        \caption{CIFAR / IID / LOO.}
    \end{subfigure}
    \hfill
    \begin{subfigure}{0.22\textwidth}
        \includegraphics[width=\linewidth]{figs/S_C_20_I_loo.png}
        \caption{CIFAR / IID / GTG.}
    \end{subfigure}
    \hfill
    \begin{subfigure}{0.22\textwidth}
        \includegraphics[width=\linewidth]{figs/S_C_20_I_loo.png}
        \caption{CIFAR / NIID / LOO.}
    \end{subfigure}
    \hfill
    \begin{subfigure}{0.22\textwidth}
        \includegraphics[width=\linewidth]{figs/S_C_20_I_loo.png}
        \caption{CIFAR / NIID / LOO.}
    \end{subfigure}

    \begin{subfigure}{0.22\textwidth}
        \includegraphics[width=\linewidth]{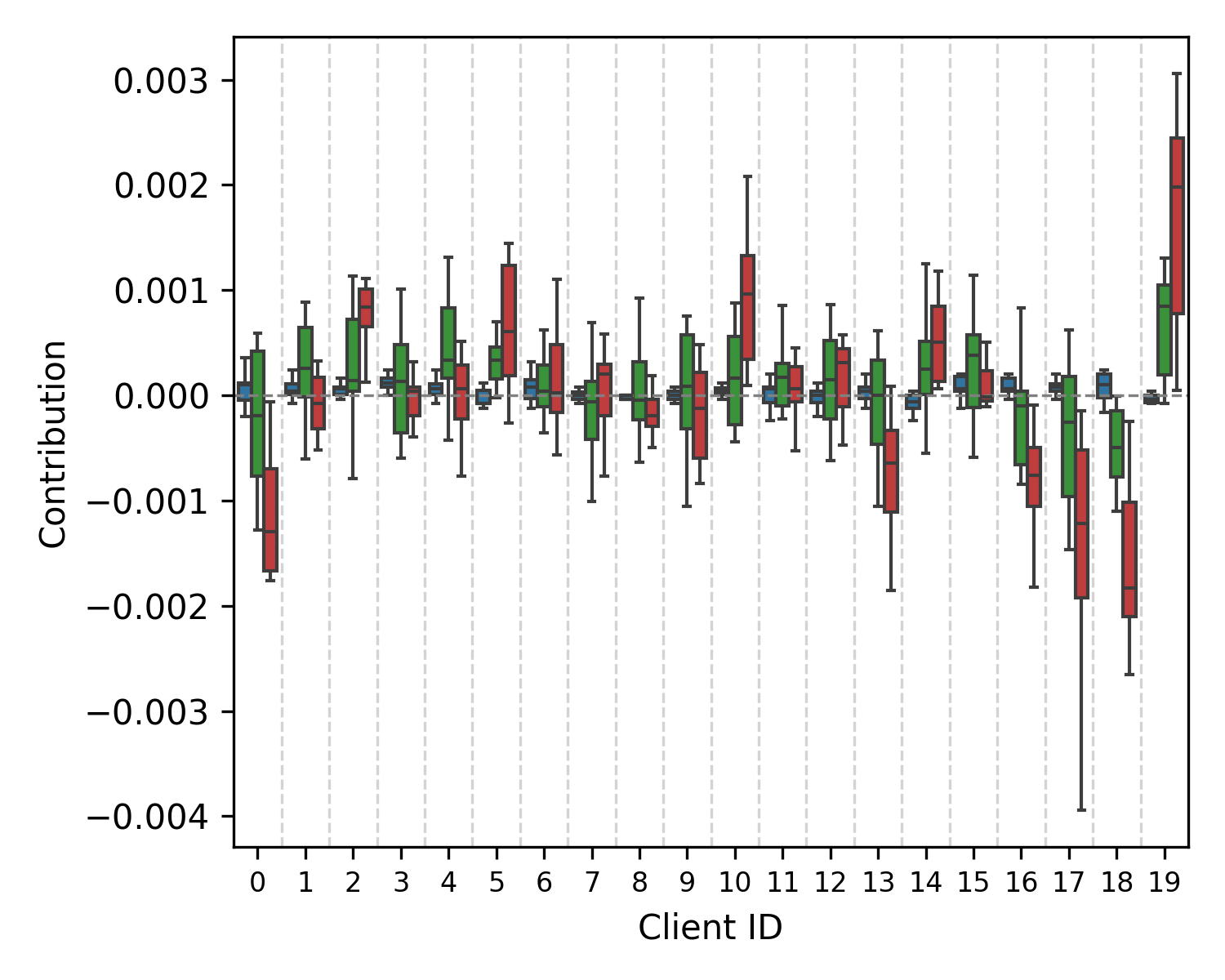}
        \caption{IMDB / IID / LOO.}
    \end{subfigure}
    \hfill
    \begin{subfigure}{0.22\textwidth}
        \includegraphics[width=\linewidth]{figs/S_I_20_I_loo.png}
        \caption{IMDB / IID / GTG.}
    \end{subfigure}
    \hfill
    \begin{subfigure}{0.22\textwidth}
        \includegraphics[width=\linewidth]{figs/S_I_20_I_loo.png}
        \caption{IMDB / NIID / LOO.}
    \end{subfigure}
    \hfill
    \begin{subfigure}{0.22\textwidth}
        \includegraphics[width=\linewidth]{figs/S_I_20_I_loo.png}
        \caption{IMDB / NIID / LOO.}
    \end{subfigure}

    \caption{Scores of the 20 clients.}
    \label{fig:score_20}
\end{figure}

\begin{figure}[h]
    \centering
    \begin{subfigure}{0.6\textwidth}
        \includegraphics[width=\linewidth]{figs/S_legend.png}
    \end{subfigure}
    
    \begin{subfigure}{0.3\textwidth}
        \includegraphics[width=\linewidth]{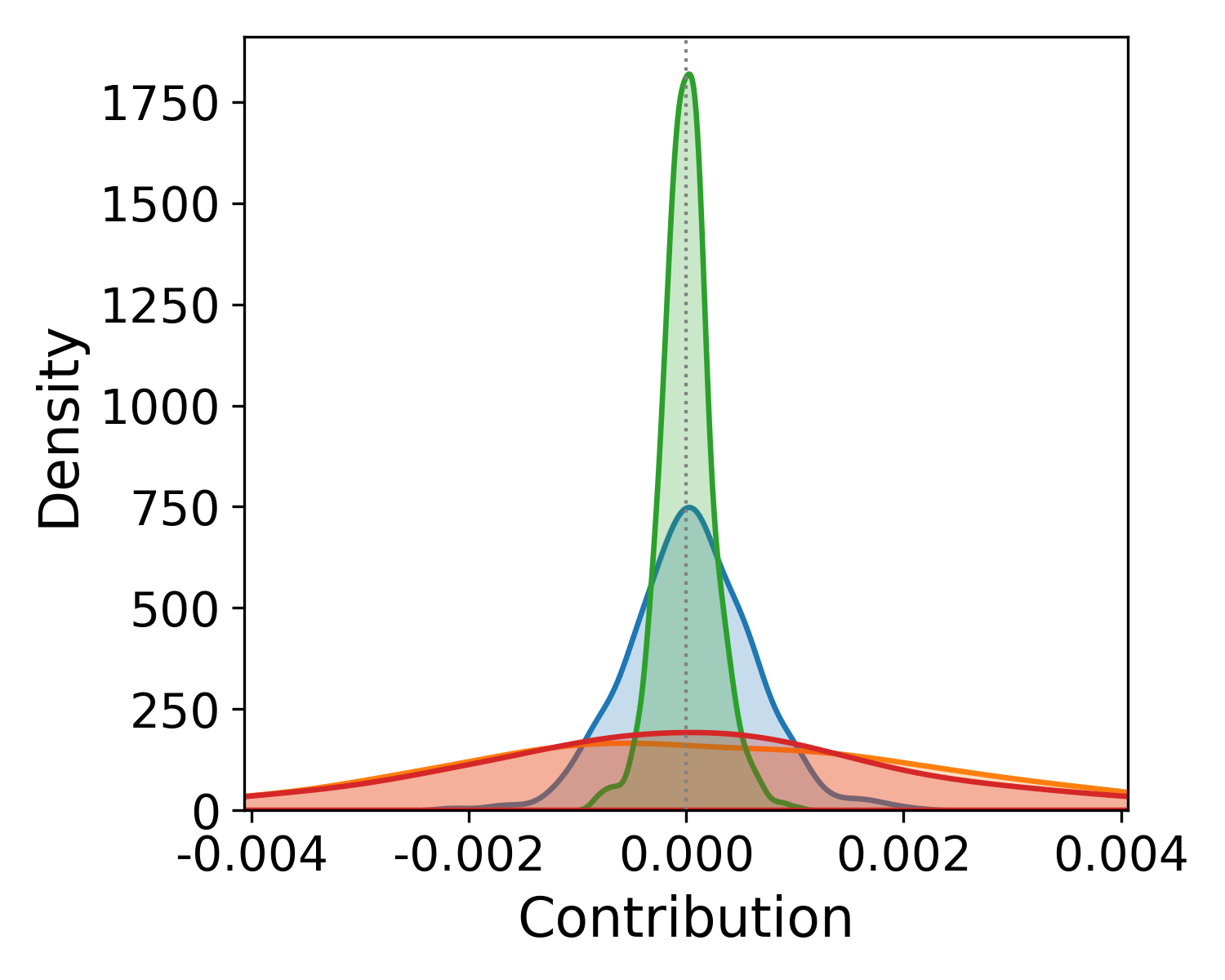}
        \caption{ADULT / LOO.}
    \end{subfigure}
    \hfill
    \begin{subfigure}{0.3\textwidth}
        \includegraphics[width=\linewidth]{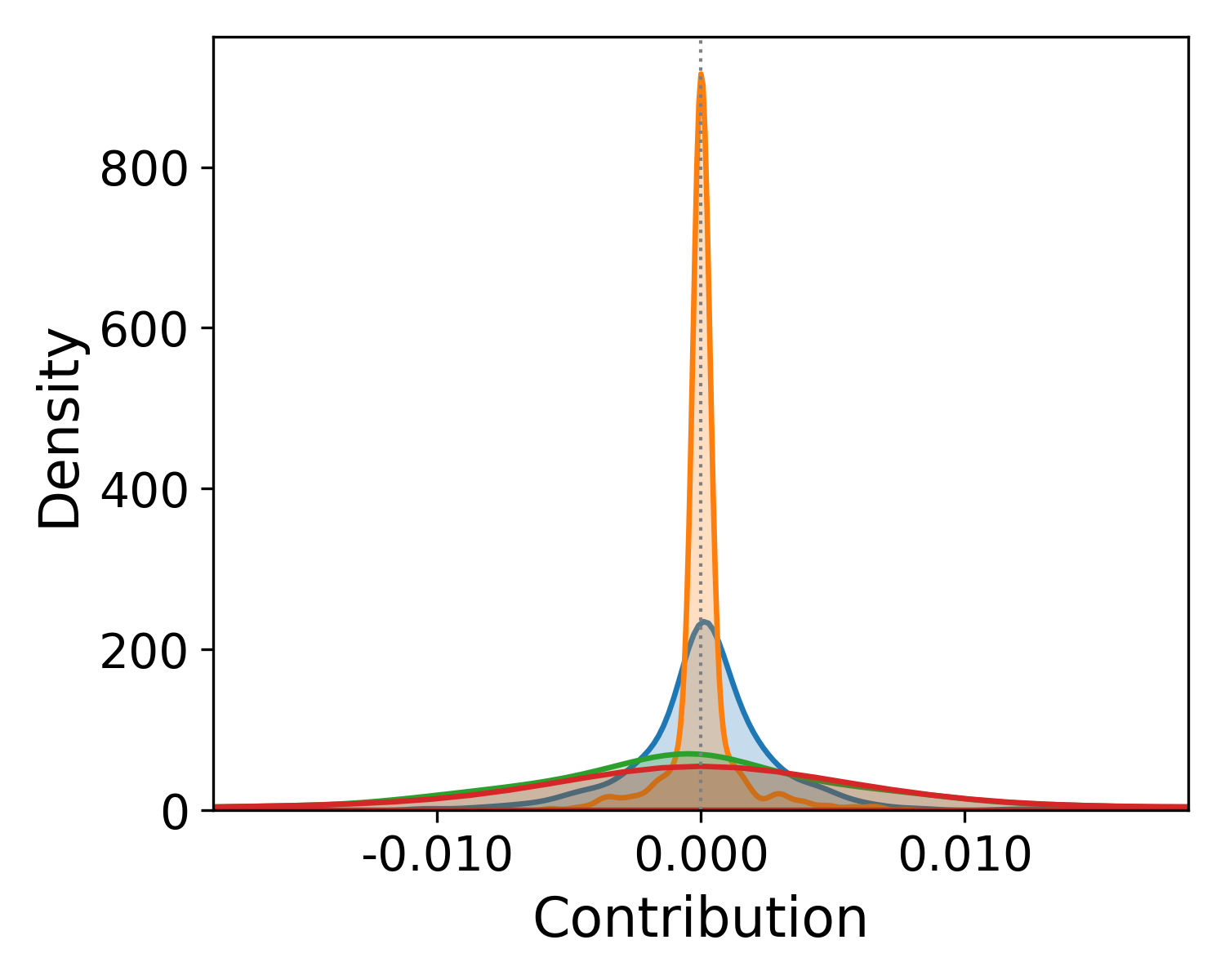}
        \caption{CIFAR / LOO.}
    \end{subfigure}
    \hfill
    \begin{subfigure}{0.3\textwidth}
        \includegraphics[width=\linewidth]{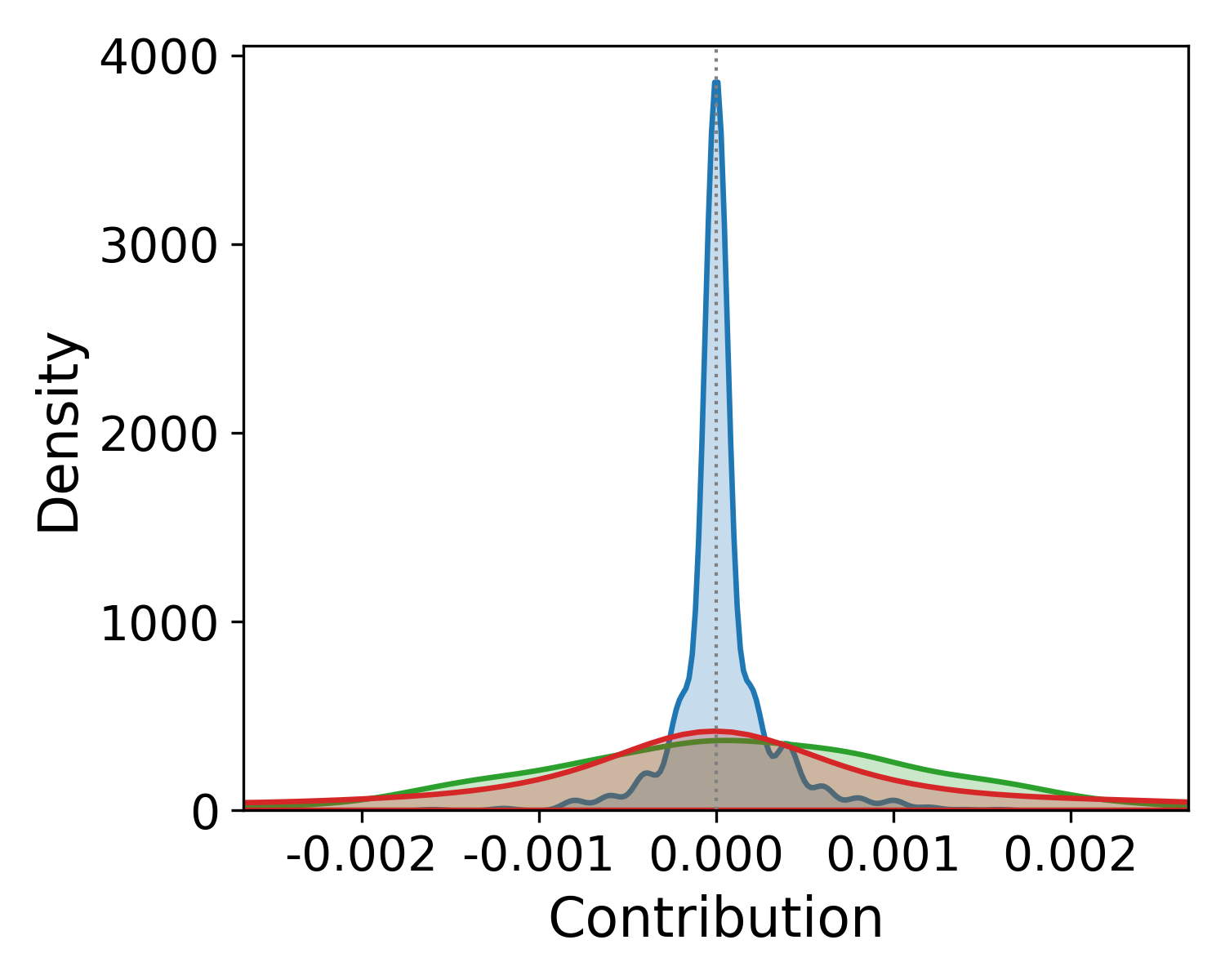}
        \caption{IMDB / LOO.}
    \end{subfigure}
    
    \begin{subfigure}{0.3\textwidth}
        \includegraphics[width=\linewidth]{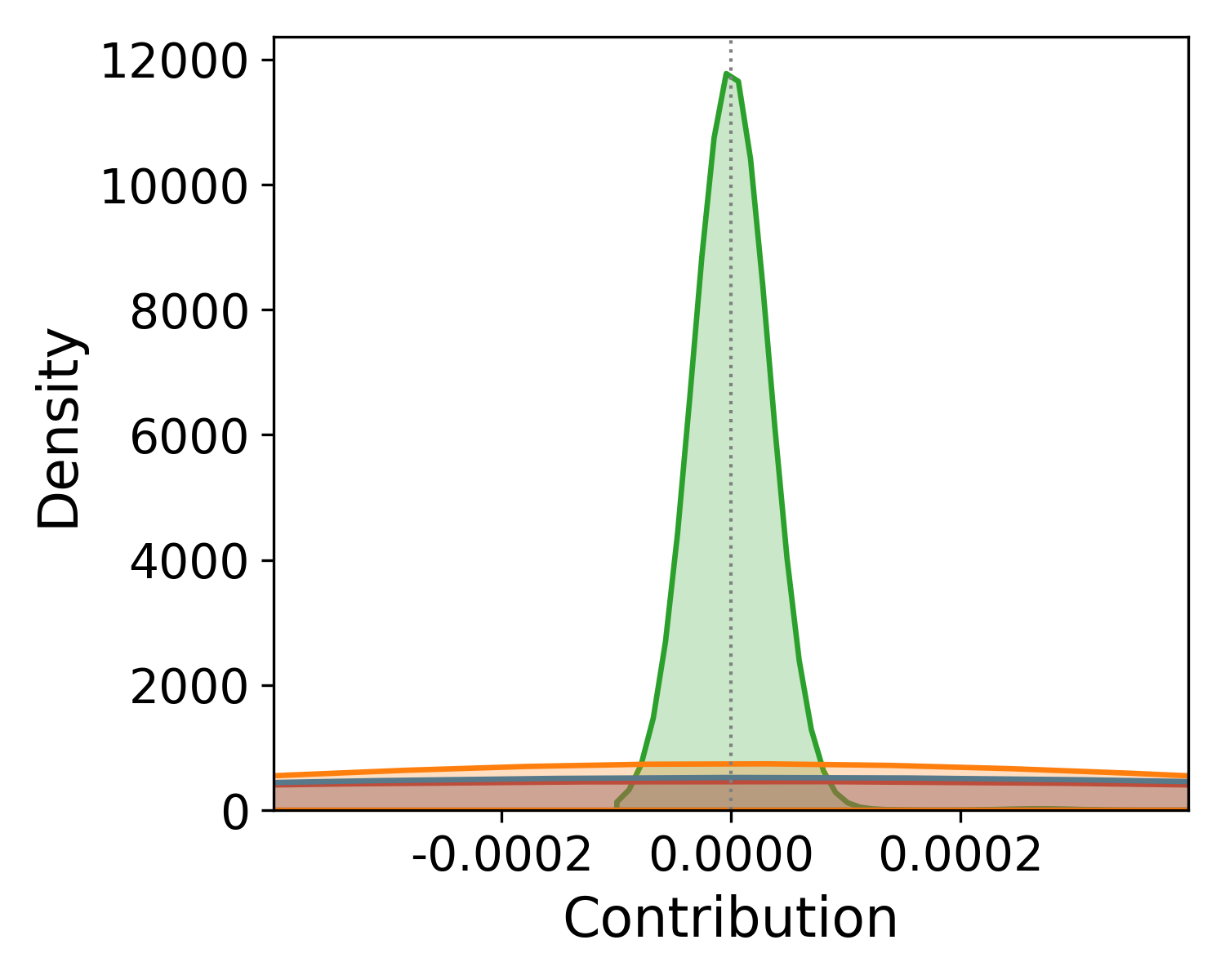}
        \caption{ADULT / GTG.}
    \end{subfigure}
    \hfill
    \begin{subfigure}{0.3\textwidth}
        \includegraphics[width=\linewidth]{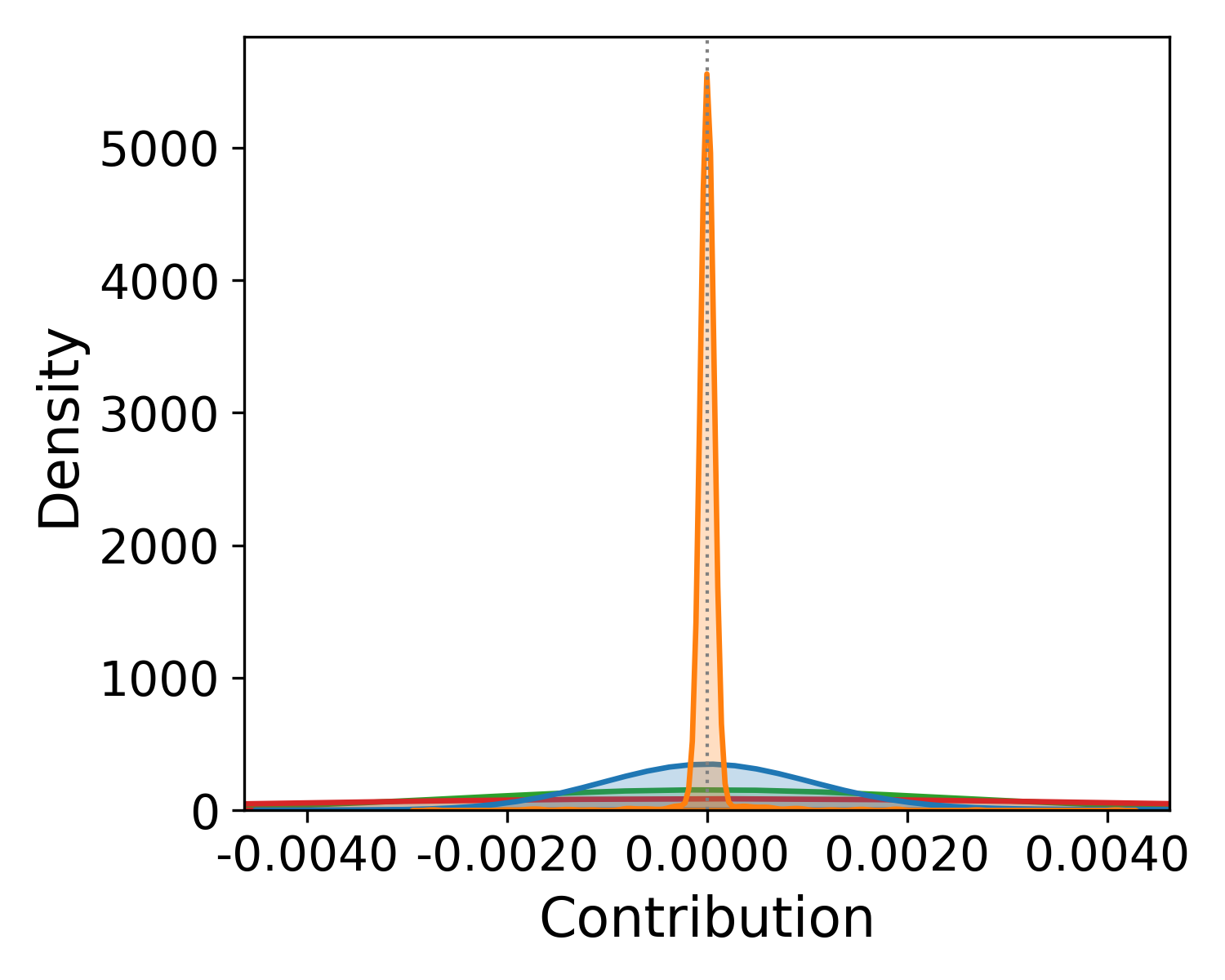}
        \caption{CIFAR / GTG.}
    \end{subfigure}
    \hfill
    \begin{subfigure}{0.3\textwidth}
        \includegraphics[width=\linewidth]{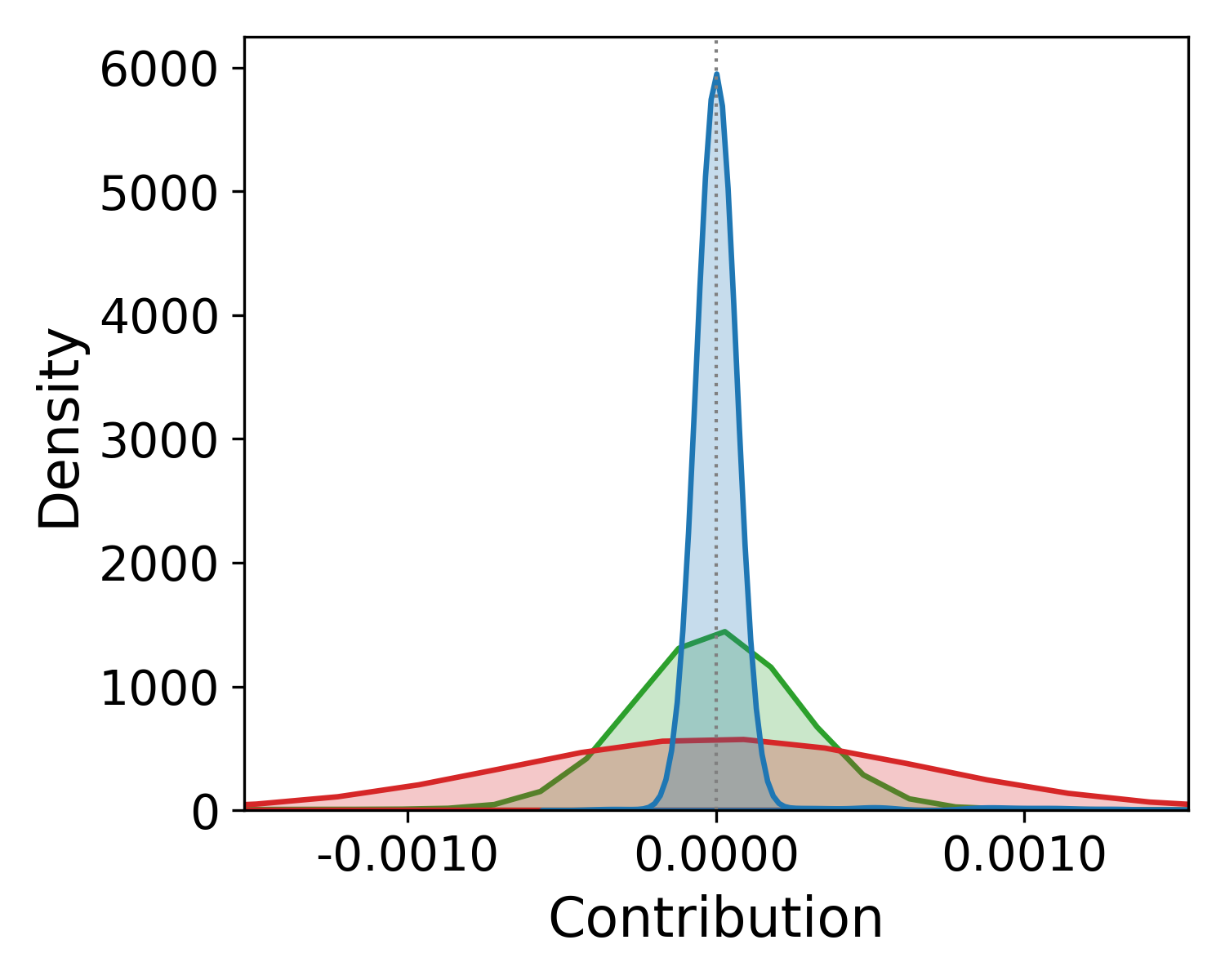}
        \caption{IMDB / GTG.}
    \end{subfigure}    
    \caption{Score distributions of the 20 IID clients.}
    \label{fig:score_dist_20_IID}
\end{figure}

\begin{figure}[h]
    \centering
    \begin{subfigure}{0.6\textwidth}
        \includegraphics[width=\linewidth]{figs/S_legend.png}
    \end{subfigure}
    
    \begin{subfigure}{0.22\textwidth}
        \includegraphics[width=\linewidth]{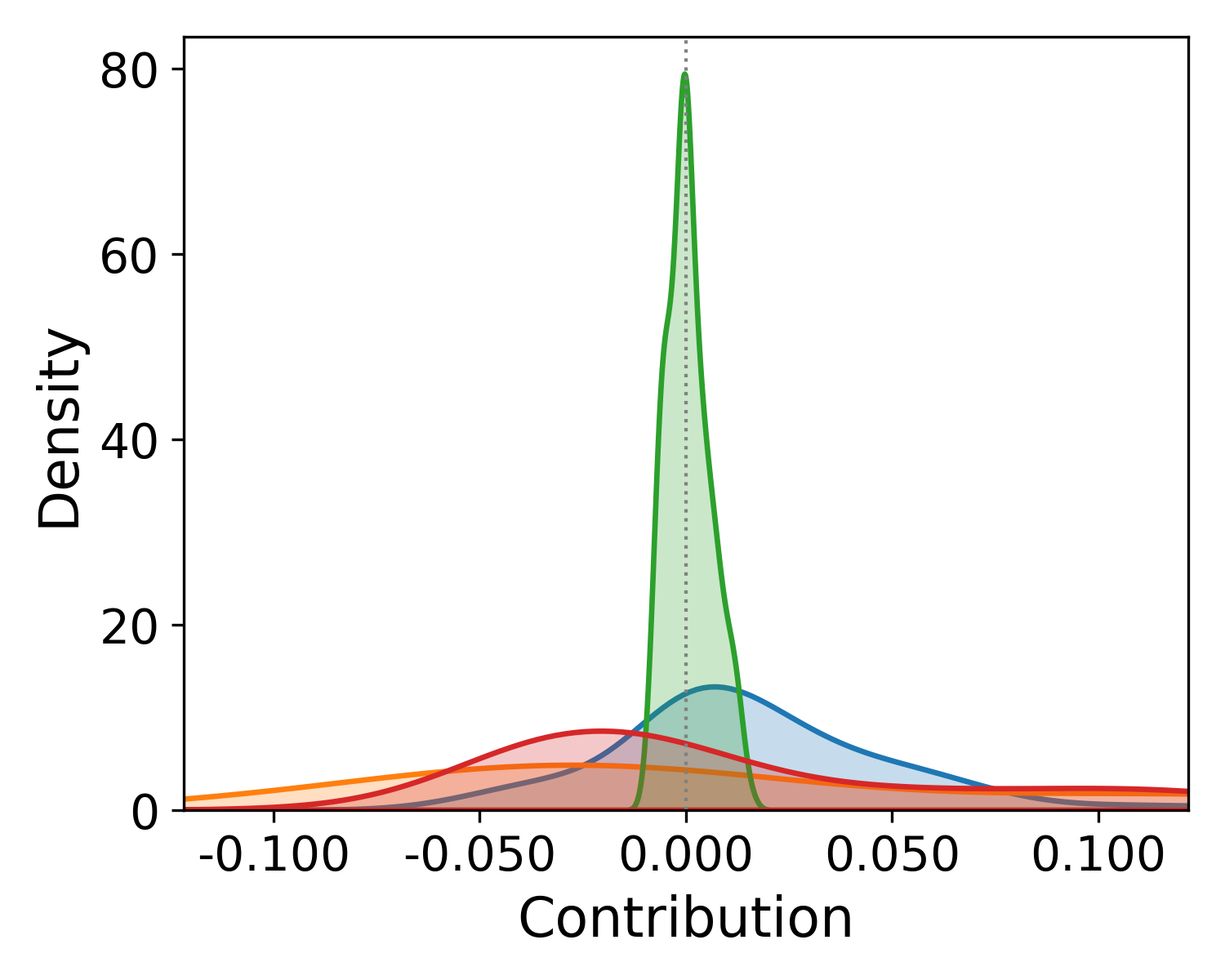}
        \caption{ADULT / NIID / LOO.}
    \end{subfigure}
    \hfill
    \begin{subfigure}{0.22\textwidth}
        \includegraphics[width=\linewidth]{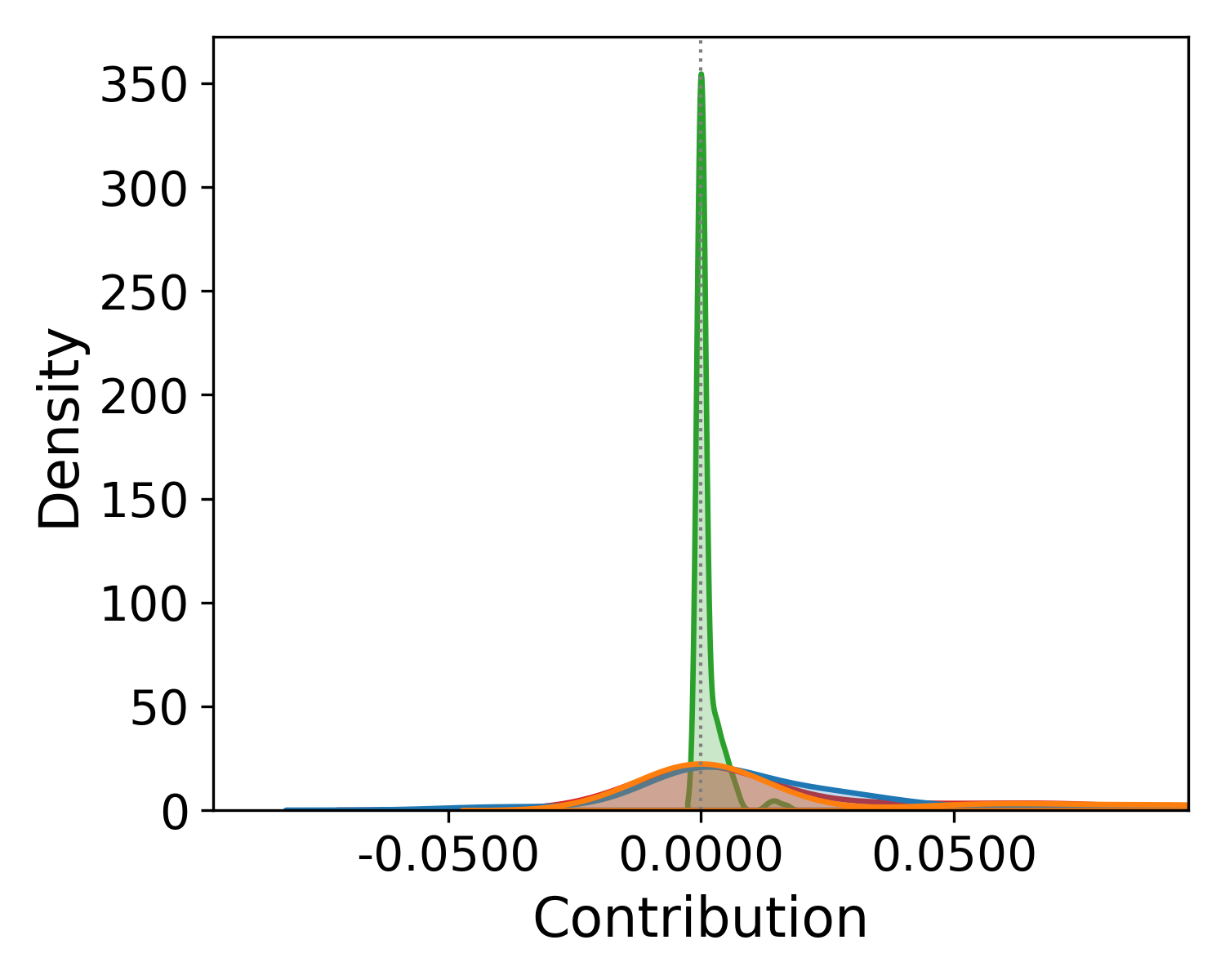}
        \caption{ADULT / NIID / GTG.}
    \end{subfigure}
    \hfill
    \begin{subfigure}{0.22\textwidth}
        \includegraphics[width=\linewidth]{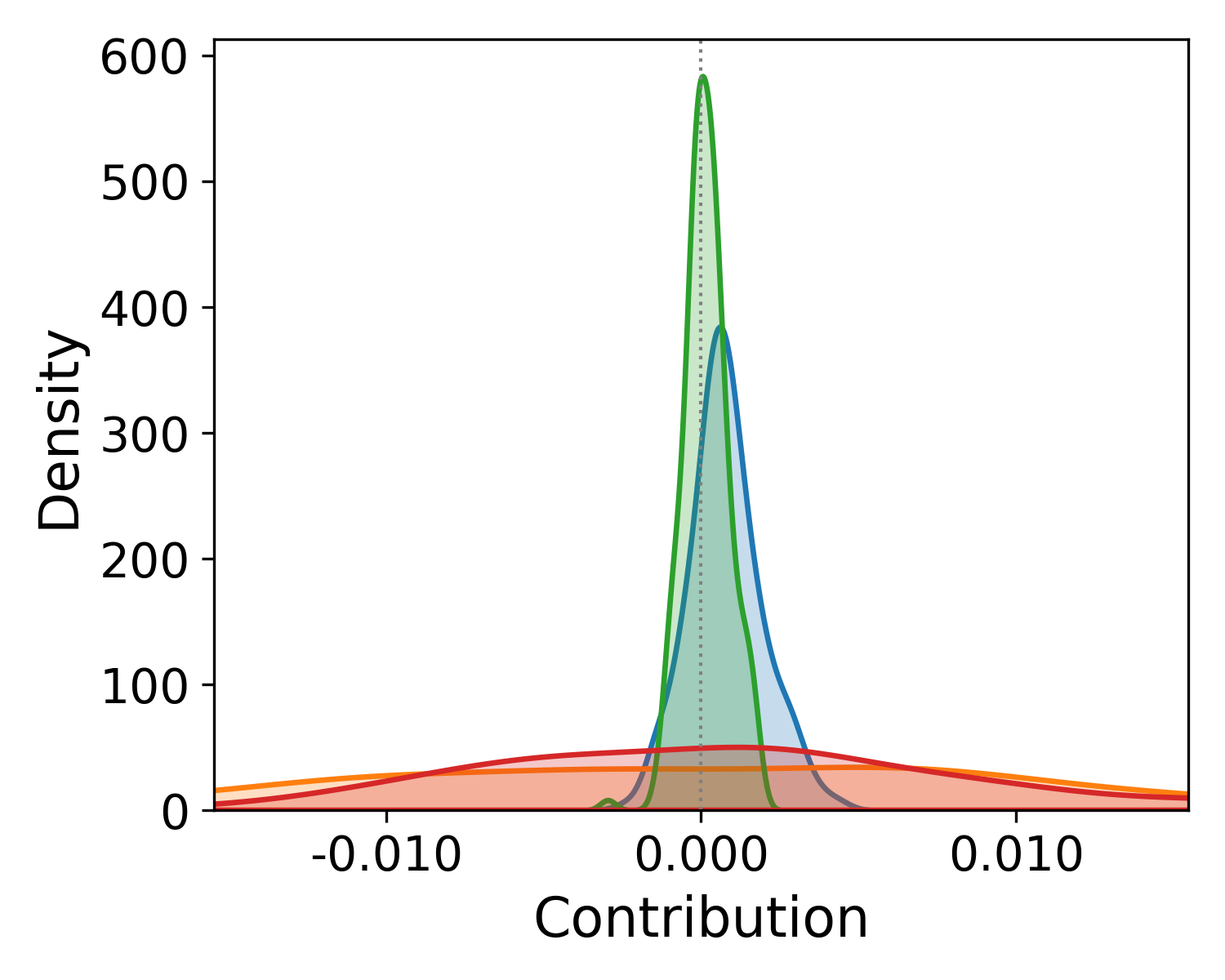}
        \caption{ADULT / IID / LOO.}
    \end{subfigure}
    \hfill
    \begin{subfigure}{0.22\textwidth}
        \includegraphics[width=\linewidth]{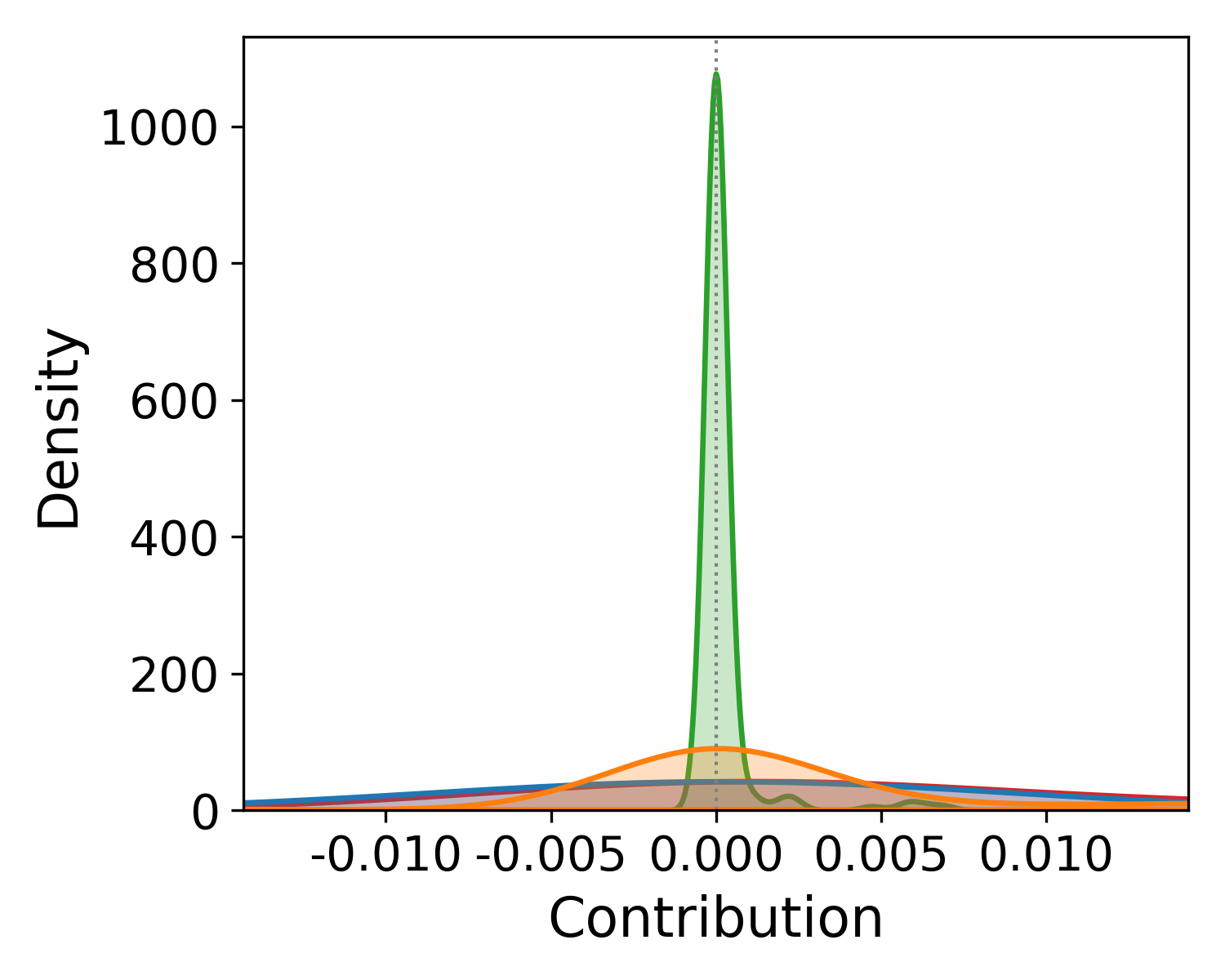}
        \caption{ADULT / IID / GTG.}
    \end{subfigure}

    \begin{subfigure}{0.22\textwidth}
        \includegraphics[width=\linewidth]{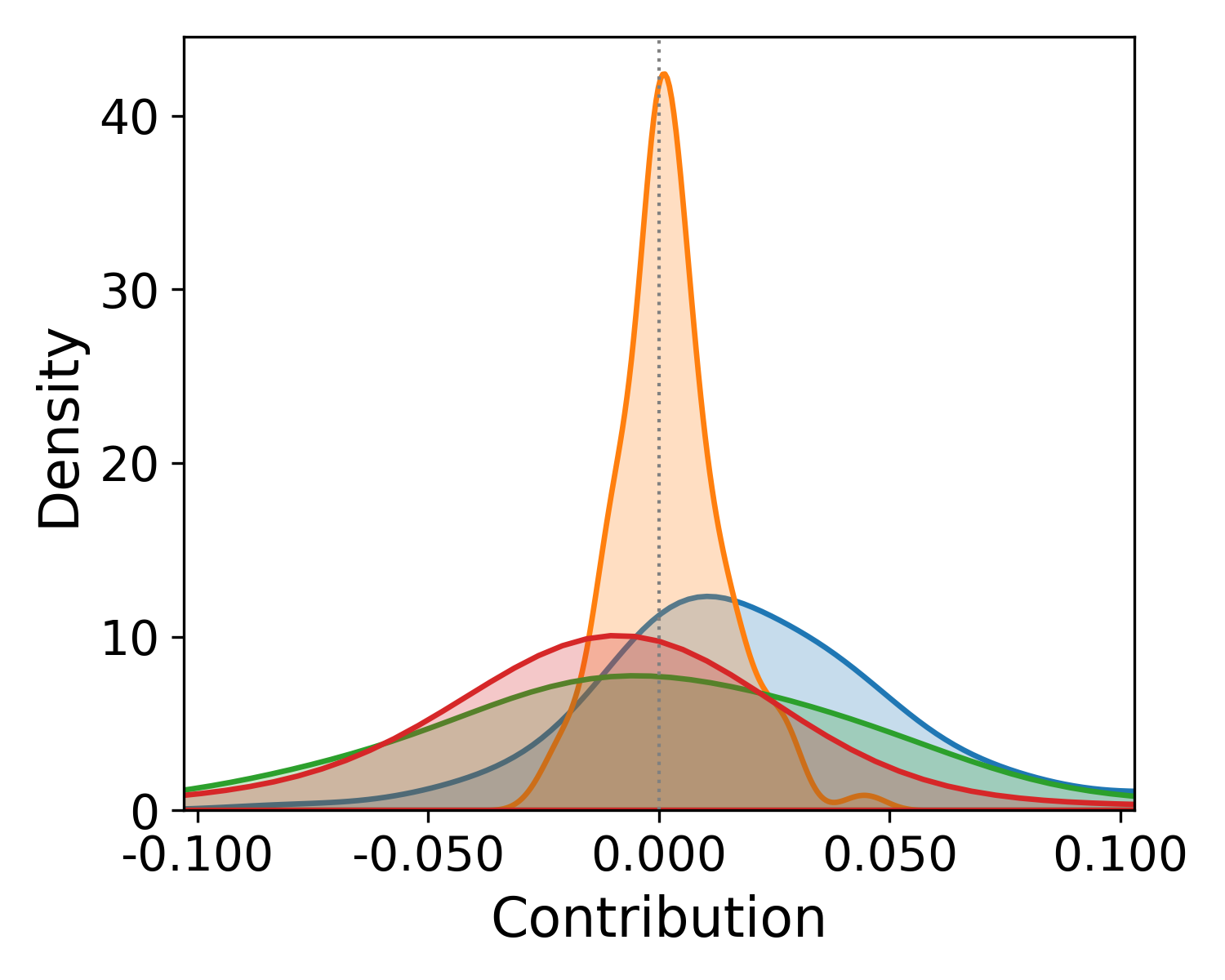}
        \caption{CIFAR / NIID / LOO.}
    \end{subfigure}
    \hfill
    \begin{subfigure}{0.22\textwidth}
        \includegraphics[width=\linewidth]{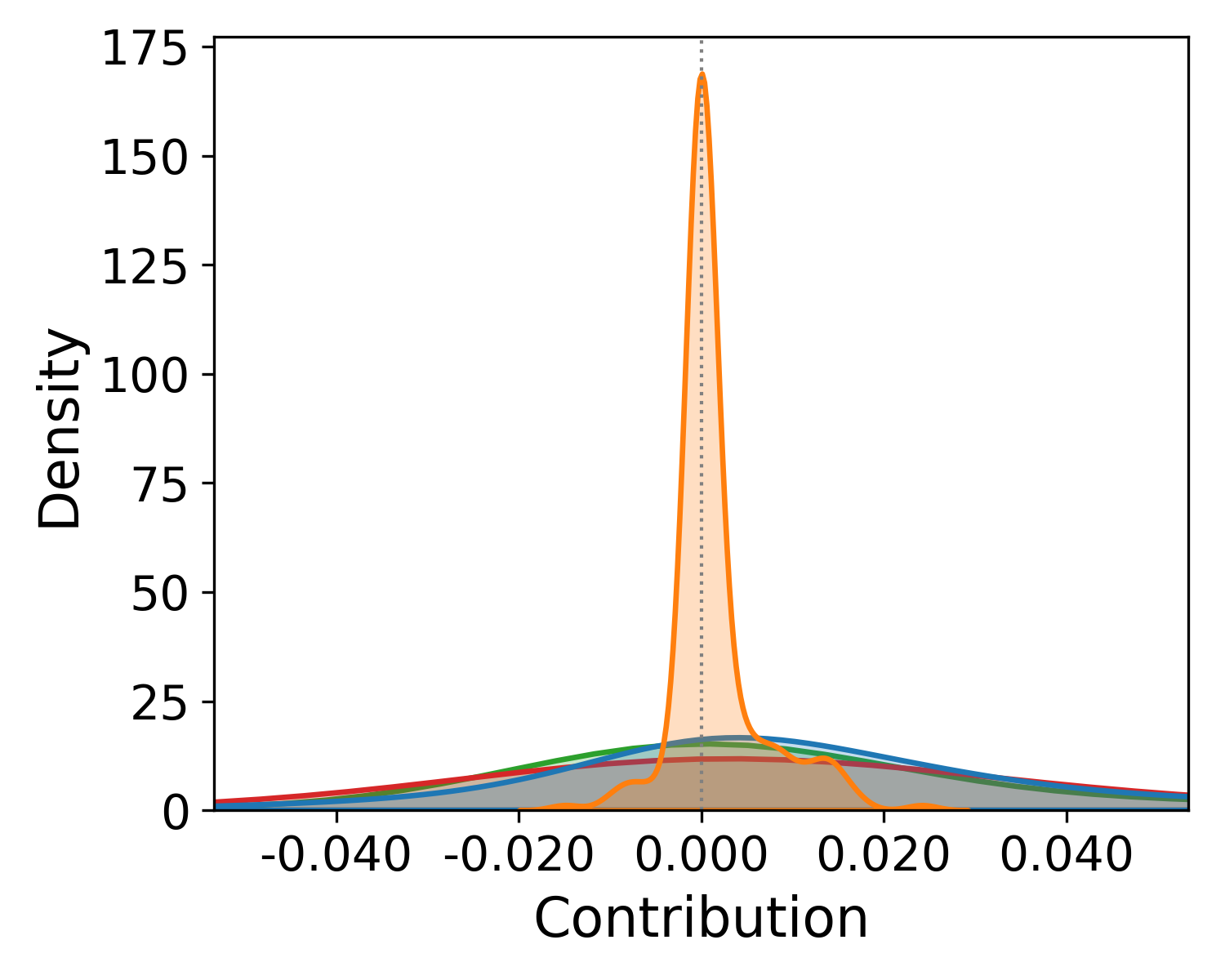}
        \caption{CIFAR / NIID / GTG.}
    \end{subfigure}
    \hfill
    \begin{subfigure}{0.22\textwidth}
        \includegraphics[width=\linewidth]{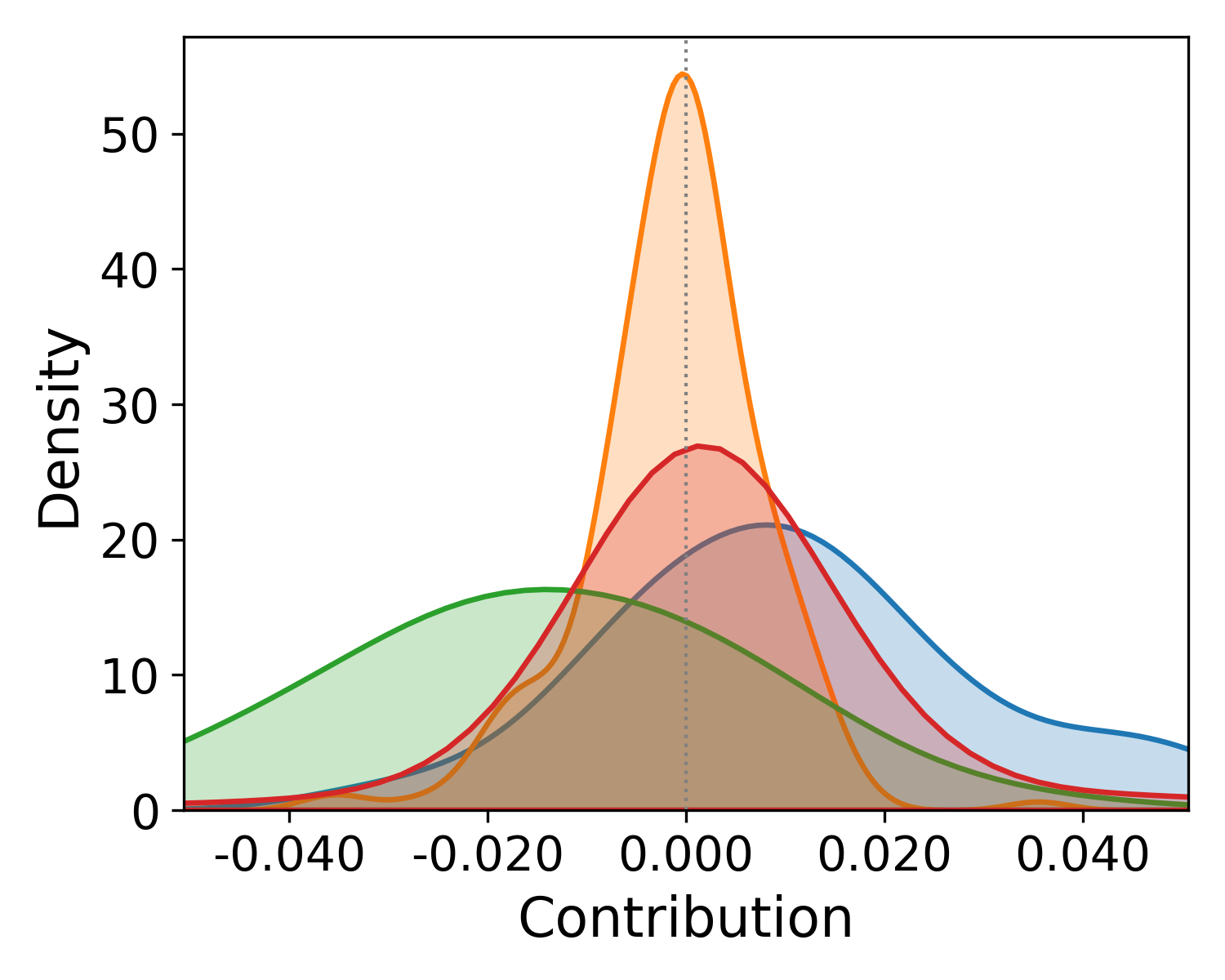}
        \caption{CIFAR / IID / LOO.}
    \end{subfigure}
    \hfill
    \begin{subfigure}{0.22\textwidth}
        \includegraphics[width=\linewidth]{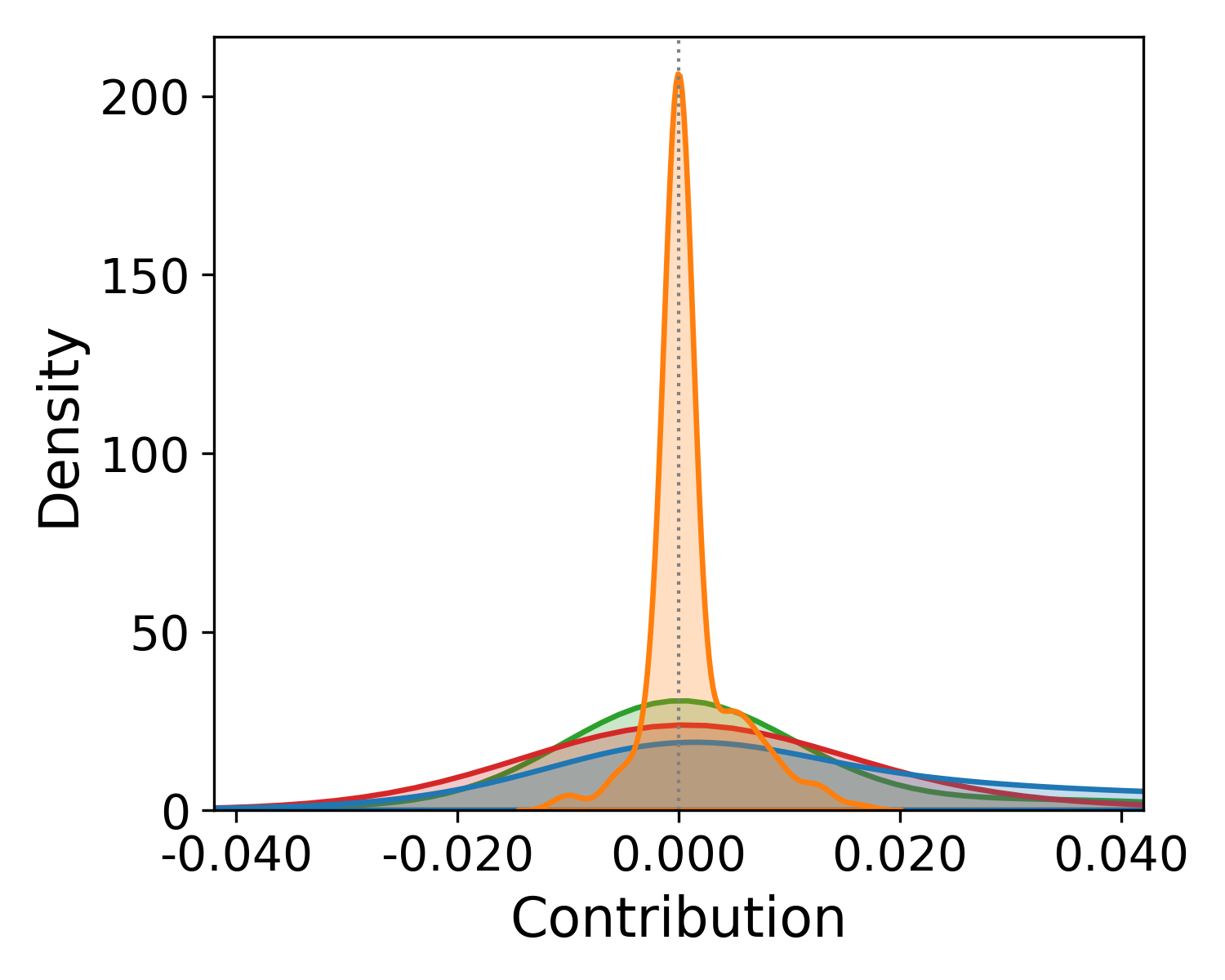}
        \caption{CIFAR / IID / GTG.}
    \end{subfigure}
    
    \begin{subfigure}{0.22\textwidth}
        \includegraphics[width=\linewidth]{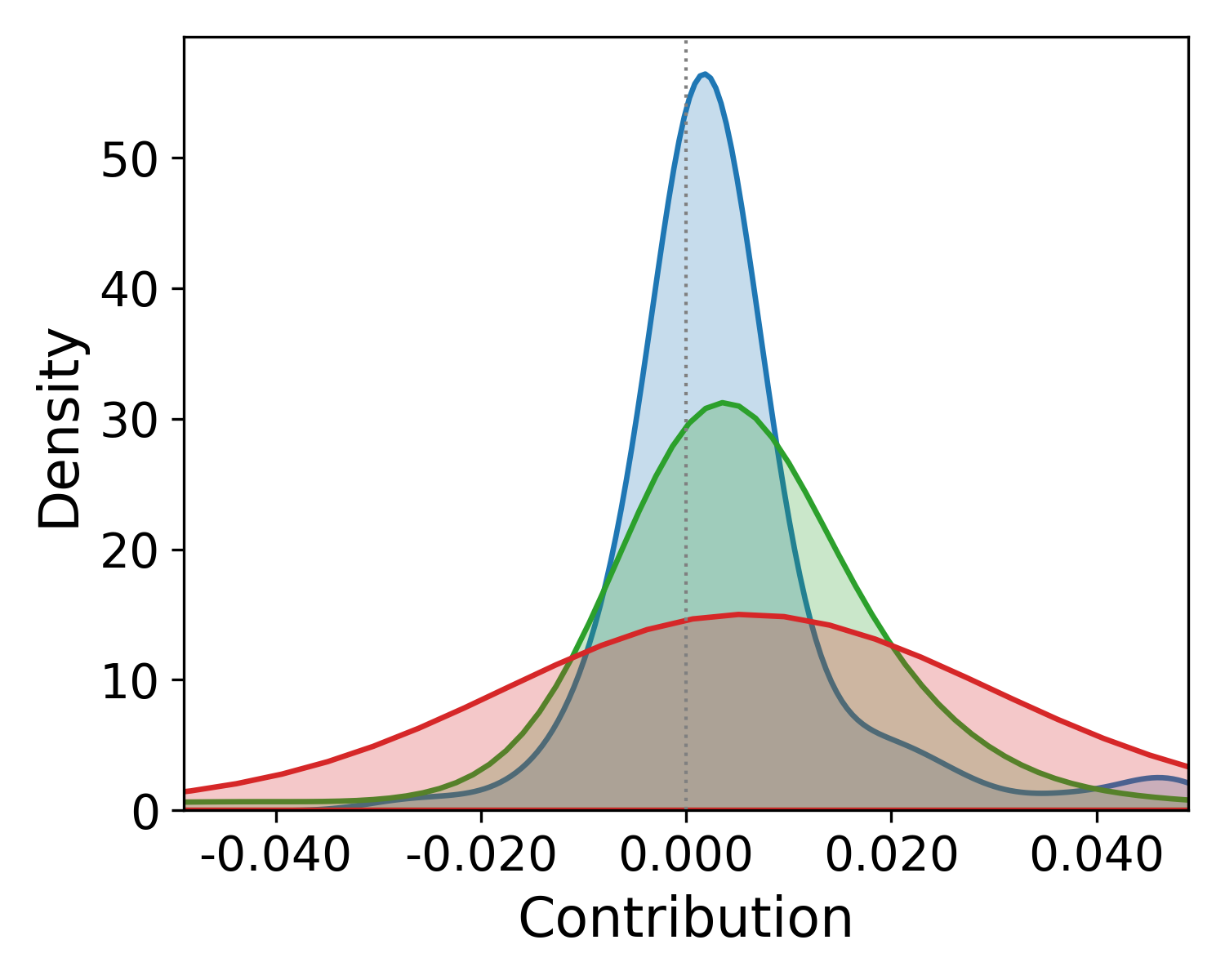}
        \caption{IMDB / NIID / LOO.}
    \end{subfigure}
    \hfill
    \begin{subfigure}{0.22\textwidth}
        \includegraphics[width=\linewidth]{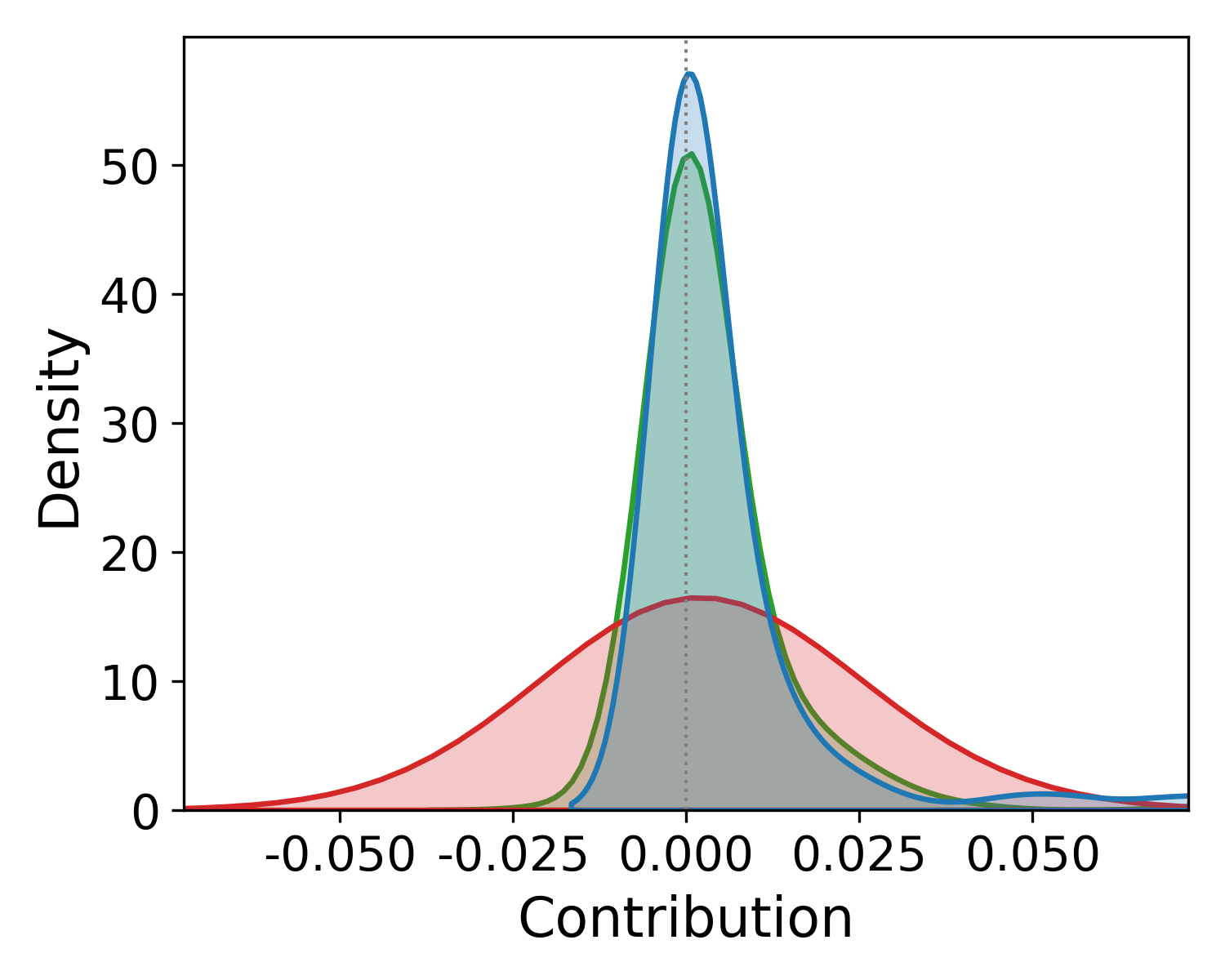}
        \caption{IMDB / NIID / GTG.}
    \end{subfigure}
    \hfill
    \begin{subfigure}{0.22\textwidth}
        \includegraphics[width=\linewidth]{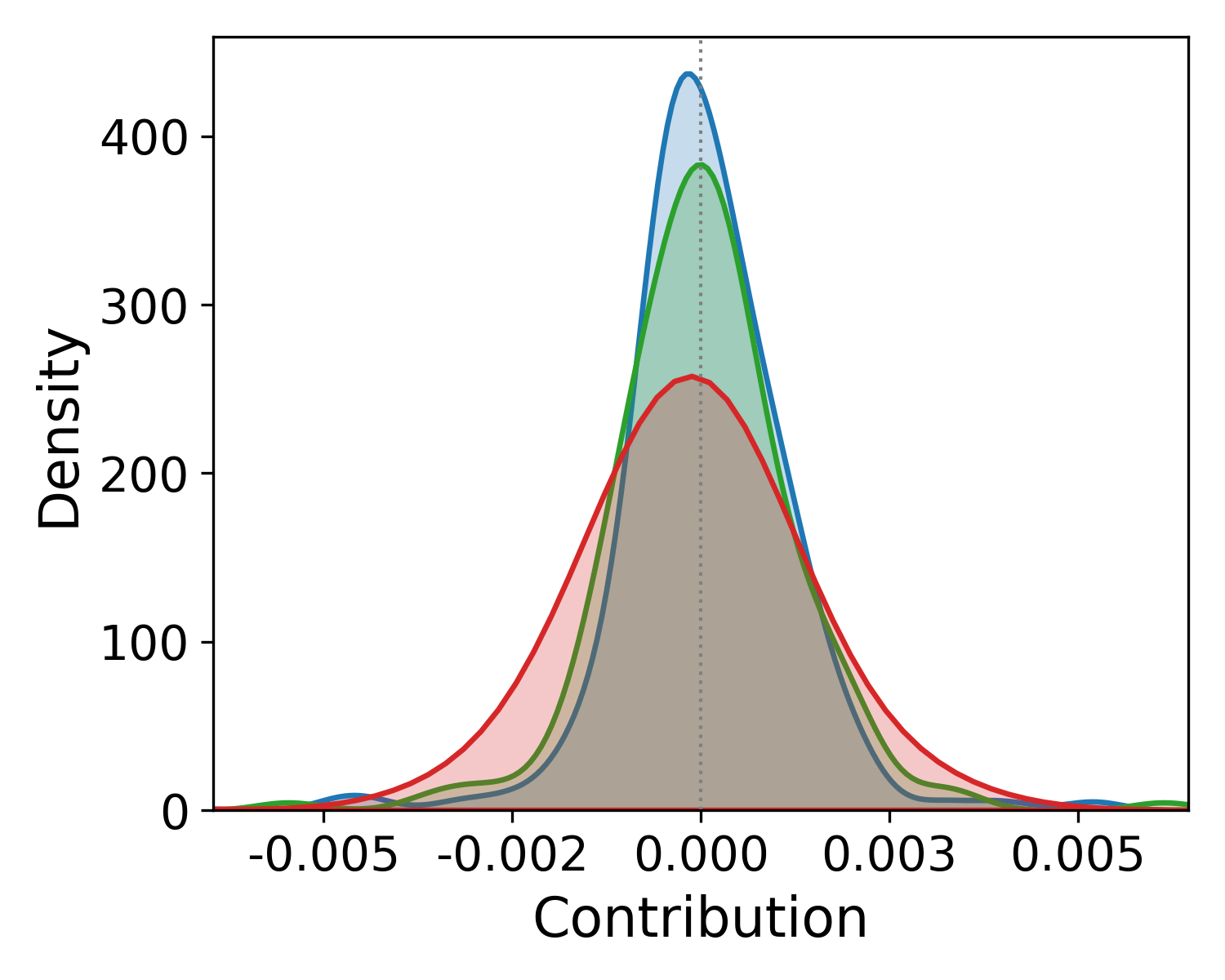}
        \caption{IMDB / IID / LOO.}
    \end{subfigure}
    \hfill
    \begin{subfigure}{0.22\textwidth}
        \includegraphics[width=\linewidth]{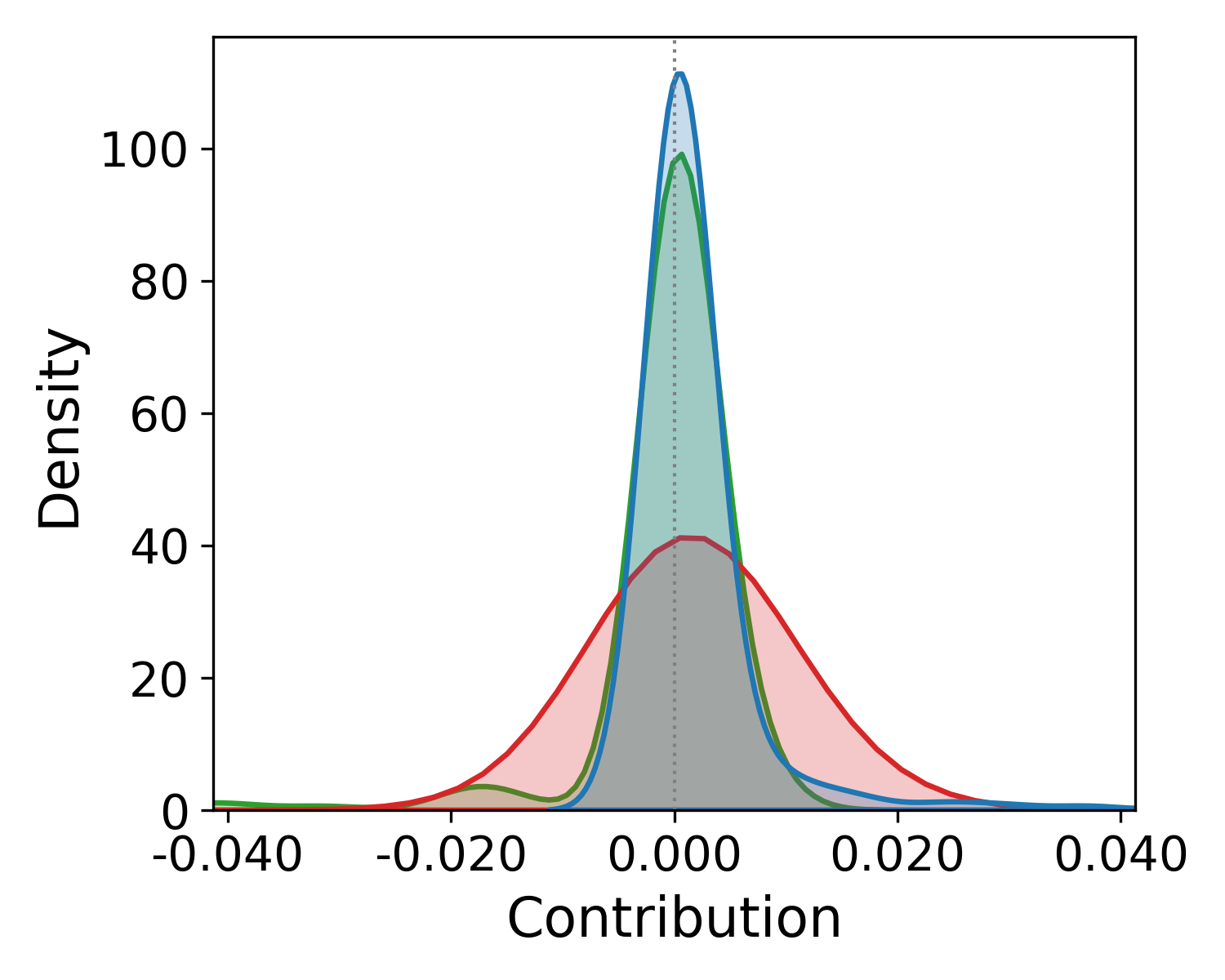}
        \caption{IMDB / IID / GTG.}
    \end{subfigure} 
    
    \caption{Score distributions of the 4 clients.}
    \label{fig:score_dist_4}
\end{figure}

\begin{table}[!b]
    \centering
    \setlength{\tabcolsep}{1.9pt}
    \begin{tabular}{ccc||cc|cc|cc}
        \multicolumn{3}{c||}{\multirow{2}{*}{\textbf{Setting}}}
        & \multicolumn{2}{c|}{\textbf{$\mathtt{fair}$}}
        & \multicolumn{2}{c|}{\textbf{$\mathtt{rel}$}}
        & \multicolumn{2}{c}{\textbf{$\mathtt{res}$}} \\
        &&& $\Phi$ & $L_2$ & $\Phi$ & $L_2$ & $\Phi$ & $L_2$ \\
        \hline\hline

        \multirow{4}{*}{\rotatebox{90}{ADULT}} & \multirow{2}{*}{4} & I &
        0.0006 & 0.0099 & 0.0912 & 0.0017 & -0.0982 & 0.0079 \\
        && N &
        -0.0451 & 0.1099 & -0.2477 & 0.0481 & -0.0309 & 0.0602 \\
        \cline{2-9}
        & \multirow{2}{*}{20} & I &
        -0.0655 & 0.0024 & -0.0845 & 0.0006 & -0.2608 & 0.0023 \\
        && N &
        -0.2091 & 0.0959 & -0.3382 & 0.0238 & 0.0364 & 0.0745 \\
        \hline

        \multirow{4}{*}{\rotatebox{90}{CIFAR}} & \multirow{2}{*}{4} & I &
        0.0595 & 0.0269 & -0.0190 & 0.0465 & -0.1259 & 0.0401 \\
        && N &
        0.0546 & 0.0500 & -0.3671 & 0.0599 & -0.4226 & 0.0762 \\
        \cline{2-9}
        & \multirow{2}{*}{20} & I &
        -0.1337 & 0.0032 & -0.2508 & 0.0094 & -0.3152 & 0.0177 \\
        && N &
        -0.0362 & 0.0187 & -0.3736 & 0.0577 & -0.5544 & 0.2035 \\
        \hline

        \multirow{4}{*}{\rotatebox{90}{IMDB}} & \multirow{2}{*}{4} & I &
        \textemdash & \textemdash & -0.0669 & 0.0015 & -0.0517 & 0.0028 \\
        && N &
        \textemdash & \textemdash & 0.2548 & 0.0257 & 0.1957 & 0.0630 \\
        \cline{2-9}
        & \multirow{2}{*}{20} & I &
        \textemdash & \textemdash & 0.0097 & 0.0011 & 0.0261 & 0.0018 \\
        && N &
        \textemdash & \textemdash & 0.1116 & 0.0420 & -0.0556 & 0.1404 \\

    \end{tabular}
    \caption{Comparing the $\mathtt{L1O}$-based performance score to the trustworthy scores where I and N stands for IID and non-IID, respectively.}
    \label{tab:score_diff_LOO}
\end{table}

\begin{figure}[!t]
    \centering    
    \begin{subfigure}{0.3\textwidth}
        \includegraphics[width=\linewidth]{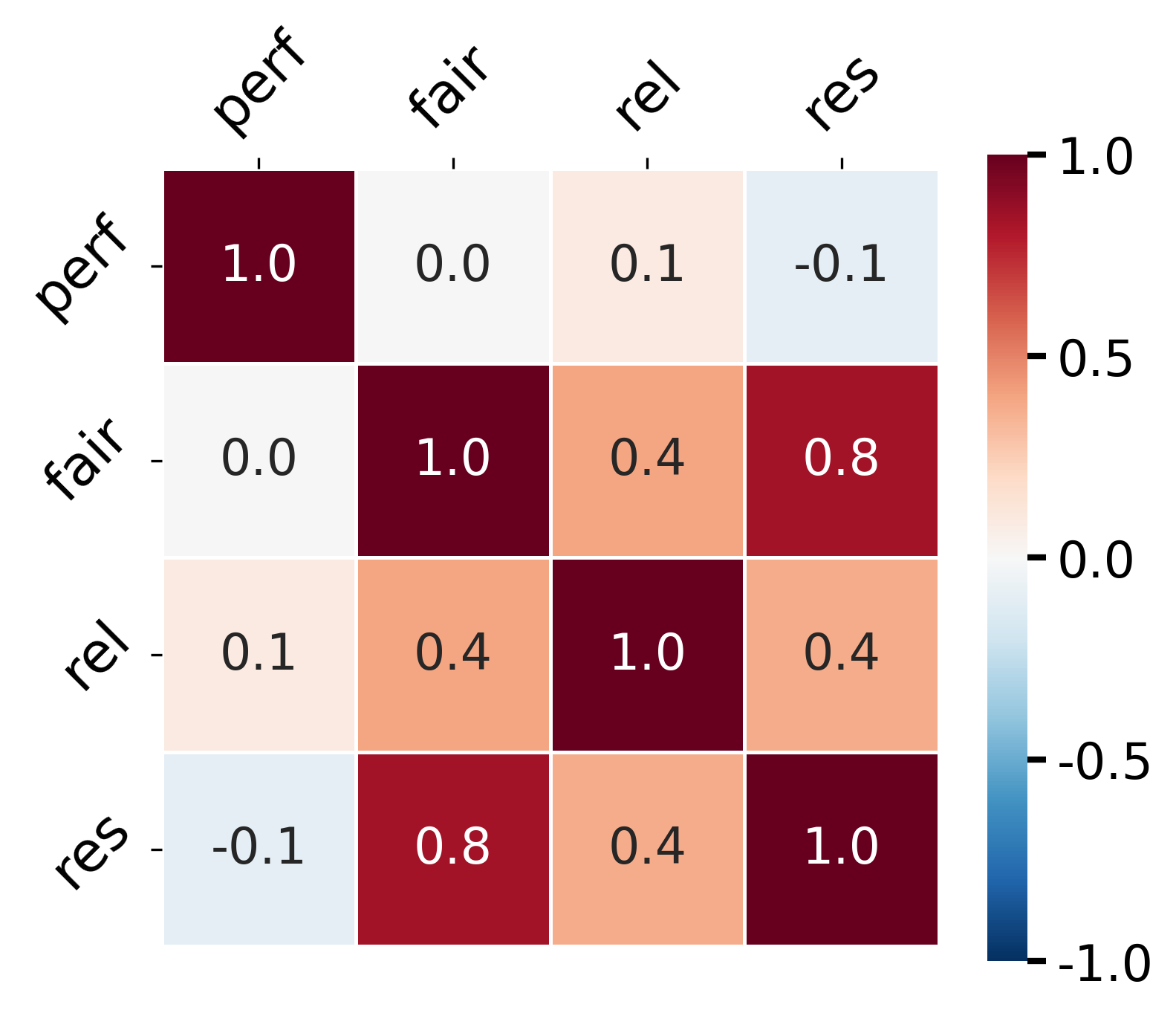}
        \caption{ADULT / LOO.}
    \end{subfigure}
    \hfill
    \begin{subfigure}{0.3\textwidth}
        \includegraphics[width=\linewidth]{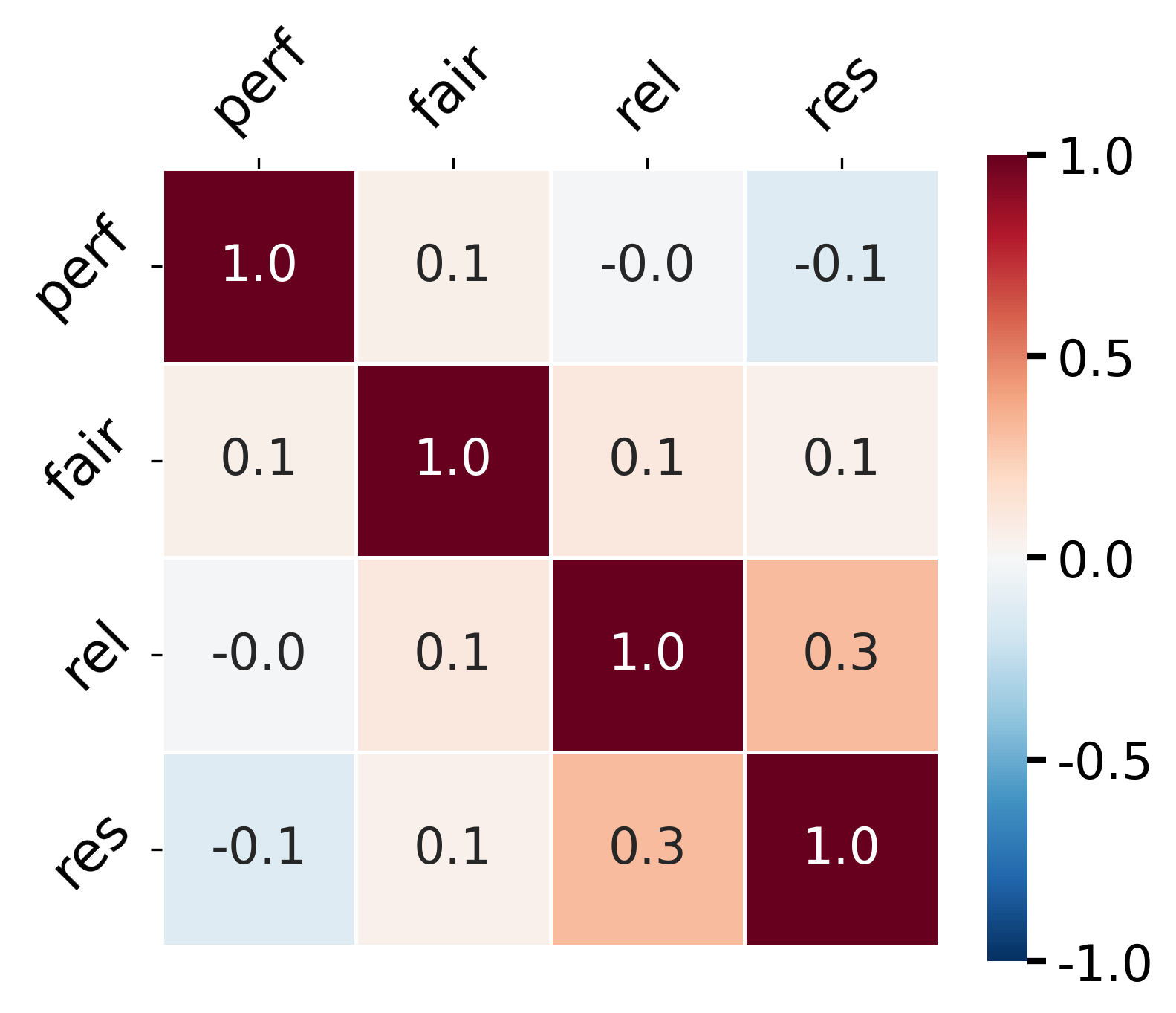}
        \caption{CIFAR / LOO.}
    \end{subfigure}
    \hfill
    \begin{subfigure}{0.3\textwidth}
        \includegraphics[width=\linewidth]{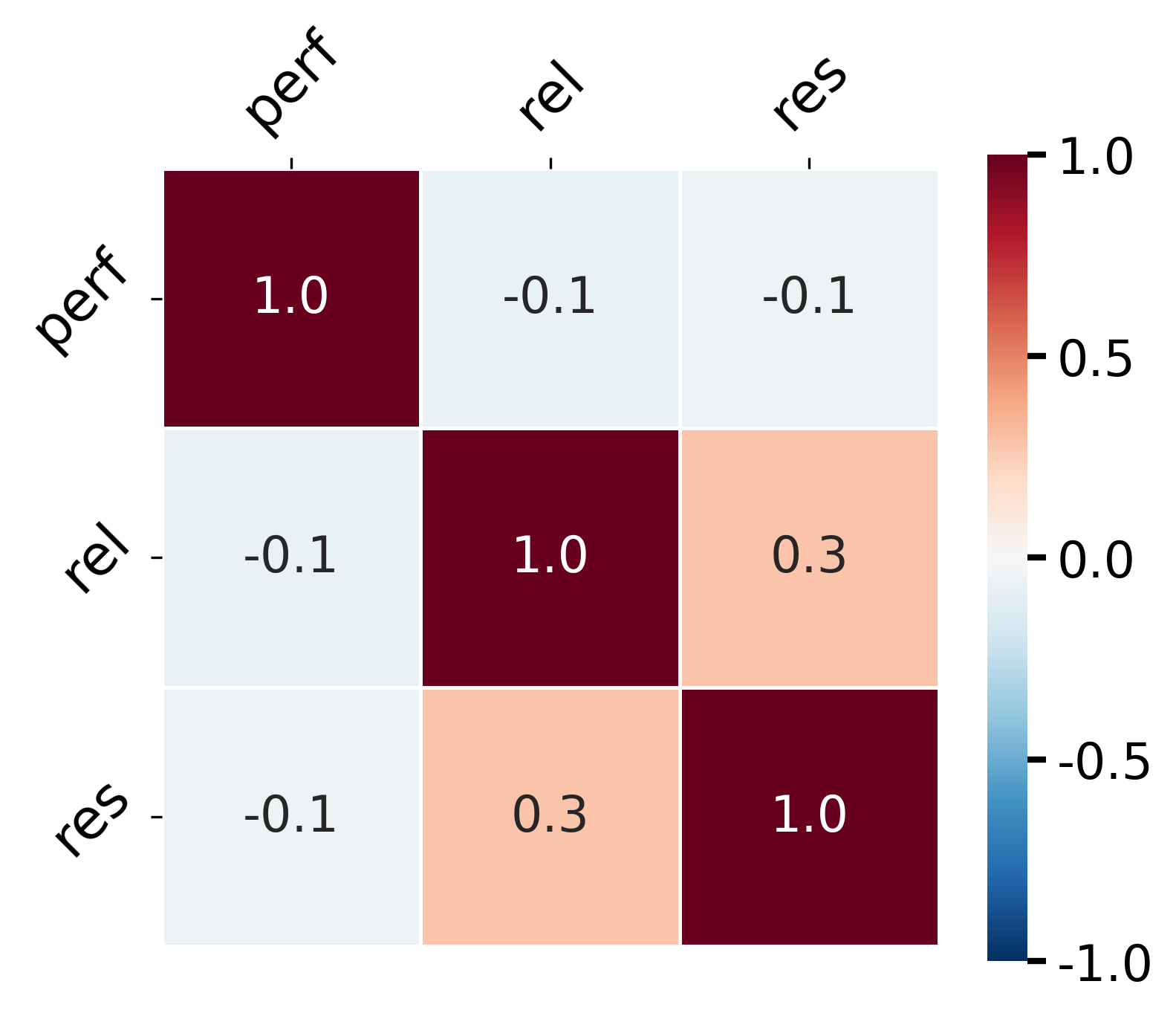}
        \caption{IMDB / LOO.}
    \end{subfigure}
    
    \begin{subfigure}{0.3\textwidth}
        \includegraphics[width=\linewidth]{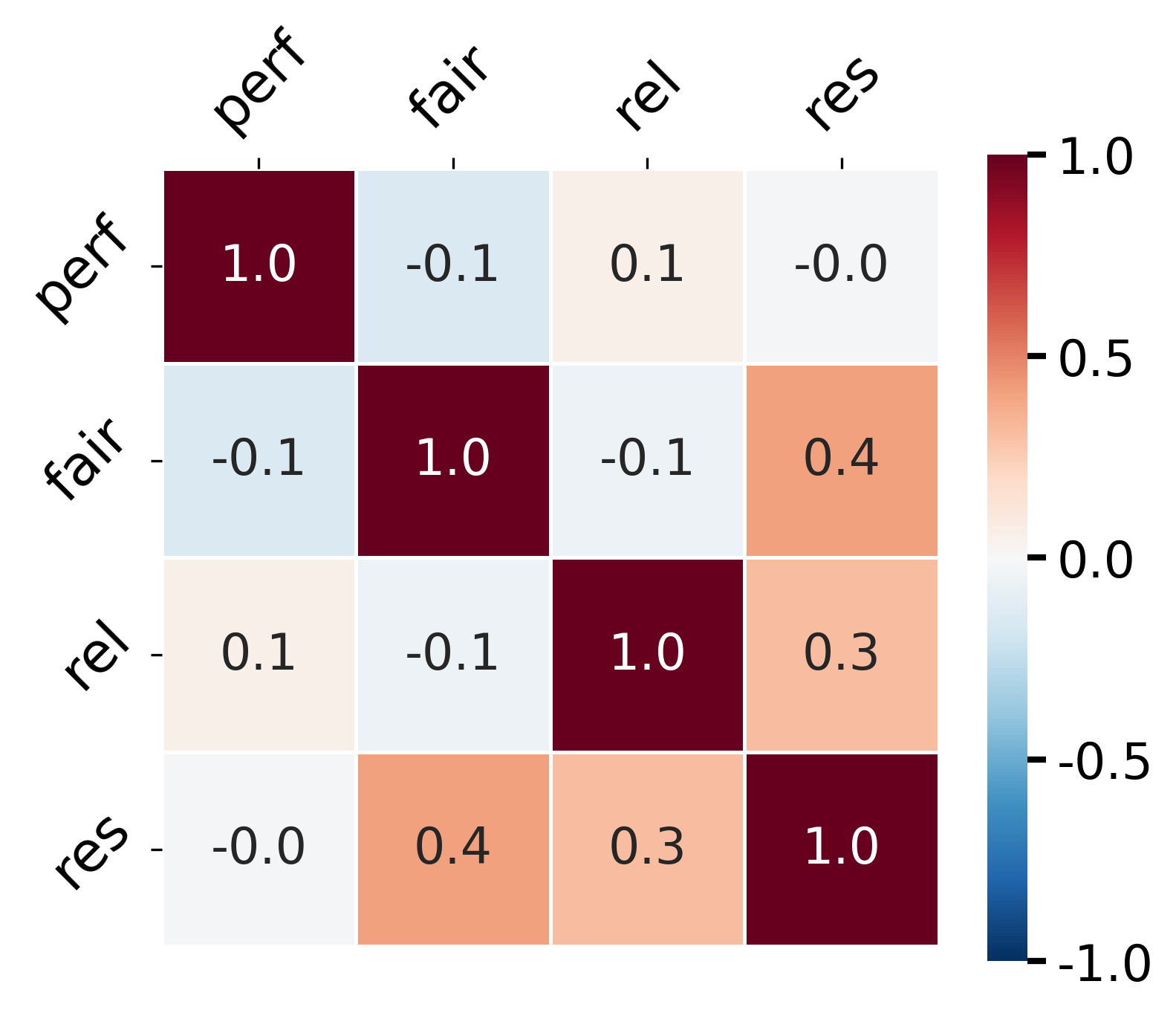}
        \caption{ADULT / GTG.}
    \end{subfigure}
    \hfill
    \begin{subfigure}{0.3\textwidth}
        \includegraphics[width=\linewidth]{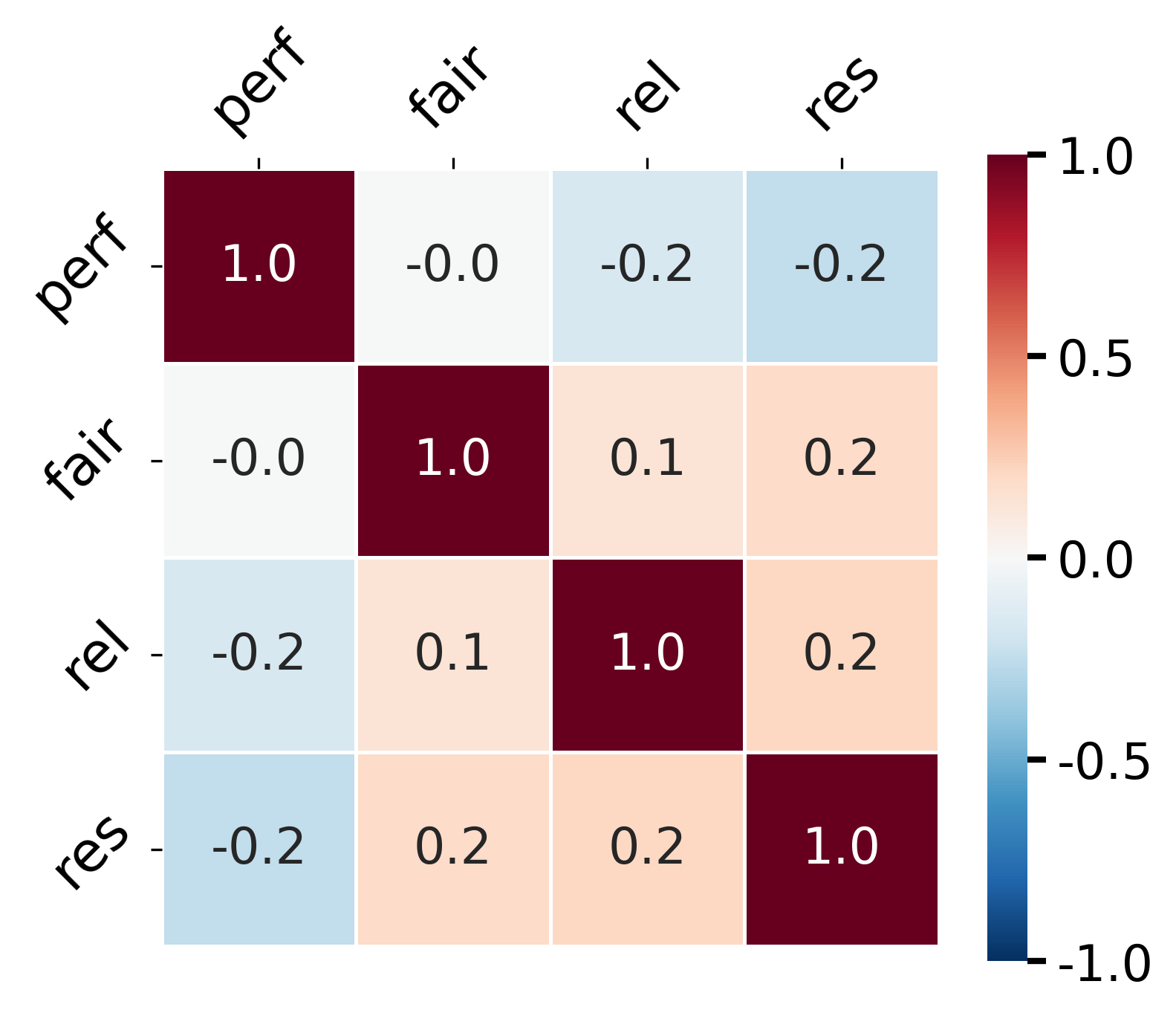}
        \caption{CIFAR / GTG.}
    \end{subfigure}
    \hfill
    \begin{subfigure}{0.3\textwidth}
        \includegraphics[width=\linewidth]{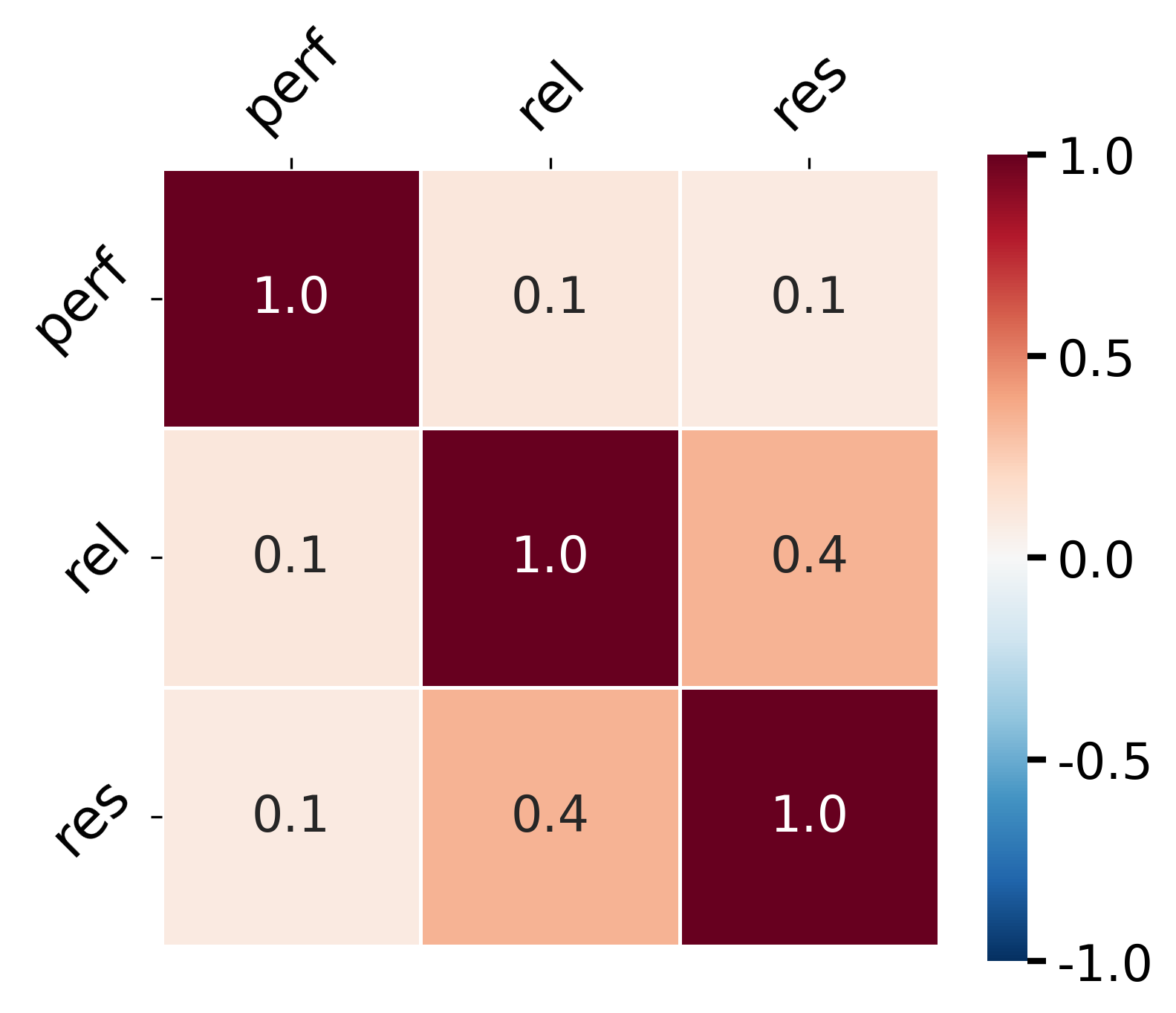}
        \caption{IMDB / GTG.}
    \end{subfigure}
    \caption{Heatmap of the pair-wise score correlations for the 4 IID client setting.  .}
    \label{fig:score_heat_4_I}
\end{figure}

\begin{figure}[!t]
    \centering    
    \begin{subfigure}{0.22\textwidth}
        \includegraphics[width=\linewidth]{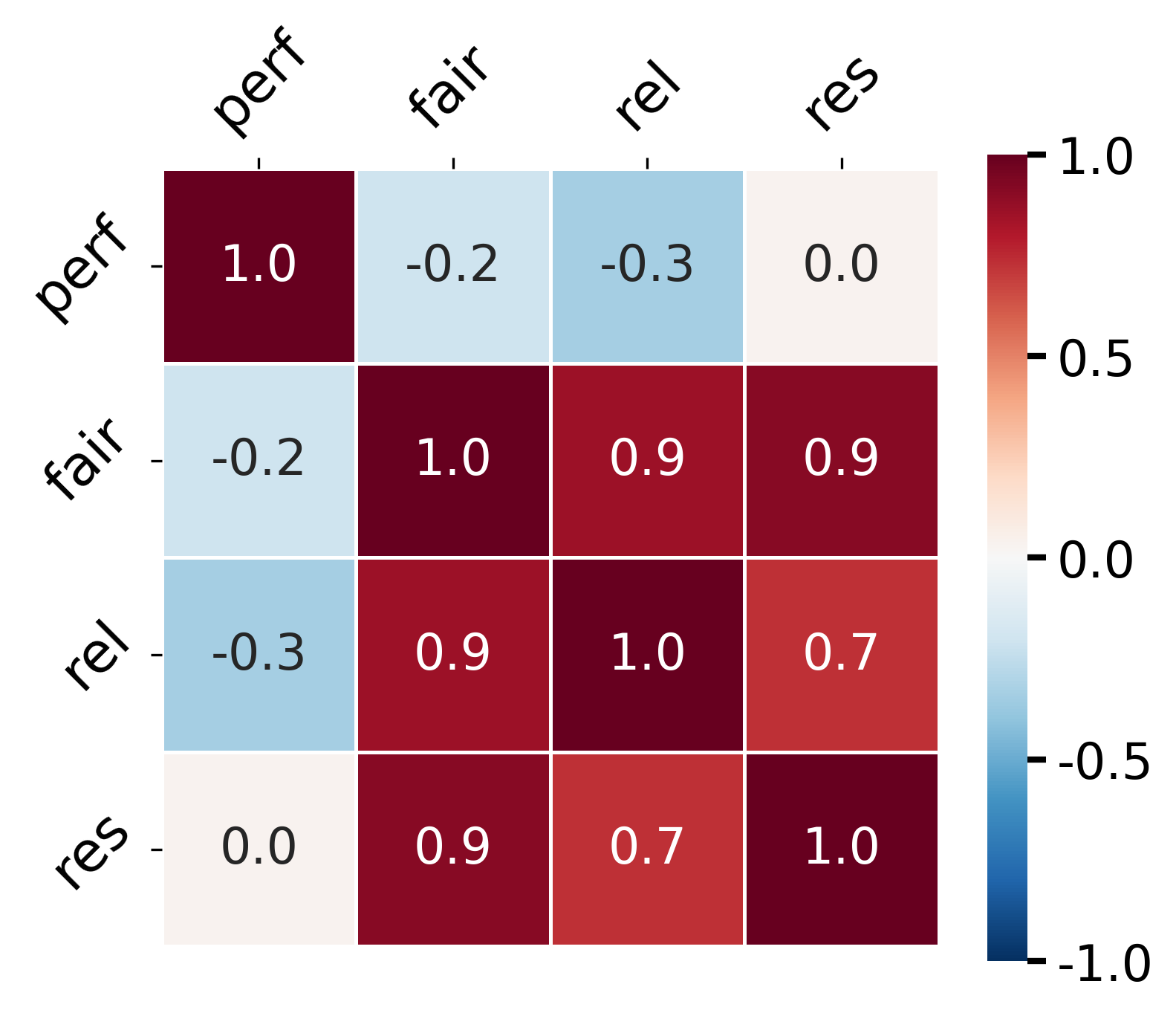}
        \caption{ADULT / NIID / LOO.}
    \end{subfigure}
    \hfill
    \begin{subfigure}{0.22\textwidth}
        \includegraphics[width=\linewidth]{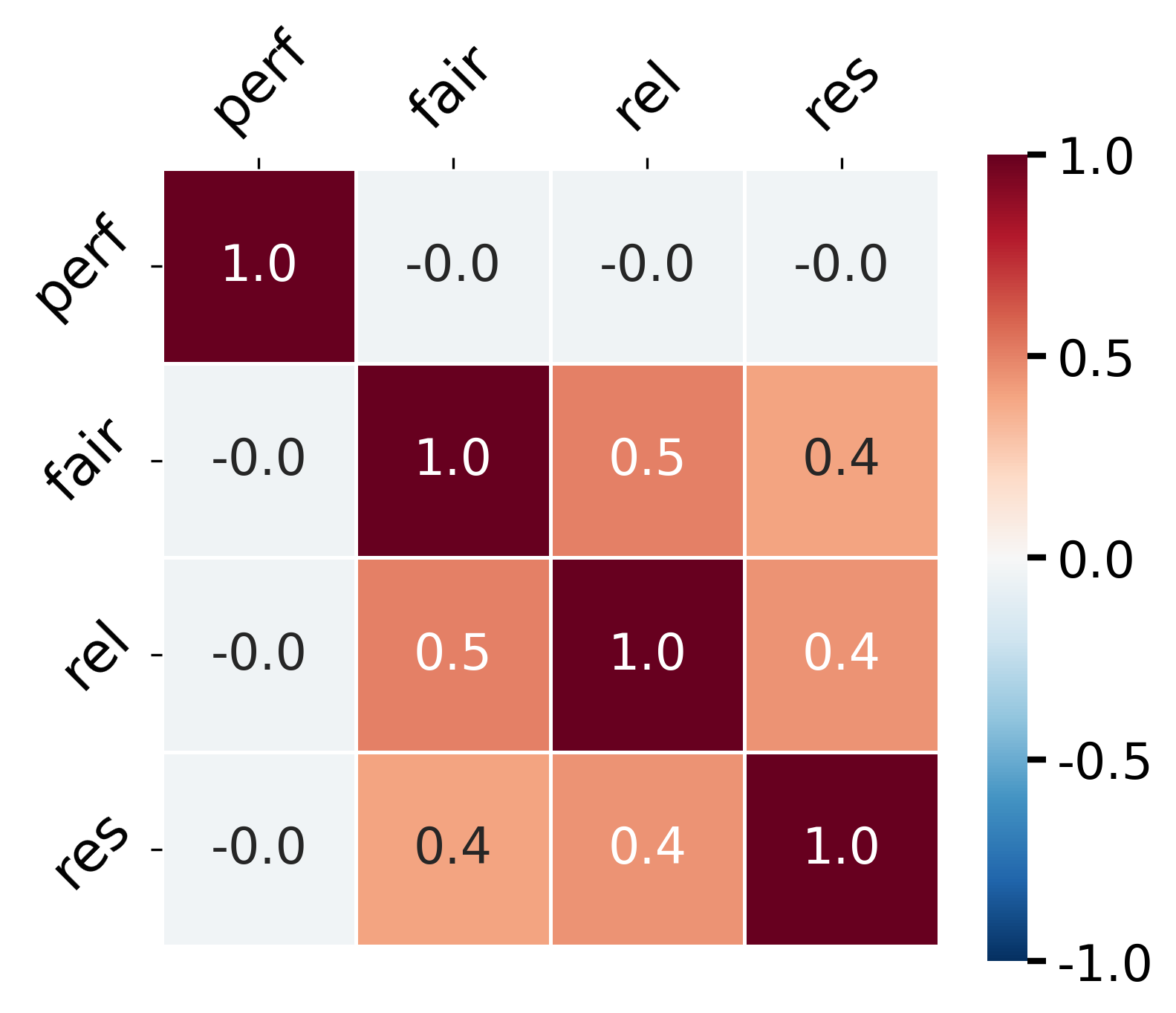}
        \caption{ADULT / NIID / GTG.}
    \end{subfigure}
    \hfill
    \begin{subfigure}{0.22\textwidth}
        \includegraphics[width=\linewidth]{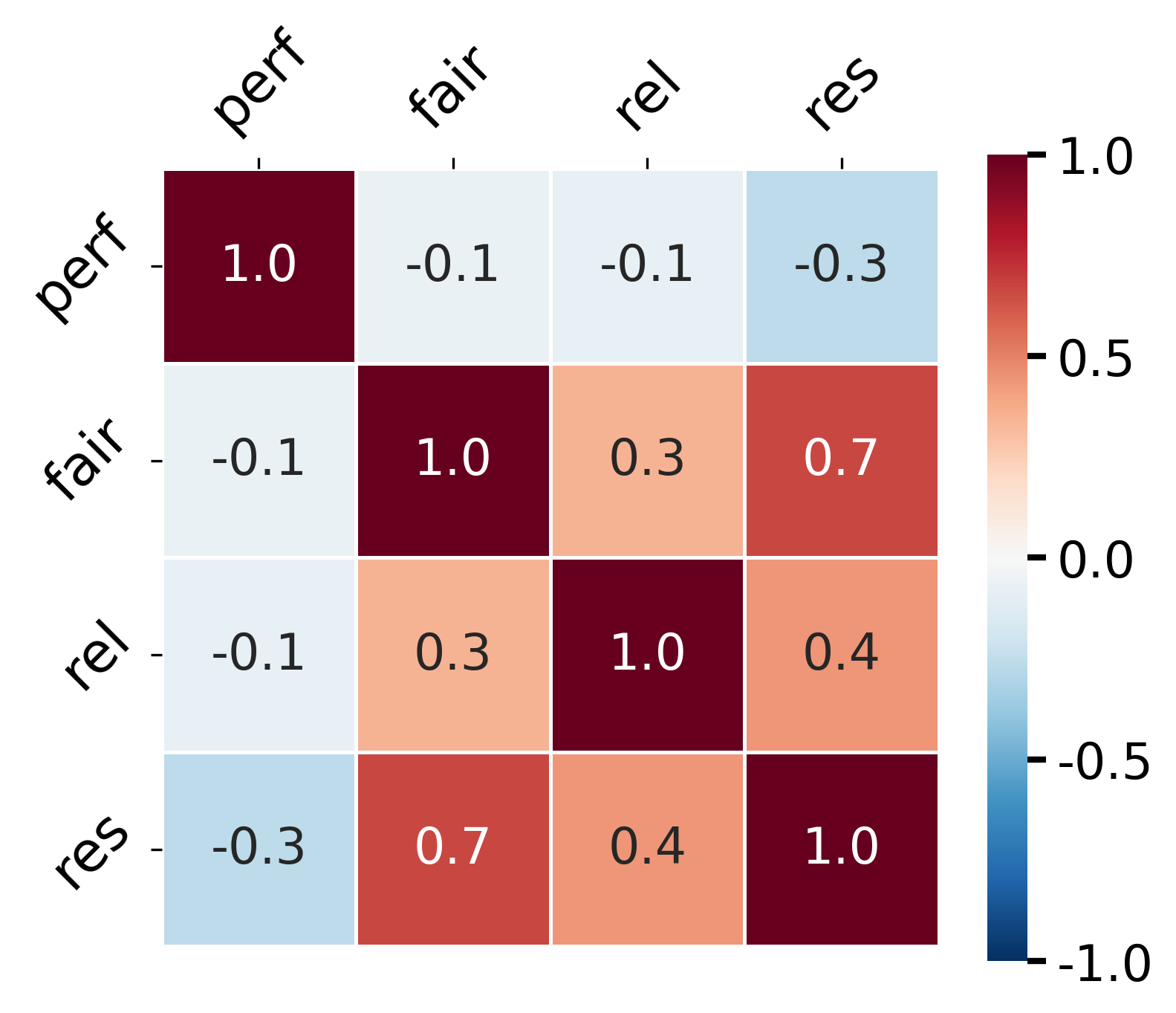}
        \caption{ADULT / IID / LOO.}
    \end{subfigure}
    \hfill
    \begin{subfigure}{0.22\textwidth}
        \includegraphics[width=\linewidth]{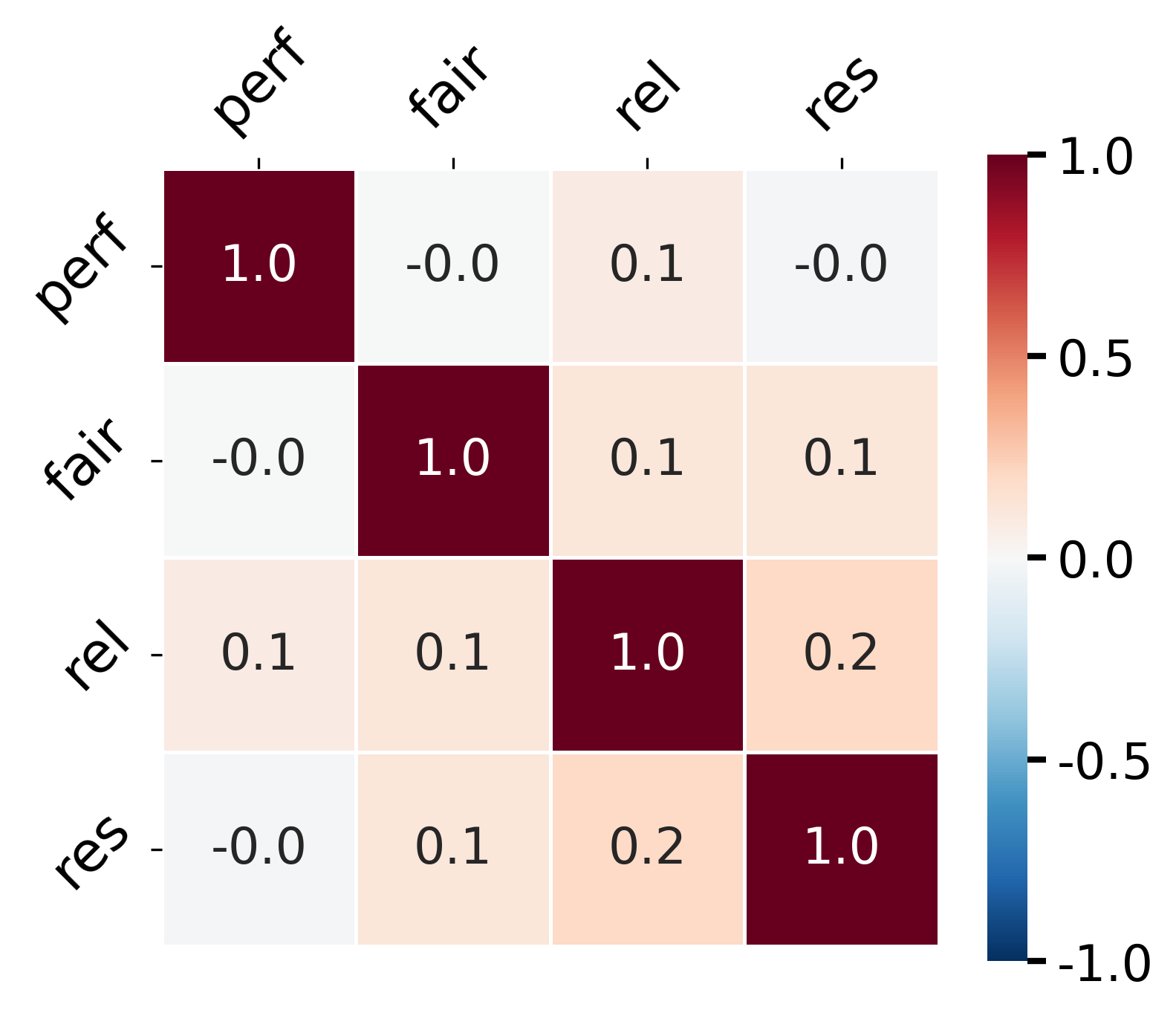}
        \caption{ADULT / IID / GTG.}
    \end{subfigure}
    
    \begin{subfigure}{0.22\textwidth}
        \includegraphics[width=\linewidth]{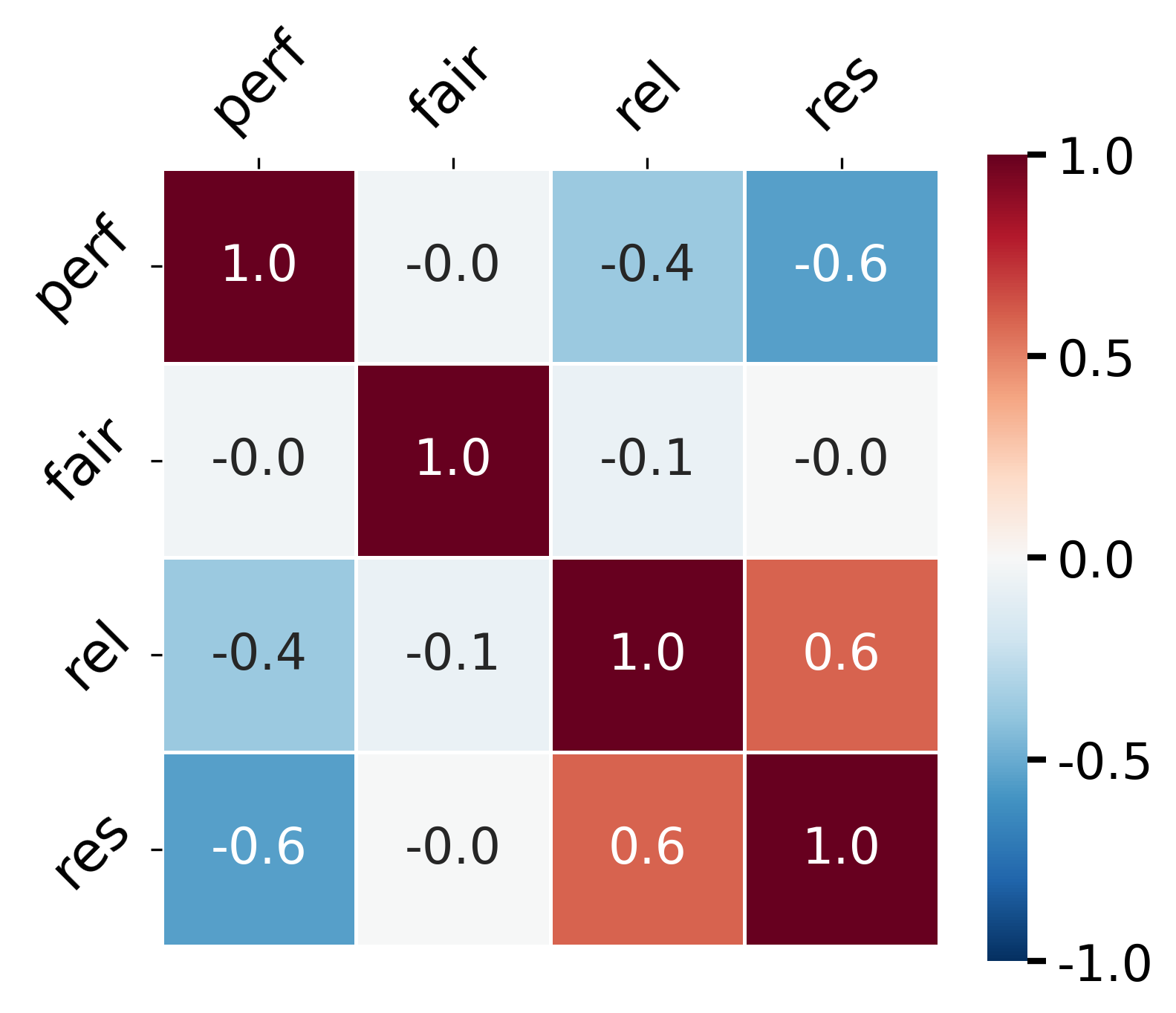}
        \caption{CIFAR / NIID / LOO.}
    \end{subfigure}
    \hfill
    \begin{subfigure}{0.22\textwidth}
        \includegraphics[width=\linewidth]{figs/H_20_C_N_loo.png}
        \caption{CIFAR / NIID / GTG.}
    \end{subfigure}
    \hfill
    \begin{subfigure}{0.22\textwidth}
        \includegraphics[width=\linewidth]{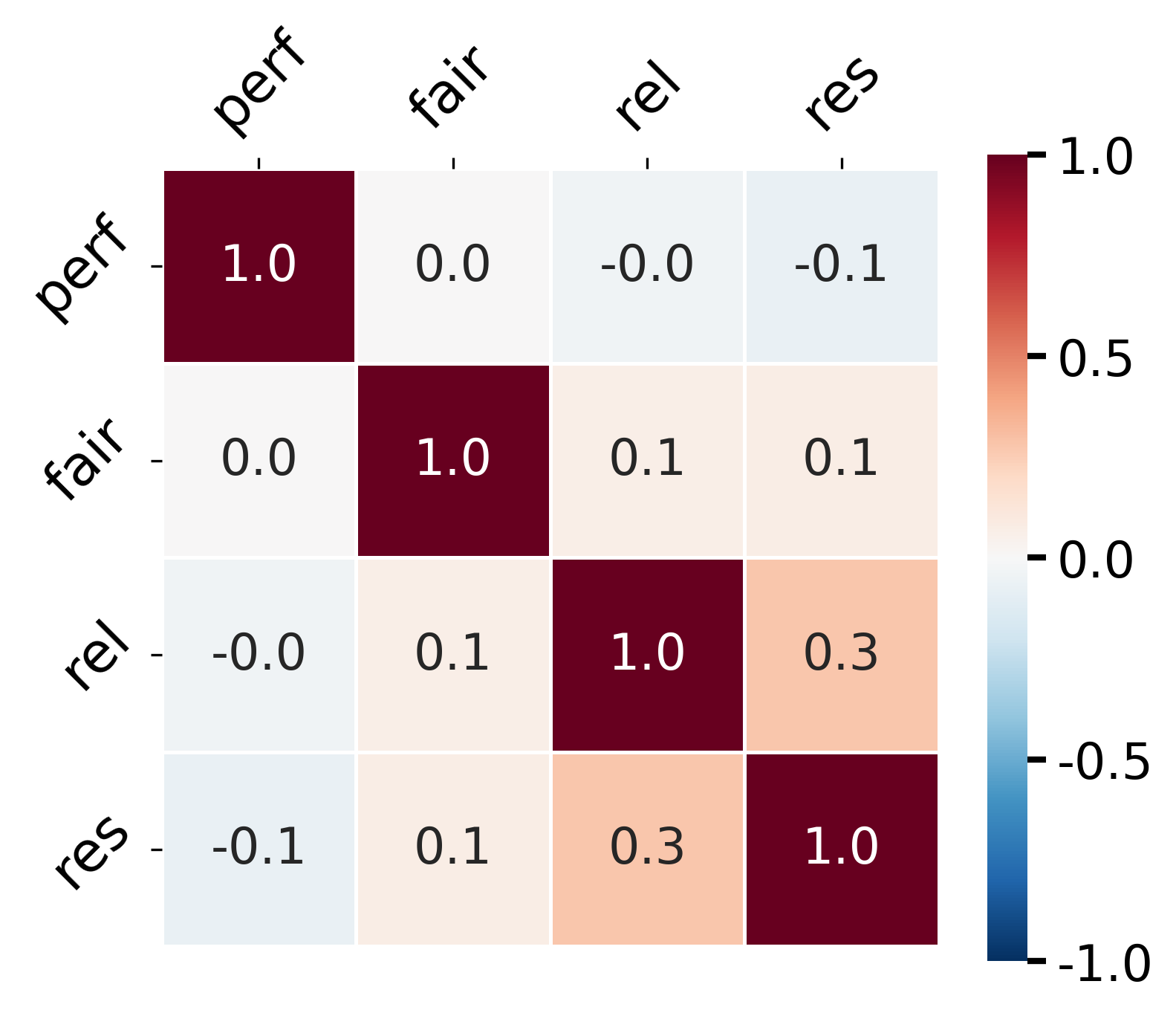}
        \caption{CIFAR / IID / LOO.}
    \end{subfigure}
    \hfill
    \begin{subfigure}{0.22\textwidth}
        \includegraphics[width=\linewidth]{figs/H_20_C_I_gtg.png}
        \caption{CIFAR / IID / GTG.}
    \end{subfigure}
    
    \begin{subfigure}{0.22\textwidth}
        \includegraphics[width=\linewidth]{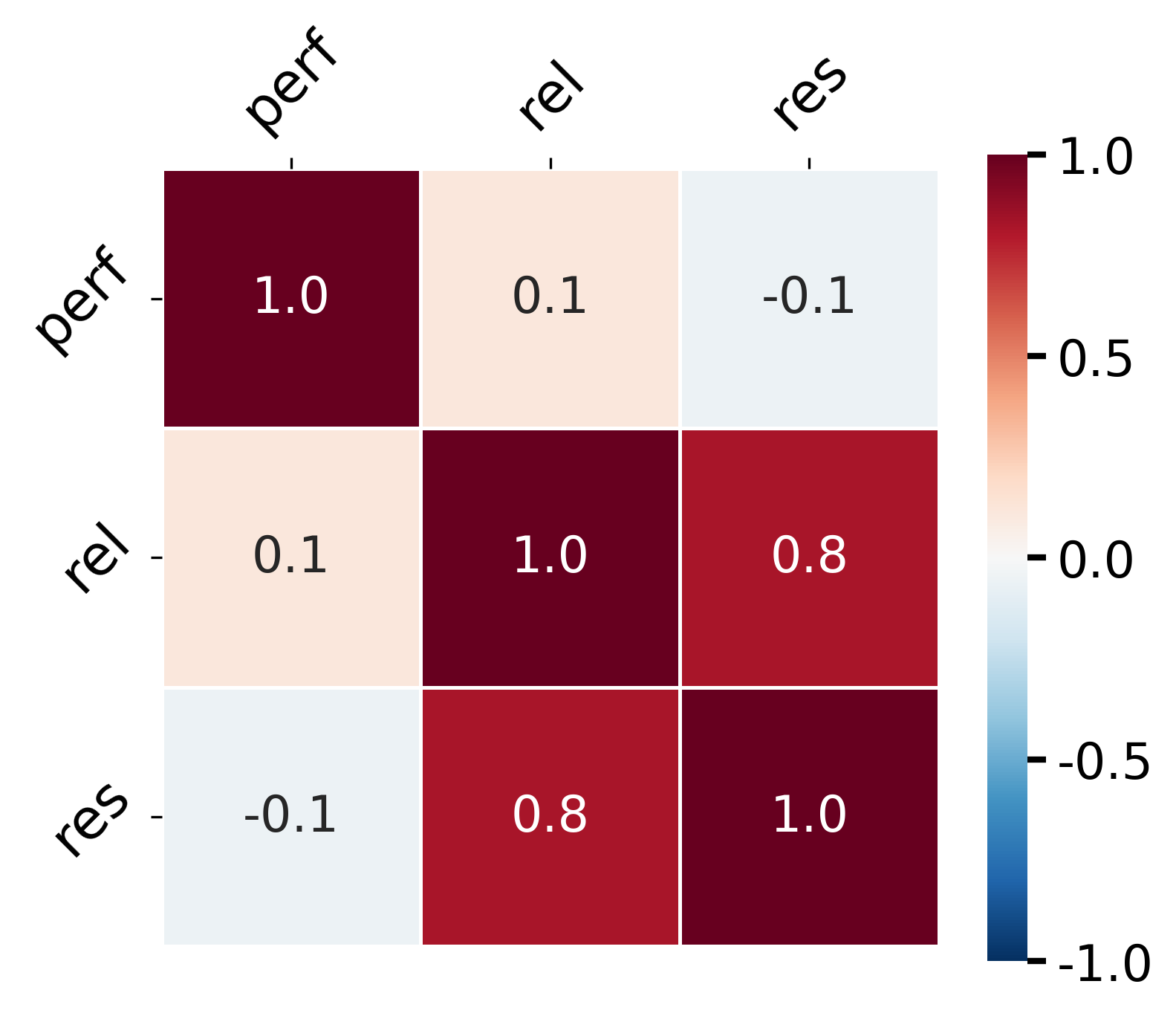}
        \caption{IMDB / NIID / LOO.}
    \end{subfigure}
    \hfill
    \begin{subfigure}{0.22\textwidth}
        \includegraphics[width=\linewidth]{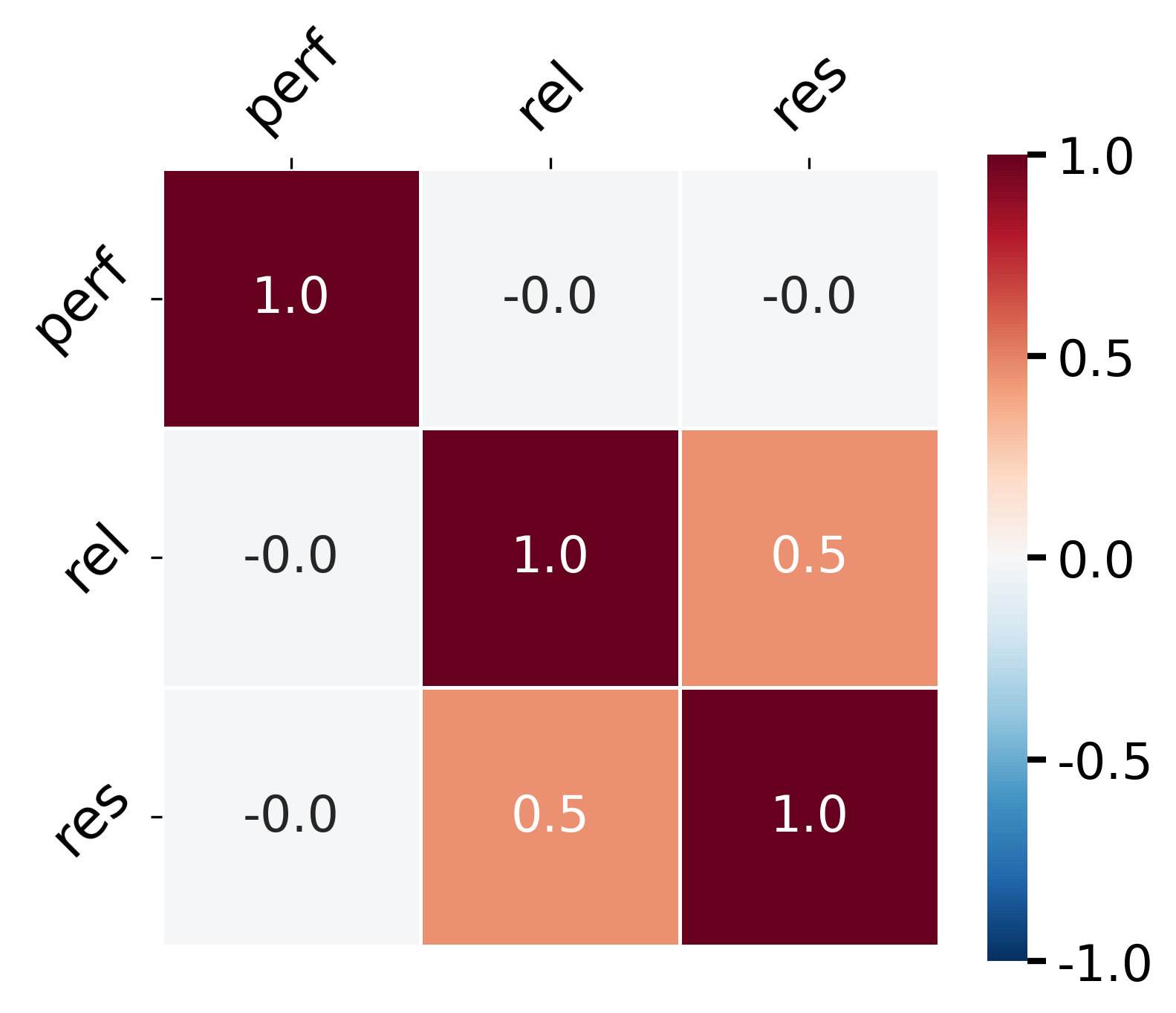}
        \caption{IMDB / NIID / GTG.}
    \end{subfigure}
    \hfill
    \begin{subfigure}{0.22\textwidth}
        \includegraphics[width=\linewidth]{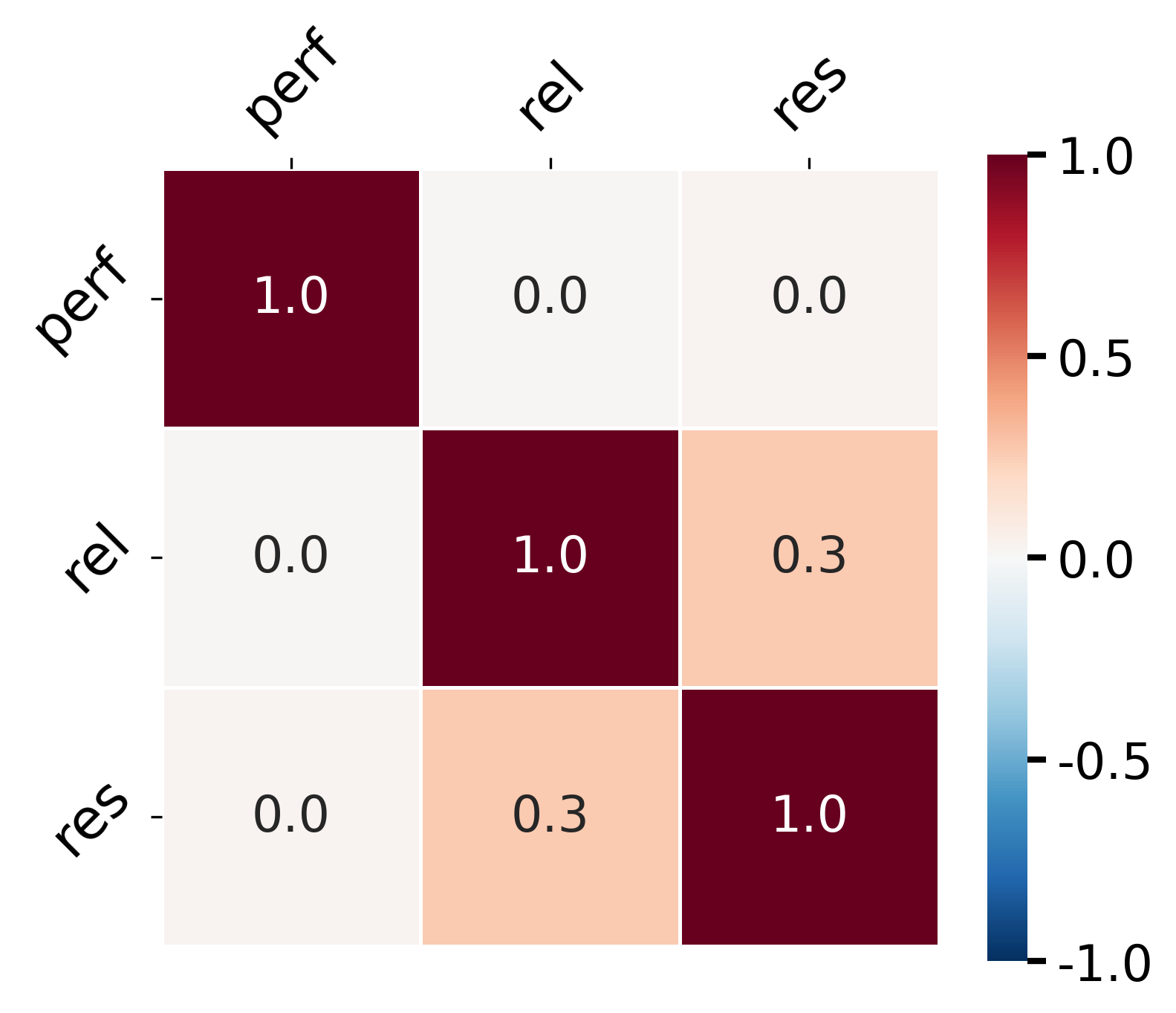}
        \caption{IMDB / IID / LOO.}
    \end{subfigure}
    \hfill
    \begin{subfigure}{0.22\textwidth}
        \includegraphics[width=\linewidth]{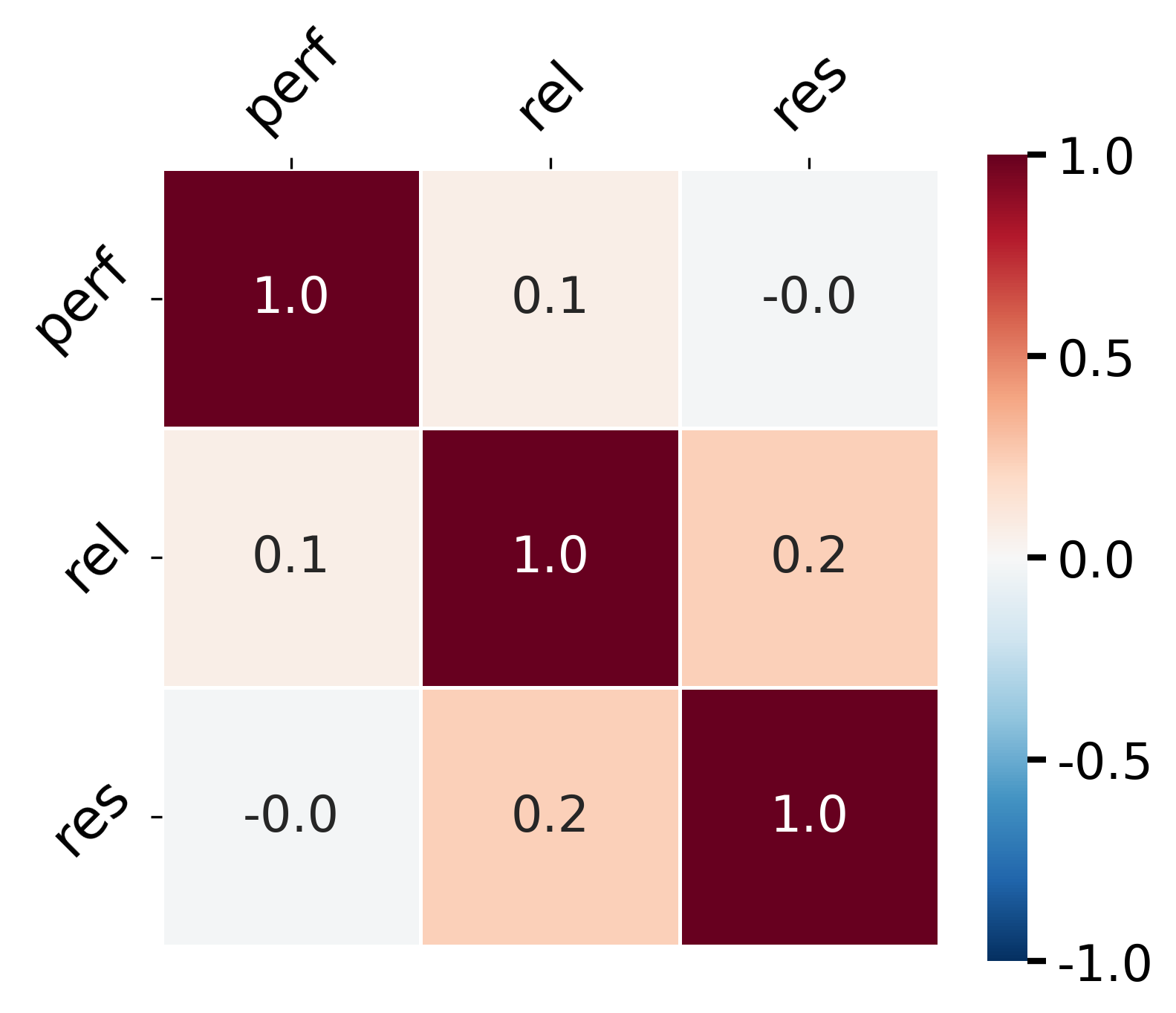}
        \caption{IMDB / IID / GTG.}
    \end{subfigure}
    \caption{Heatmap of the pair-wise csore correlations for the 20 client setting. }
    \label{fig:score_heat_20}
\end{figure}

\end{document}